\documentclass[11pt, a4paper, numbering]{deepmind}

\pdftrailerid{redacted}

\makeatletter
\renewcommand\bibentry[1]{\nocite{#1}{\frenchspacing\@nameuse{BR@r@#1\@extra@b@citeb}}}
\makeatother

\usepackage{kantlipsum, lipsum}
\usepackage{dsfont}
\usepackage{dm-colors}

\newcommand{\videourl}{https://www.youtube.com/watch?v=KHMwq9pv7mg}
\newcommand{\videourltext}{\url{https://www.youtube.com/watch?v=KHMwq9pv7mg}}
\newcommand{\episodesurl}{https://youtu.be/aWr5ADI_5sY}
\newcommand{\episodesurltext}{\url{https://youtu.be/aWr5ADI\_5sY}}
\newcommand{\drillepisodesurl}{https://youtu.be/BPzFJ3YhnZ4}
\newcommand{\drillepisodesurltext}{\url{https://youtu.be/BPzFJ3YhnZ4}}

\usepackage[numbers, sort&compress, round]{natbib}

\usepackage{times}

\usepackage[font=footnotesize,labelfont=bf]{caption}
\usepackage{subfigure} 
\usepackage{color}

\usepackage[utf8]{inputenc} %
\usepackage[T1]{fontenc}    %
\usepackage{hyperref}       %
\usepackage{url}            %
\usepackage{booktabs}       %
\usepackage{amsfonts}       %
\usepackage{nicefrac}       %
\usepackage{microtype}      %
\usepackage{diagbox}
\usepackage{xcolor}
\usepackage{float}
\usepackage{graphicx}
\usepackage{titlesec}
\usepackage{multirow}
\usepackage{array}
\usepackage{adjustbox}

\usepackage{algorithm}
\usepackage{algorithmicx}
\usepackage{algpseudocode}
\usepackage{amsmath}
\usepackage{tabularx}
\usepackage{footnote}
\makesavenoteenv{tabular}
\makesavenoteenv{table}
\usepackage{graphicx,picture}

\graphicspath{{figures/}}

\title{From Motor Control to Team Play in Simulated Humanoid Football}

\correspondingauthor{guylever@google.com, liusiqi@google.com}


\reportnumber{} %

\renewcommand{\today}{2021-05-24} 

\author[*,1]{Siqi Liu} %
\author[*,1]{Guy Lever} %
\author[*,1]{Zhe Wang} %
\author[1]{Josh Merel} %
\author[1]{S. M. Ali Eslami} %
\author[1]{Daniel Hennes} %
\author[1]{Wojciech M. Czarnecki} %
\author[1]{Yuval Tassa} %
\author[1]{Shayegan Omidshafiei} %
\author[1]{Abbas Abdolmaleki} %
\author[1]{Noah Y. Siegel} %
\author[1]{Leonard Hasenclever} %
\author[1]{Luke Marris} %
\author[1]{Saran Tunyasuvunakool} %
\author[1]{H. Francis Song} %
\author[1]{Markus Wulfmeier} %
\author[1]{Paul Muller} %
\author[1]{Tuomas Haarnoja} %
\author[1]{Brendan D. Tracey} %
\author[1]{Karl Tuyls} %
\author[1]{Thore Graepel} %
\author[*,1]{Nicolas Heess} %

\affil[*]{Equal contributions}
\affil[1]{DeepMind}

\begin{abstract}

Intelligent behaviour in the physical world exhibits structure at multiple spatial and temporal scales.
Although movements are ultimately executed at the level of instantaneous muscle tensions or joint torques, they must be selected so as to serve goals defined on much longer timescales, and in terms of relations that extend far beyond the body itself, ultimately involving coordination with other agents. Recent research in artificial intelligence has shown the promise of learning-based approaches to the respective problems of complex movement, longer-term planning, and multi-agent coordination. However, there is limited research aimed at their integration.
We study this problem by training teams of physically simulated humanoid avatars to play football in a realistic virtual environment.
We develop a method that combines \emph{imitation learning}, single- and multi-agent \emph{reinforcement learning} and \emph{population-based training}, and makes use of transferable representations of behaviour for decision making at different levels of abstraction. 
In a sequence of training stages, players first learn to control a fully articulated body to perform realistic, human-like movements such as running and turning; they then acquire mid-level football skills such as dribbling and shooting; finally, they develop awareness of others and learn to play as a team, successfully bridging the gap between low-level motor control at a time scale of milliseconds, and coordinated goal-directed behaviour as a team at the timescale of tens of seconds. We investigate the emergence of behaviours at different levels of abstraction, as well as the representations that underlie these behaviours using several analysis techniques, including statistics from real-world sports analytics. Our work constitutes a complete demonstration of integrated decision-making at multiple scales in a physically embodied multi-agent setting. We provide footage of the learned football skills in the \textcolor{blue}{\href{\videourl}{supplementary video}}.\footnote{ {\href{\videourl}{\videourltext}}.}

\end{abstract}

\begin{document}

\maketitle

\newcommand{\expect}[2]{\mathds{E}_{{#1}} \left[ {#2} \right]}
\newcommand{\myvec}[1]{\boldsymbol{#1}}
\newcommand{\myvecsym}[1]{\boldsymbol{#1}}
\newcommand{\vx}{\myvec{x}}
\newcommand{\vy}{\myvec{y}}
\newcommand{\vz}{\myvec{z}}
\newcommand{\vtheta}{\myvecsym{\theta}}

\newcommand{\bE}{\mathbb{E}}
\newcommand{\bI}{\mathbb{I}}
\newcommand{\bR}{\mathbb{R}}
\newcommand{\cC}{\mathcal{C}}
\newcommand{\cR}{\mathcal{R}}
\newcommand{\cG}{\mathcal{G}}
\newcommand{\cF}{\mathcal{F}}
\newcommand{\cX}{\mathcal{X}}
\newcommand{\cP}{\mathcal{P}}
\newcommand{\cA}{\mathcal{A}}
\newcommand{\cM}{\mathcal{M}}
\newcommand{\cS}{\mathcal{S}}
\newcommand{\cT}{\mathcal{T}}
\newcommand{\cO}{\mathcal{O}}
\newcommand{\cU}{\mathcal{U}}
\newcommand{\cW}{\mathcal{W}}
\newcommand{\cJ}{\mathcal{J}}
\newcommand{\vpi}{\vec{\pi}}
\newcommand{\rulesep}{\unskip\ \vrule\ }
\newcommand{\prior}{\mu}
\newcommand{\noti}{\backslash{i}}
\newcommand{\notj}{\backslash{j}}

\newcommand{\npmp}{\pi}
\newcommand{\objective}{\Omega}
\newcommand{\traj}{\xi}
\newcommand{\coplayers}{{\pi'}}
\newcommand{\expert}{\bar\pi}

\newcommand{\toadd}[1]{{\color{green} #1}}
\newcommand{\maybedelete}[1]{{#1}}
\newcommand{\alternative}[1]{}

\newcommand{\guycom}[1]{{\color{blue} Guy: #1}}
\newcommand{\todo}[1]{{\color{red} To-Do: #1}}
\newcommand{\kt}[1]{{\color{orange} Karl: #1}}
\newcommand{\siqi}[1]{{\color{purple} Siqi: #1}}
\newcommand{\yuval}[1]{{\color{brown} yuval: #1}}
\newcommand{\nicolas}[1]{{\color{green} Nicolas: #1}}
\newcommand{\zw}[1]{{\color{pink} Zhe: #1}}
\newcommand{\so}[1]{{\color{orange} [Shayegan: #1]}}
\newcommand{\hennes}[1]{{\color{blue} [Daniel: #1]}}
\newcommand{\jm}[1]{{\color{red} [Josh: #1]}}
\newcommand{\ali}[1]{{\color{red} Ali: #1}}

\section{Introduction}
\label{sec:Introduction}

Allen Newell, in his classic remarks describing the foundations of both cognitive science and AI \citep{newell1990unified}, pointed out that human behaviour can be understood at multiple levels of organisation, ranging from millisecond-level muscle twitches, to cognitive-level decisions occurring on the order of hundreds of milliseconds or seconds, to longer-term socially informed goal-directed sequences, playing out over minutes, hours or days. 
To illustrate this, consider %
 three friends carrying a sofa up a flight of stairs to their new apartment: 
Although they are aiming at a goal which persists over many minutes, their actions will also be shaped by shorter-term considerations (get the sofa around a corner), and those actions are ultimately executed as muscle contractions on a much finer time-scale.  Furthermore, although the impact of each muscle contraction is most immediately on the body itself, each must be chosen to result in outcomes defined in terms of a much larger dynamic context, including both inanimate objects (the stairs, the sofa), as well as other agents (the friends) who are executing actions of their own. As Newell observed, the ability to coordinate across all these levels of abstraction is one of the most remarkable aspects of human behaviour, 
and it raises 
the question of how low-level motor commands are organised to support cognitive-level decisions, and ultimately high-level goals and social coordination.  %

In the years since Newell’s writing, remarkable progress has been made in understanding how intelligent behaviour can be generated, primarily through a research strategy that focuses on individual levels of abstraction at a time. Entire disciplines have dedicated themselves to understanding particular aspects of the full problem individually, studying motor control and goal-directed behaviour, \cite[e.g.][]{lashley1951problem,rosenbaum1987hierarchical,schank1977scripts,fuster2001prefrontal,merel2019hierarchical}, the origins of cooperative behaviour \cite[e.g.][]{smith2012evolution}, or the mechanisms underlying movement coordination in groups of animals and humans \cite[e.g.][]{sebanz2006joint}. In a similar vein, enabling machines to produce agile, animal-like movement has long been a goal of robotics research \cite[e.g.][]{raibert1986legged,kuindersma2016optimization}; and the naturalistic movement of physically simulated characters has been studied in the computer graphics community since the early days of animation \cite[e.g.][]{sims1994evolving,Faloutsos2001, yin2007simbicon, coros2010generalized, liu2012terrain}. Recently, learning-based approaches have successfully tackled a number of challenges in artificial intelligence, including problems requiring hierarchically structured behaviour and long-horizon planning \citep[e.g.][]{OpenAI_dota,VinyalsStarCraft} and multi-agent coordination \citep[e.g.][]{JaderbergCTF,MordatchEmergence}. They have also shown promise generating complex movement strategies for simulated \citep[e.g.][]{heess2017emergence,BansalEmergent, peng2018deepmimic,merel2020} and real-world embodied systems \citep[e.g.][]{lee2020learning,openai2019solving,peng2020learning}. However, the multi-scale organization of behaviour, highlighted by Newell and inherent to many real-world scenarios, continues to pose a problem for designers of embodied artificial intelligence systems. 

Although it has long been acknowledged that intelligent embodied systems require the integration of multiple levels of control \cite[e.g.][]{brooks1986robust,albus1993reference}, %
the principles that underlie the design of such systems remain poorly understood \cite[e.g.][]{merel2019hierarchical}, and successful examples are limited \citep[e.g.][]{stone2000layered,openai2019solving,gupta2019relay,merel2018hierarchical}.
For instance, a divide-and-conquer approach would suggest a decomposition into a hierarchy of modules with well-defined interfaces. But for many scenarios,  including the example above, a ``natural'' decomposition is often non-obvious, and an unsuitable one can significantly impair the performance of the system. Similarly, for learning-based approaches, the specification of a single objective function that would allow efficient learning of complex, multi-level behaviour can be difficult, as will its optimisation. And while a decomposition into multiple smaller learning problems may facilitate specification, credit assignment and exploration, it %
raises the question how such sub-problems should be integrated.
In this paper, we build on prior work on learning intelligent humanoid control \cite{peng2018deepmimic,peng2019mcp,merel2018hierarchical,MerelNPMP,merel2020} and investigate this problem through a case study of football with simulated humanoid players. We develop a framework based on deep reinforcement learning (Deep-RL)  \cite{MnihDQN, MnihA3C, SchulmanPPO, heess2015learning, LillicrapDDPG, AbdolmalekiMPO} that addresses several of the challenges associated with the acquisition of coordinated long-horizon behaviours and leads to the emergence of coordinated 2v2 humanoid football play.

Modern team sports %
highlight many of the challenges for integrated and coordinated decision making and motor control present in ethologically important activities. This has long been recognised in the robotics community where football, in particular, has been a grand challenge %
since 1996, with the aim of the RoboCup community to beat a human football team by 2050 \cite{KitanoRoboCup, RoboCupWeb}. 
Playing competitively in a game of football requires decisions at different levels of spatial and temporal abstraction -- ``low-level'' fast timescale control of the complex human body produces ``mid-level'' skills such as kicking and dribbling in the service of ``high-level'', long-term, goal-directed behaviour such as scoring as a team. Importantly, these levels of decision making are intimately coupled: for instance, the success or failure of a pass depends as much on a shared %
understanding of the situation and the players' ability to agree on a joint course of action as it depends on their ability to precisely control their movements. %
We introduce a simulated football environment that reflects a subset of the challenges of the full game of football, focusing especially on the problem of movement coordination. It extends the environment suite of \cite{merel2018hierarchical,MerelNPMP,merel2020,tassa2020dm_control} and comprises teams of fully articulated humanoid football players, capable of %
agile, naturalistic movements, while realistically simulated physics and the presence of other players allow complex coordinated strategies to emerge. The richness of possible behaviours, the need to coordinate movement with respect to a dynamic context including ball, goals, and other players, and the fact that low-level movement and high-level coordination are tightly coupled, without any obvious well-defined behavioural abstractions, make this setting a suitable testbed to study multi-scale decision making for embodied AI. While prior work has studied motor control, long-horizon behaviour, and multi-agent coordination in isolation, the football environment combines them into a single challenge. %

Our training framework consists of a three-stage procedure %
during which learning progresses gradually from imitation learning for low-level movement skills, to reinforcement learning of training drills for the acquisition of mid-level skills, to multi-agent reinforcement learning for full game play. %
This makes use of prior knowledge from imitation where available, while the auto-curriculum
that emerges from self-play in populations of learning agents allows the discovery of complex solutions that would be difficult to specify through reward or learn from imitation. 
With the gradual acquisition of skills of increasing complexity, the mix of different forms of learning that lie on a spectrum between %
imitation and deliberate practice as well as the repurposing of existing skills,
our framework bears some loose similarity to human learning \citep[e.g.][]{ericsson1993role,baker201420years,ashford2006observational,diedrichsen2015motor}. In particular, it provides a practical solution to challenges including behaviour specification, credit assignment, and exploration. Importantly, the framework exploits the modularity of the learning problem, and relies on explicit representations of low- and mid-level skills, but it still allows for seamless integration of the final behaviour across all levels of abstraction.
Although we instantiate our framework for football, the underlying principles %
are general and should be applicable in other domains, including other team sports or collaborative work scenarios \citep[e.g.][]{merel2020}.

We demonstrate that the training framework results in the emergence of sophisticated movement, football skills and team-level coordination. The players exhibit human-like, agile, and robust context-dependent movement and ball-handling skills such as getting up from the ground, rapid changes of direction, or dribbling around opponents to make accurate shots. These movement skills enable cooperative play, that progresses from individualistic behaviour to more coordinated team tactics such as moving into space, defensive positioning, and passing. 
We develop several techniques for the  quantitative analysis of the players' performance as well as their behavioural strategies and internal representations. We combine techniques previously employed in AI research \cite[e.g.][]{LiuSoccer,JaderbergCTF} with techniques from real world sports analytics \cite[e.g.][]{tuyls2020game,spearman2018beyond}. 
Game performance is positively correlated with robust movement skills, but also with coordination and team-level tactics as well as the ability to predict the behaviour of opponents and teammates. The players show an understanding of the value of teammates possessing the ball, and their intention to score or move the ball up-field, similar to observations made for human football players \citep[e.g.][]{gonccalves2015anticipation,roca2012developmental}. %

The \textcolor{blue}{\href{\videourl}{supplementary video}} provides an overview of the environment, training framework and agent behaviours.\footnote{{\href{\videourl}{\videourltext}}.} The paper is structured as follows: in Section \ref{sec:Environment} we introduce our novel multi-agent environment and in Section \ref{sec:Method} we discuss our training framework. In Section~\ref{sec:Experiments} we outline the experimental procedure. In Section \ref{sec:behaviour} we present results and provide a quantitative analysis of the evolution of individual players' movement skills and team-level strategies. In Section~\ref{sec:HowAgentsWork} we analyse the players' learned representations and ``understanding'' of the game. We provide an ablation of different components of our framework in Section \ref{sec:Ablation}. In Section \ref{sec:Related} we review different lines of research that are brought together in our work. We conclude with a discussion in Section \ref{sec:Discussion}.

\section{Environment}
\label{sec:Environment}

\begin{figure}
  \centering
  \includegraphics[width=\textwidth]{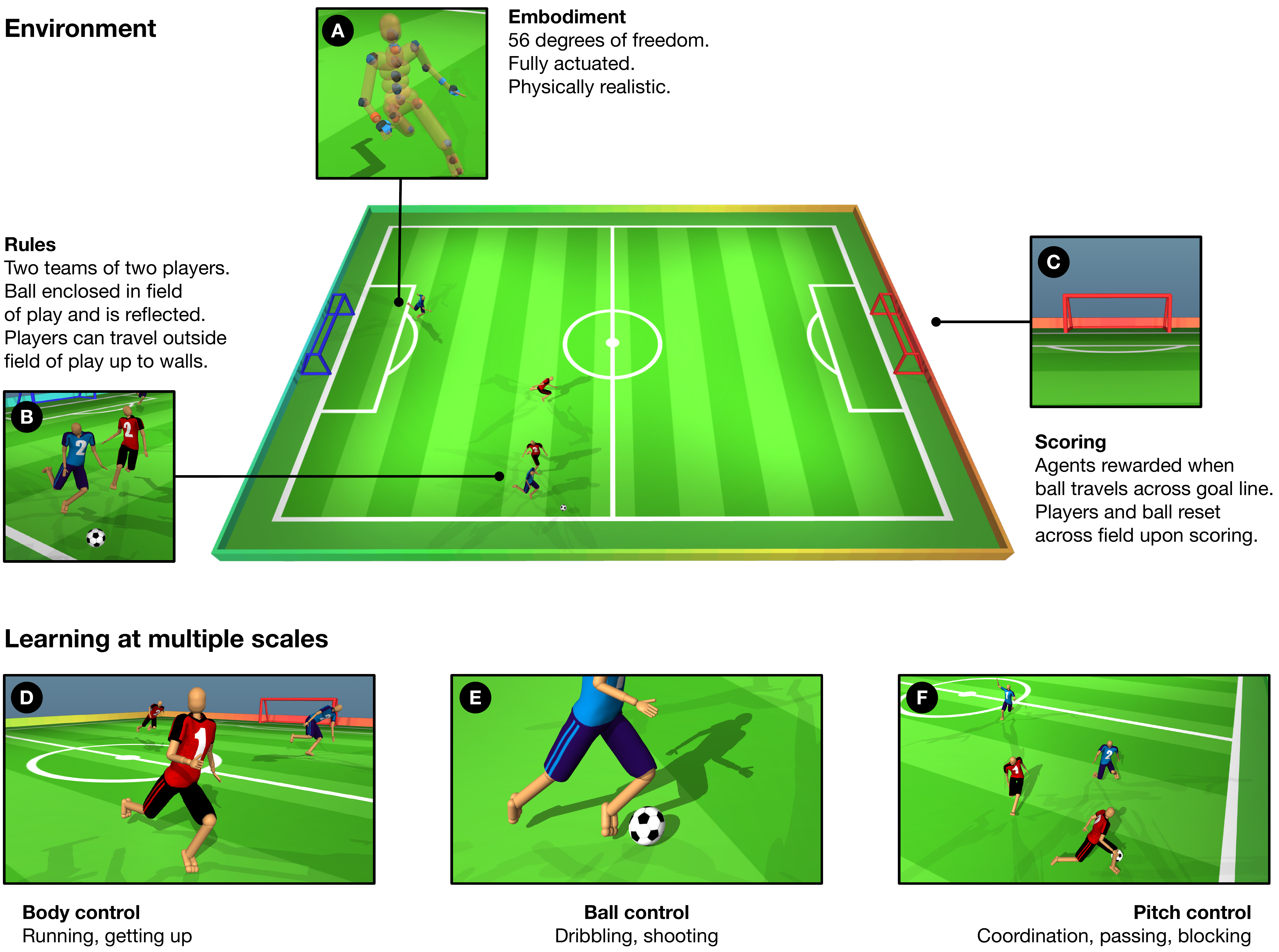}
  \caption{Overview of the humanoid football environment. {\bf (A)} Humanoids each have 56 degrees of freedom that are fully actuated (i.e.\ controllable by agents), and are situated in a high-fidelity MuJoCo physics simulation. {\bf(B)} We study 2v2 football. The ball reflects from the touchlines and goal lines (i.e.\ no throw-ins), however players can travel beyond, up to the physical hoardings. Contacts between all players, walls, goals and the ball are simulated. {\bf(C)} Both agents on a team receive reward when the ball travels through their opponents' goal, at which point all players and the ball are reset in random locations and the episode continues. We show that with an appropriate learning scheme in this setting, agents can be trained that exhibit proficiencies at multiple scales: {\bf(D)} Body control: moving their highly articulated bodies to stand, run, get up from falls and to jostle with opponents; {\bf(E)} Ball control: using their bodies to manipulate the position of the ball, moving it across the pitch, passing with teammates, and shooting towards the goal; {\bf(F)} Pitch control: using the aforementioned skills to pressure the opposing team, to take advantage of open space, and to gain situational advantages in order to score goals.}
  \label{fig:environment}
\end{figure}

For this study we extend the suite of simulated humanoid environments of \cite{merel2018hierarchical,MerelNPMP,merel2020,tassa2020dm_control} with a multi-agent football environment. It is designed to embed sophisticated motor control in a task that requires context dependent behaviour, multi-scale decision making, and multi-agent coordination in a setting suitable for end-to-end learning. We build on the environment of \cite{LiuSoccer}, and replace the 3 degrees of freedom player body with a fully articulated, 56 degrees of freedom humanoid used in studies of humanoid control \cite{merel2018hierarchical, MerelNPMP, merel2020}.\footnote{The environment implementation with both the simple embodiment and the complex, 56 degrees of freedom humanoid embodiment is open sourced at \url{http://git.io/dm_soccer}} Figure \ref{fig:environment} provides an overview of the environment. 

The environment adheres to a standardized environment interface \cite{tassa2020dm_control}
and is simulated by the MuJoCo physics engine \cite{TodorovMujoco}, which is used extensively in the machine learning and robotics research community \cite{heess2016learning, heess2017emergence, BansalEmergent, OpenAIGym, TassaControlSuite, riedmiller2018learning}.  The standardized environment interface makes it easy to use in reinforcement learning experiments, and it can be said to be realistic in the following sense: we make no simplifications of the rigid body dynamics; players bodies have realistic masses and joint force limits, and must locomote using torques applied at the joints, causing foot contact and friction forces with the pitch. That said, there are many aspects which are unrealistic with respect to human or robot football. Most notably, there are no neural delays, muscle dynamics, tendon-driven actuation or fatigue. However the principle of locomotion via controlled torques and friction remains intact.

Compared to existing simulation environments \cite{KitanoRoboCup, RoboCupWeb, LiuSoccer, kurach2019google}, including the RoboCup 2D and 3D leagues, ours emphasizes a specific subset of the full football challenge. We focus on movement coordination among small groups of highly articulated players rather than the full football problem.
On the one hand, the use of high-fidelity simulated physics provides potential for rich emergent behaviour, the complexity of which goes beyond that of prior work. The environment requires highly agile movements with a high-dimensional body for which behaviours would be difficult to handcraft. The primitive, joint-level action space without any form of action abstraction constitutes a significant learning challenge. The choice of action space imposes few restrictions and thus allows complex movements to emerge, including skilled dribbling and shooting, headers, or players shielding the ball with their body. This emphasizes the multi-scale nature of football, where movements have to be tightly coupled with higher level tactics and strategy without clearly separated levels of abstraction. 
Similar to real football, successful execution of a tackle or kick requires careful close-quarter positioning, foot placement, and balancing relative to the ball and opponent.

On the other hand, we simplify the football problem in ways not essential for the focus of this work. We do not attempt to model the full set of football rules, and also use a simpler set of rules than e.g.\ the RoboCup 3D simulation league (see Section~\ref{sec:Related} for further discussion).  Fewer interruptions (e.g. handballs are not illegal, there are no fouls and the ball is prevented from leaving the pitch to avoid special cases like throw-ins or goal kicks, see below) enable continuous gameplay and thus facilitate end-to-end learning. Furthermore, in the form used in this paper the agents perceive the environment (partially) via state features, relieving the agents from performing state estimation.\footnote{The environment permits egocentric vision but we do not use this feature. It is likely that the lack of a more realistic observation model discourages certain types of behaviours, like running backwards, or hanging back.} Finally, although our environment admits an arbitrary number of players, in this work we focus on teams with two players. This reduces the computational burden but still allows us to study the problem of movement coordination.

\paragraph{Environment Dynamics} At the start of an episode, the positions and orientations of four humanoid players as well as the ball are uniformly initialized across a central portion of the football pitch. The radius of the ball as well as the goal sizes follow the \textit{FIFA} regulation sizes adjusted in proportion to the humanoid body height.\footnote{We use FIFA 5-vs-5 regulation ball radius of 11cm, goal length of 3.66m scaled by the ratio of simulated humanoid body height of 1.5m to the average human body height of 1.75m.} 
The pitch size is sampled within a range at the start of each episode, scaled proportionally to the number of players.\footnote{For each episode, we randomly sample a per-player area between 100sqm and 350sqm, or 40\% of the 5-vs-5 \textit{FIFA} regulation pitch sizes.} To emulate the football rules, the players can travel outside of the boundaries of the pitch (but cannot travel outside of the gradient-coloured physical hoardings), whereas the ball ``bounces off'' of the pitch boundary. This simplification removes the need for a throw-in mechanism, and leaves the physics simulation to determine the range of strategies that players can execute (including deliberately bouncing the ball off the pitch boundary). Within an episode, if either team scores, the game resets with the same initialization logic as executed at the start of an episode and continues to the next timestep. 
Episodes can consist of multiple scoring events and training matches last 45 seconds.

\paragraph{Observation and Actuation} Each agent observes their own physical state through a set of proprioceptive measurements: joint angles, joint velocities, root orientation with respect to the world vertical axes, as well as sensory readings including accelerometer, velocimeter, and gyroscope. Exteroceptively, an agent observes other players and physical objects in the scene such as the ball and goal posts via a narrower set of physical observations: position, velocities, and orientation projected onto their own egocentric coordinate frame.\footnote{Observations are such that the position of the ball, goal posts and pitch boundaries can be precisely determined, but other players are only partially observed via positions of hands, feet and pelvis.} Every 30ms of simulation time, each agent perceives the environment's current physical state, partially, via the state observations, and samples a 56 dimensional, bounded, continuous action, corresponding to desired joint positions. The desired joint positions  are then converted to torques at the 56 joints using proportional-position actuators with realistic gains and maximum-torque values.

\section{Learning Framework} \label{sec:Method}

\begin{figure}
  \centering
  \includegraphics[width=\textwidth]{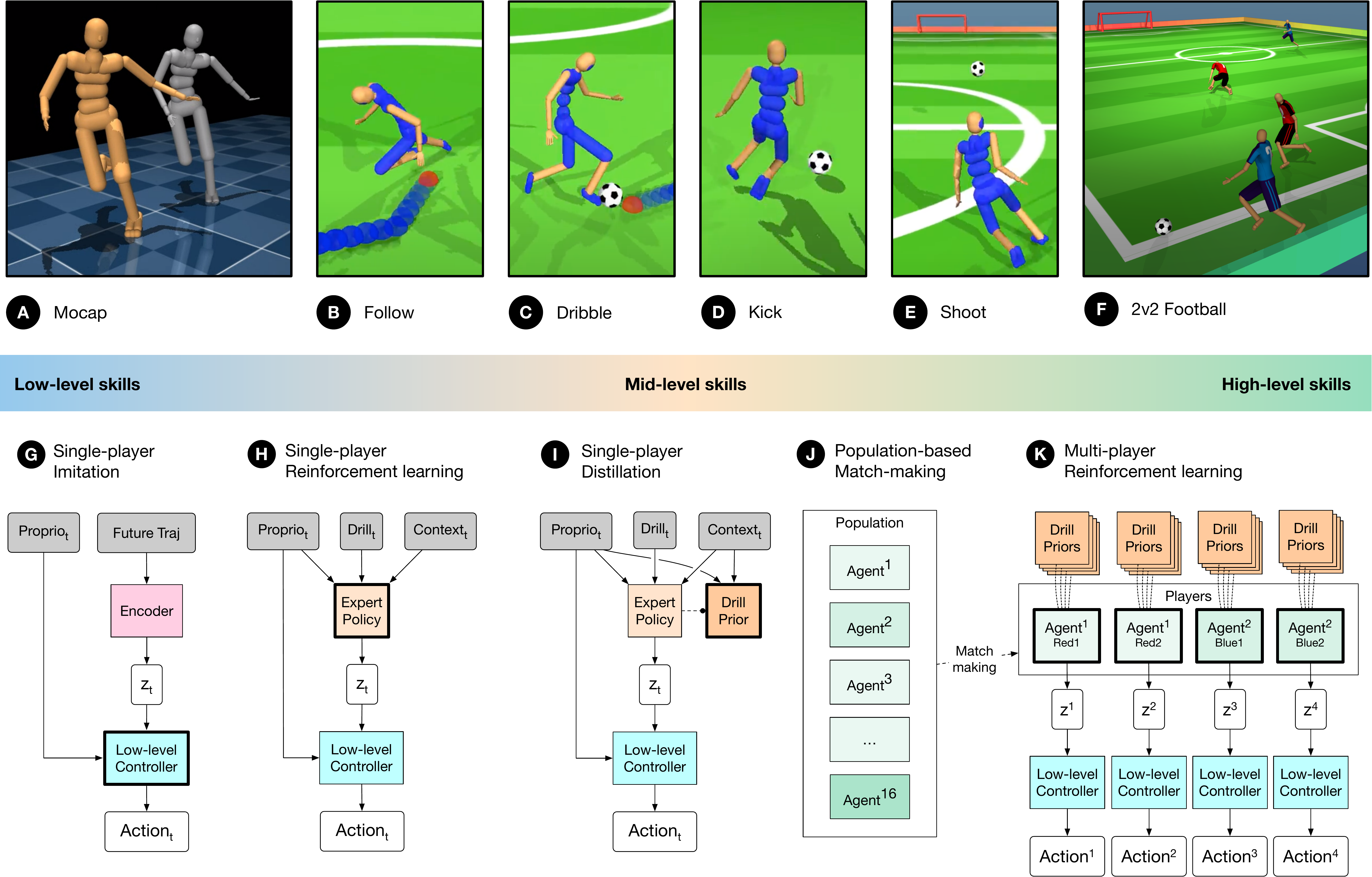}
  \caption{An overview of the proposed learning framework. Components shown in bold are optimized during their corresponding stage and transferred to the subsequent stages. {\bf (A, G)} an example frame of the motion capture behaviour (grey) and the low-level controller optimized to reproduce matching behaviours (beige). {\bf (B-E, H-I)} illustrations of the four mid-level drill tasks and their corresponding training procedures to obtain per-task expert policy and subsequently per-task, uninformed drill prior. {\bf (F, J-K)} an example frame of a 2v2 football match with the two teams sampled from the population of agents. The agents are optimized end-to-end by RL, reusing the low-level controller while regularized towards pre-trained drill priors.}
  \label{fig:learning}
\end{figure}

\begin{table}[ht]
    \centering
    \begin{tabular}{llllll}
        \toprule
        \textbf{Stage} & \textbf{Methods} & \textbf{Domain} & \textbf{Output} & \textbf{Details}\\
        \toprule
        \multirow{2}{*}{Low-level} & 1.\ Imitation via RL & Reference tracking & Per-clip expert & \multirow{2}{*}{Sec.~\ref{sec:Method:LowLevel}}\\
        & 2.\ Distillation & Supervised learning & General NPMP & \\
        \midrule
        \multirow{2}{*}{Mid-level} & 3.\ RL and PBT & Football drills & Per-drill expert & \multirow{2}{*}{Sec.~\ref{sec:Method:Drills}}\\
        & 4.\ Distillation & Football drills & Per-drill prior & \\
        \midrule
        High-level & 5.\ RL and PBT & 2v2 football & Full agent & Sec.~\ref{sec:Methods:MARL}\\
        \bottomrule
    \end{tabular}
    \caption{A summary of the methods and domains used at each stage of training.}
\end{table}

In this section we describe our three-stage learning framework, during which agents acquire increasingly complex competencies. In Section \ref{sec:Method:LowLevel} we describe the acquisition of low-level movement skills via imitation learning from human motion capture data. The result is a general-purpose \emph{motor module} that controls the players' movements and translates a \textit{motor intention} into joint actuation that leads to realistic human-like movements.
We use this motor module as part of the agents that we train in the second and third stage. In the second stage (Section \ref{sec:Method:Drills}) we train agents to solve a suite of football-specific training drills. The resulting behaviours are compressed into reusable skill policies, or \emph{drill priors} that capture general locomotion and ball-handling skills. Finally, in the third stage (Section~\ref{sec:Methods:MARL}), agents acquire long-horizon coordinated football play. This is achieved by training them in the full game of football while guiding exploration with mid-level drill priors. \maybedelete{The final stage takes advantage of explicit representations of low- and mid-level skills to speed up training and avoid local optima but circumvents common issues where such representations restrict the final solution in undesirable ways.}

An overview of the three stages of the learning procedure is provided in Figure~\ref{fig:learning}. In stage 2 and 3 we train populations of players by reinforcement learning using a hybrid, two-timescale optimization scheme similar to that employed in \cite{JaderbergCTF}. We describe the general problem definition for this training setup, the agent architecture, and the meta-optimization scheme in Section \ref{sec:Methods:Preliminaries}.

\subsection{General Training Setup}
\label{sec:Methods:Preliminaries}

We model each of the $K$ tasks (the football game and the training drills) as a multi-agent reinforcement learning (MARL) problem using the framework of $n$-player \textit{Stochastic Games} \cite{shapley1953stochastic} (see Appendix~\ref{sec:Appendix:Training:Abstraction} for more detail). For the football task $n=4$, but our training drills are single-agent tasks with $n=1$. In game $\cG_k$, at each timestep $t$, each player $i \in \{1,\ldots, n\}$ observes $o_t^i = \phi^k_i(s_t)$, which are features extracted from the game state $s_t$, by the players observation function $\phi^k_i$, as described in Section~\ref{sec:Environment}. Each player $i$ then independently selects a 56-dimensional continuous action $a^i_t\in\cA$ to control the humanoid (see Section~\ref{sec:Environment}). Actions are sampled from the player's (stochastic) \emph{policy}, $a^i_t \sim \pi^i(\cdot| h_t^i)$, as a function of their observation-action history $h^i_t := (o_0^i, a_0^i, o_1^i, \dots, o_t^i)$. These interactions give rise to a trajectory $\traj = \left((s_t, a_t^1, \ldots , a_t^n, r_{t,1}, \ldots , r_{t,n})\right)_{t\in[T]}$, over a horizon $T$, where the game state transitions according to the system dynamics $s_{t+1} \sim P^k(\cdot|s_{t}, a^1_{t}, \ldots, a^n_{t})$, and the player receives reward as a function of state $r_{t,i} = r^k_i(s_t)$. Action sets are consistent across tasks, and observation and dynamics are partially consistent, which enables skill transfer across the family of tasks.

The objective for a policy $\pi$ in task $\cG_k$ is to maximize expected cumulative reward,
\begin{align}
    \label{eqn:Methods:MDPObjective}
    \cF^k(\pi):=\mathbb{E} \Bigl[ \sum^T_{t=1} r^k_{i}(s_t) \Bigr],
\end{align}
where expectation is over the system dynamics, the sampling of player index $i$ to be controlled by policy $\pi$, and coplayer policies $\pi^{\noti}:=(\pi^1, \ldots,\pi^{i-1}, \pi^{i+1}, \ldots, \pi^n )$, and the sampling of actions from the appropriate player policies $\{ \pi^j : j\in[n]\}$.\footnote{In drill tasks $n=1$ and, in the football task, players are not assigned roles within a team, so that rewards are invariant to permutations of players which preserve teams. Our agents do not learn separate policies for each player index, or observe their player index.}
In the football task, denoted by $\cG_0$, players receive a reward of +1 (-1) on the terminal state $s_T$ when their team wins (loses) the match, or 0 if the match ends in a tie.

\subsubsection{Outer-Loop Optimization with Population-Based Training}

\label{sec:Methods:MetaO10n}
For task $\cG_k$, we train a population of reinforcement learning agents, $\cW = \{w_j : j\in[|\cW|]\}$, which learn the task, as described in Section \ref{sec:Methods:InnerO10n}. In practice, each agent is a collection, $w_j = (\theta_j, \theta_j^Q, \theta_j^h)$, of policy network parameters $\theta_j$, network parameters of an auxiliary action-value function $\theta_j^Q$, and hyper-parameters of the learning process $\theta_j^h$. In the multi-agent football task the population of agents play against each other as opponents but, otherwise, the agents are deployed independently in the single-agent drills.

For several tasks \autoref{eqn:Methods:MDPObjective} is difficult to optimize with RL. This is particularly true for the football task where the reward is sparse and has high variance. The reward also provides no direct information on how to play football well, or which behaviours may be useful, resulting in a challenging long-horizon exploration problem. We therefore define parameterized \emph{surrogate objectives} $\{\cJ^k : k\in[K] \}$ which are easier to optimize by reinforcement learning. The surrogate objective for agent $w_j$ is denoted by $\cJ^k(\theta_j; \theta^h_j)$, with $\theta^h_j$ including parameters of the surrogate objective. In the inner loop,  $\cJ^k$ is optimized with respect to $\theta_j$ via RL as described in Section~\ref{sec:Methods:InnerO10n}.

For each task, we use population-based training (PBT, \cite{jaderberg2017population}) as an optimizer over the population $\cW$ to perform several functions:
\begin{itemize}
    \item Optimize the hyper-parameters $\theta^h_j$ of the surrogate objective $\cJ^k$ to provide a well-shaped learning signal for individual learning agents such that by optimizing $\cJ^k$, with RL, the agents effectively optimize the original objective of the task (\autoref{eqn:Methods:MDPObjective}). 
    \item Optimize other hyper-parameters that control the learning dynamics, such as learning rates.
    \item Implement a mechanism that increases the proportion of high-performing agents in the population, resulting in an automatic curriculum over the strength of coplayers in the football task.
\end{itemize}

In Algorithm~\ref{alg:pbt} we specify the implementation of the outer loop in terms of several subroutines: {\it Eligible} controls the frequency of evolution events based on the number of environment interactions that took place since the last evolution. {\it Select} samples a pair of agents for evolution, where the child agent corresponds to the agent with the minimum fitness and the parent selected uniformly at random. {\it Mutate} and {\it Crossover} define the exploration strategy for the inherited hyper-parameters $\theta^h_j$. {\it UpdateFitness} implements the rules for population fitness updates, based on the resulting episodic returns following interactions between population members. Concrete implementations of {\it UpdateFitness} depend on the task, and specific hyper-parameters and are provided in Section~\ref{sec:Method:DrillPBTRL} and Appendix~\ref{app:FitnessMeasure}.

\begin{algorithm}[t]
  \begin{small}
  \caption{\small{Meta-Optimization with Population-based Training.}}\label{alg:pbt}
  \begin{algorithmic}[1]
    \Procedure{PBT}{}
    \State Let $\{w_j\}$ denote independent agents forming a population $\cW$, optimizing for a task $k$.
    \For{agent $w_j$ in $\cW$}
        \State Initialize agent network parameters $\theta_j$ and agent fitness $f_j$ to fixed initial fitness $F_{init}$.
        \State{Sample initial hyper-parameters $\theta^h_j$ from the initial hyper-parameter distribution.}
    \EndFor
    
    \While {true}
        \State Select agents from $\{w_j\}$ to play in {\it Episodes}. Submit data to replay buffers $\{\cR_j\}$. \Comment{Matchmaking}
        \For{agent $j$ with opponent $\notj$, states $(s_0, \dots, s_T) \in \mathit{Episodes}$}
        \State $\mathit{UpdateFitness}(j, \notj, \sum_{t=1}^T r^k(s_t))$  \Comment{Fitness Update}
        \EndFor
        \State For each $w_j$ optimize $\cJ^k(\theta_j; \theta^h_j)$ using data from replay $\cR_j$. \Comment{Inner-loop Optimization}
        \If{$\mathit{Eligible}(\cW)$}  \Comment{Population Eligibility Criteria}
            \State $j, \ell \gets \mathit{Select}(\cW)$  \Comment{Parent-Child selection}
            \State $(\theta_j, \theta^Q_j) \gets (\theta_\ell, \theta^Q_\ell)$  \Comment{Inherit Parameters}
            \State $\theta_j^h = \mathit{Mutate}(\textit{Crossover}(\theta^h_j, \theta^h_\ell))$  \Comment{Mutate-Inherit Hyper-parameters}
        \EndIf
    \EndWhile
    \EndProcedure
  \end{algorithmic}
  \end{small}
\end{algorithm}

\subsubsection{Inner-Loop Optimization with Reinforcement Learning}

\label{sec:Methods:InnerO10n}

For each task $\cG_k$ we introduce a set of shaping rewards $\{ \hat{r}^{(k,\ell)} : \ell \in [M_k] \}$ each associated with a discount factor $\gamma_{k,\ell}$ and a coefficient $\alpha_{k,\ell}$, which enable adjusting the relative importance and horizon of each reward component individually. The surrogate objective for an agent with policy parameter $\theta$ and hyper-parameters $\theta^h$, is a discounted infinite horizon sum of the form
\begin{equation}
    \cJ^k(\theta; \theta^h) := \mathbb{E}\Bigl[ \sum_{\ell=1}^{M_k} \alpha_{k,\ell} \sum^\infty_{t=0} \gamma_{k,\ell}^t \hat{r}_{i}^{(k,\ell)}(s_{\tau + t}) \Bigr], 
    \label{eq:SurrogateObjectiveGeneral}
\end{equation}
where expectation is over the system dynamics, sampling of actions from player policies, and, in the multi-agent football task, the assignment of policy $\pi_\theta$ to control player $i$ and sampling of coplayers $\pi^{\noti}$.\footnote{The distribution over $s_\tau$ in \autoref{eq:SurrogateObjectiveGeneral} is the state visitation distribution of the policy, determined in practice by the matchmaking scheme and the sampling of data from replay buffers which store episode data for each agent.}
Given fixed hyper-parameters $\myvec{\alpha}, \myvec{\gamma} \in \theta^h $, determined by the outer-loop optimization, the network parameters $\theta$, are optimized by RL using Maximum a Posteriori Policy Optimization \cite{AbdolmalekiMPO} as described in Appendix~\ref{sec:Appendix:Training:MPO}. The precise reward functions optimized by $\cJ^k$ depend on the specific task and are described in Sections~\ref{sec:Method:Drills} and \ref{sec:Methods:MARL}.

\subsubsection{Agent Architecture}
\label{sec:MethodsPreliminaries:Architecture}

The agent first processes proprioceptive and task-specific observations using feature encoders parameterized using multi-layer perceptrons (MLP). An order-invariant attention module is used to further process observations of coplayers. Since optimal policies in multi-agent environments are, in general, a function of the interaction history, LSTM modules \cite{HochreiterLSTM} then process history in the policy and action value functions. See Appendix~\ref{app:AgentArchitecture} for more details.

\subsection{Stage 1: Learning Low Level Motor Control Using Human Data}
\label{sec:Method:LowLevel}

To play football well, dynamic movements are required. To bias our agent's behaviour towards realistic and useful movements, we learn a motor primitive module by imitation of football motion capture data. We used roughly 1 hour and 45 minutes of football motion capture data collected from ``vignettes'' of semi-natural scripted scenes of football gameplay. Although the data contains ball interactions, the ball was not part of the tracked data. We registered all of the point-cloud data onto the humanoid model.  For more details see Appendix \ref{sec:Appendix:NPMP}. 

To build the low-level controller (see Figure~\ref{fig:learning}G) we used a two-stage pipeline consisting of tracking motion capture clips with individual policies, followed by distillation of the tracked behaviours into a single low-level controller. This approach follows previous work \cite{MerelNPMP, merel2020}; the architecture is referred to as a neural probabilistic motor primitive (NPMP) model. In particular, these previous works have demonstrated that distilling trajectories into an inverse model with a latent bottleneck that is trained to reconstruct the action as a function of the current state and the future-state trajectory produces a reusable motor controller.  
To achieve this, we first cut the motion capture data into 4-8s snippets and trained separate time-indexed ``tracking'' policies by reinforcement learning to imitate each snippet. The reward function for tracking was the same as that used in \cite{Hasenclever2020}.

In the second step, we sampled multiple trajectories from each tracking policy, with noise added to the actions to induce variations and witness the corrective behaviour of the tracking policy; we then used a supervised training approach to distil these sampled trajectories into a single neural network controller.
More specifically, each training trajectory was obtained from a tracking policy by starting the episode at a random time within the corresponding reference motion snippet and performing a rollout until the end of the reference clip, acting according to the tracking policy in the presence of action noise.  Given a set of $\cT$ trajectories $\{\small((x^{(i)}_t, a^{(i)}_t)\small)_{t\in\{0,\ldots,T\}}\}_{i\in [\cT]}$, consisting of proprioceptive state features $x_t \in \mathcal{X}$ and noiseless actions $a_t \in \mathcal{A}$ from the tracking policies, we can train the motor module, $\npmp$, according to the supervised objective
\begin{align}
    \mathbb{E}_q
    \Bigl[
    \sum_{t=1}^{T}
    \underbrace{\log \npmp(a_t|x_t, z_t)}_{decoder~/~LL~controller} + \beta \big( \underbrace{\log p_{z}(z_t|z_{t-1})}_{latent~prior} - \underbrace{\log q(z_t|z_{t-1}, x_{t+1:t+k})}_{encoder}\big)
    \Bigr], \label{eqn:ELBO}
\end{align}
where $z_t$ is a latent variable that represents the future trajectory and the distribution 
$q(z_t|z_{t-1}, x_{t+1:t+k})$, which is optimized, corresponds to an encoder that produces these latents, given short look-aheads into the future. 
As a result of the training on expert motor behaviour, and specifically by encoding the future lookaheads, the latent variable $z_t$ can be interpreted as a {\it motor intention}, because the latent variable determines what behaviour the low-level controller will generate for a short horizon into the future. See Appendix \ref{sec:Appendix:NPMP} for specific network details.  The relevant part of the model that is subsequently used as a low-level controller is the decoder $\npmp(a_t|x_t, z_t)$, which produces actions in response to both the current state and a latent variable. 

The above training procedure yields a plug-and-play low-level controller, used, without further training, in the remainder of the present work. For football and drill training, instead of producing behaviours by operating at the level of raw per-joint actions, football agents are trained to produce 60-dimension continuous \emph{latent motor intentions}, $z_t$, that are fed into the fixed low-level policy, together with the proprioception input $x\in\cX$. Reinforcement learning is performed in the latent motor-intention space. This approach effectively reconfigures the control space so that random exploration in the latent motor intention space is more likely to produce realistically correlated actions and useful humanoid motion.

\subsection{Stage 2: Acquiring Transferable Mid Level Skills}
\label{sec:Method:Drills}

By design, the motor module from Section \ref{sec:Method:LowLevel} constrains the space of low level movements but does not directly produce temporally extended movement patterns. The motor module does not directly encourage interaction with the ball or any other goal directed behaviour relevant for football. To learn \emph{mid-level skills} including running, turning and balance, as well as football-specific dribbling and kicking skills, we train players to learn a syllabus of training drills, listed in Table \ref{tab:TrainingDrills}. These prelearned skills are then used to accelerate learning of the full football task.

\begin{table}[ht]
    \centering
    \begin{tabular}{lp{12cm}}
        \toprule
         {\bf Drill name} & {\bf Description } \\ 
         \toprule
         \emph{Follow} & The agent must follow a moving target that moves at fixed velocity for a short episode and in variable directions. The target velocity is randomized at the start of the episode. The agent observes the current target and the future position of the target, so that the agent can anticipate where the target will move and prepare accordingly.
         \\
         \midrule
         \emph{Dribble} & The environment is similar to the \emph{follow} drill but the agent must keep the ball close to the moving target. \\
         \midrule
         \emph{Shoot} & The ball is initialized randomly on the pitch and the agent has a budget of three ball contacts with which to score a goal.\footnote{This does not translate to precisely three kicks since a kick will typically involve contact over two or three consecutive timesteps.} \\
         \midrule
         \emph{Kick-to-target} & The agent has a small window of time (randomized between two and six seconds) in which to manoeuvre the ball and kick it to a distant fixed target. \\
        \bottomrule
    \end{tabular}
    \caption{Training drills employed in Stage 2 of the training framework to induce mid-level football skills. See Appendix~\ref{sec:Appendix:Drills} for more details.}
    \label{tab:TrainingDrills}
\end{table}

\subsubsection{Learning Task-Specific Expert Drill Policies by PBT-RL}
\label{sec:Method:DrillPBTRL}

For each drill $\cG_k$ we train a population of task-specific {\it expert policies} (see Figure~\ref{fig:learning}H). We use the general setup of Section~\ref{sec:Methods:Preliminaries}. We reuse the low-level motor primitives developed in Section~\ref{sec:Method:LowLevel}: drill experts output motor intentions to the fixed NPMP module, and RL is effectively performed in the latent motor intention space.

\paragraph{Fitness Measure and Expert Selection} The drill objectives are defined in terms of reward functions that characterize the desired behaviour, yielding a fitness $\cF^k(\pi) = \mathbb{E}\left[\sum^T_{t=1} r^k(s_t)\right]$, recalling Section \ref{sec:Methods:MetaO10n} (\autoref{eqn:Methods:MDPObjective}). Specifically, we choose the reward function for {\it shoot} to be the binary indicator function of the ball reaching the goal; and for {\it dribble} we choose a measure of closeness between ball and the moving target. A detailed description of the reward functions for drill tasks can be found in Appendix~\ref{app:ShapingRewards}. For each drill we select the policy with maximal fitness from the population, yielding a collection of expert policies $\{\expert^k : k\in[4] \}$, one for each drill.

\paragraph{Surrogate Objective and Shaping Rewards} To optimize policies for the drill tasks via RL we introduce several shaping rewards for each drill and maximize surrogate objectives of the form of \autoref{eq:SurrogateObjectiveGeneral}. For instance, the {\it kick-to-target} drill introduces a reward for maximizing {\it ball-to-target} velocity, while the {\it shoot} drill rewards positive {\it player-to-ball} velocity to encourage ball interaction early in the training. We provide further details in Appendix~\ref{app:ShapingRewards}.

\subsubsection{Distilling Expert Drill Policies into Transferable Behaviour Priors}

Each drill utilizes task-specific context which the expert policies observe. For instance, in the \emph{follow} and \emph{dribble} drills the agent is required to follow a virtual target. To obtain {\it transferable skill representations} that are target-agnostic and can be reused in the football task we \emph{distil} the expert policies into \emph{drill priors} $\{\prior^k : k\in[4] \}$ (see Figure~\ref{fig:learning}I). The priors are trained to mimic the drill expert policies $\{\expert^k : k\in[4] \}$ but only observe features that are available in the football task. These include proprioception, and in the case of {\it dribble}, {\it shoot} and {\it kick-to-target}, the ball. In the case of {\it shoot} we exclude the goal position from the drill prior observations so that the prior learns a general kicking policy. See Appendix~\ref{sec:Appendix:DrillObservations} for details.

The priors are trained by minimizing the KL-divergence with the expert in the latent motor intention space:
\begin{align}
\bE\left[\sum_{t=1}^T D_{KL}( \expert^k(\cdot|h_t)  || \prior^k(\cdot|\tilde h_t) )\right],
\label{eq:Drill:Distillation}
\end{align}
where expectation is over the distribution over trajectories, sampled from a replay buffer, encountered when acting with the expert policy $\expert^k$ in a sampled instantiation of task $k$; $h_t = (o_1, a_1, o_2,\ldots,o_t)$ is the observation-action history at time $t$ from the perspective of the drill expert $\expert^k$; and $\tilde h_t$ is the observation-action history in terms of the prior's reduced observation set. The resulting drill priors reproduce the behaviour of the experts (dribbling the ball, turning and speeding-up and slowing down, for example) but do so without being prompted by a specific target. For instance, the dribble prior will favour behaviour patterns that involve dribbling the ball, independently of the direction and speed.\footnote{This is a consequence of the loss in \autoref{eq:Drill:Distillation} which trains the drill priors to match the mixture distribution that arises from executing the drill expert for different target choices \cite[see ][for details]{galashov2018information,tirumala2020behavior}.}

\subsection{Stage 3: Achieving Long-Horizon Coordinated Football Play}
\label{sec:Methods:MARL}

In the final stage of training, players learn the full football task using the general two-timescale optimization setup of Section~\ref{sec:Methods:Preliminaries} to train a population $\cW$ of football players (see Figure~\ref{fig:learning}J-K). We describe the fitness measure which drives PBT in the outer-loop and then describe the inner-loop optimization using Multi-Agent Reinforcement Learning. We make use of behaviour shaping, using the low- and mid-level skills acquired in the first two stages as well as additional shaping rewards.

\subsubsection{Outer-Loop Optimization for Football}
\label{sec:Method:PopulationBasedTraining}

We use the outer-loop optimization procedure described in Section~\ref{sec:Methods:MetaO10n}. In contrast to Stage 2 (cf. Section~\ref{sec:Method:Drills}), agents play against each other in multi-agent games. Competitive play within a population, combined with a mechanism to propagate high-performing policies through the population, induces an autocurriculum in which environment difficulty (determined by the strength of opponents in the population) is effectively calibrated to a practical but challenging level to learn from \cite{LeiboManifesto}. Next, we specify the matchmaking and fitness measure used.

\paragraph{Matchmaking} We concurrently optimize the population $\cW$ of agents, individually learning from their first-person experience, in football matches, in a decentralized fashion.  To form a match, we sample a pair of agents uniformly with replacement from the population $\cW$. At the start of an episode, we make two separate instantiations of each agent's policy (clones) which are paired to form a team of two players, and the two teams then compete. During execution, agents act independently, without access to other agents' actions, observations or other privileged information -- agents must observe their opponents and also their teammate from a third-person perspective while deciding on their optimal course of actions. 

\paragraph{Fitness Measure for Football} In the multi-agent football task we use \textit{Nash Averaging} \cite{balduzzi2018re} as the fitness measure to drive our PBT mechanism. This measures the performance of each agent against the distribution of agents in the Nash equilibrium, the definition and implementation details of which are provided in Appendix~\ref{app:FitnessMeasure}.

\subsubsection{Inner-Loop Optimiztion for Football with MARL}

\paragraph{Behaviour Shaping with Mid-level Skills} 

To assist exploration and the discovery of mid-level behaviours useful for soccer, we bias the behaviour of the players towards the mid-level skills described in Section~\ref{sec:Method:Drills}. This leads to sparse rewards being encountered sooner and can help avoid poor local optima in locomotion and ball handling behaviour. We define a loss $\mathcal{L}_{\rm priors}$, to be used as a regularizer, that penalizes the KL-divergence, in the latent motor intention space, between the players' football policy and a mixture distribution constructed from the four prelearned transferable drill priors. The loss for agent $w_j \in \cW$, with policy $\pi_{\theta_j}$, is defined as:
\begin{align}
    \label{eqn:behaviourPriorKLLoss}
    \textstyle
    \mathcal{L}_{\rm priors}(\theta_j; \beta_{1:4}):=\bE[ \sum_{t=1}^T D_{KL}(\pi_{\theta_j}(\cdot|h_t)||\sum_{i=1}^4 \beta_i \prior^{i}(\cdot|\tilde h^i_t))]
\end{align}
where expectation is over the sampling of trajectories from a replay buffer (Section~\ref{sec:Experiments:Infrastructure} for more details) of recent football training matches involving agent $j$; $h_t$ and $\tilde h^i_t$ denote the observation-action history at time $t$ from the perspective of the football policy $\pi_{\theta_j}$ and prior $\mu^i$, respectively; the mixture weights $\{\beta_i\}$ ( $\sum_{i=1}^4 \beta_i = 1$ and $\beta_i\ge 0$) are hyper-parameters of the objective that control the relative importance of the different drill priors.

Regularizing towards a mixture of drill priors accounts for the fact that the different priors characterize different types of behaviours and that the player will have to switch between different behavioural modes depending on the game state. The KL-divergence $D_{KL}$ to a mixture is bounded from above by the divergence to any individual mixture component up to a constant defined in terms of the mixture weights (see Appendix \ref{app:MixtureKL}). Thus, if a player's behaviour is close to one of the drill priors it will not pay a large cost for deviating from the remaining ones.

The use of drill priors bears some similarity to the use of shaping rewards. Yet, defining appropriate shaping rewards for complex behaviours such as dribbling or kicking that integrate well with the overall objective can be difficult. This probabilistic formulation in terms of drill priors provides extra flexibility since it allows context-dependent prioritization and de-prioritization of individual shaping terms, an effect which does not naturally emerge with the naive use of shaping rewards. We will discuss specific examples of this in our analyses of agent behaviour in Section~\ref{sec:HowAgentsWork}.

This setting resembles and is inspired by the Distral framework \cite{teh2017distral, tirumala2020behavior}, but rather than simultaneously co-learn all tasks, we wish to reuse the skills learned during the drills in the more challenging football environment. Hence we pre-train the simpler drill priors and, once learned, fix the prior policies \cite[as in][]{galashov2018information}. 

\paragraph{Behaviour Shaping with Shaping Rewards} 

We use a surrogate reinforcement learning objective for football, $\cJ^0(\theta; \theta^h)$, as described in Section \ref{sec:Methods:InnerO10n} \autoref{eq:SurrogateObjectiveGeneral}, specialized using shaping rewards for football. These include sparse rewards for scoring or conceding a goal, as well as dense shaping rewards for maximizing the magnitude of \textit{player-to-ball} and \textit{ball-to-goal} velocities. These dense rewards are intentionally myopic and thus relatively easy to optimize but do not encourage coordination directly. We provide detailed descriptions of the shaping rewards in Table~\ref{table:ShapingRewards} in Appendix~\ref{app:ShapingRewards}.

\paragraph{Parameterized MARL Objectives} 

Rather than optimize $\cJ^0(\theta; \theta^h)$ directly to learn football, we additionally regularize football behaviour towards drill priors, using the loss $\mathcal{L}_{\rm priors}$ \autoref{eqn:behaviourPriorKLLoss}, and optimize the regularized reinforcement learning objective,
\begin{align}
     \bar\cJ^0(\theta; \theta^h):=\cJ^0(\theta; \theta^h) - \lambda \mathcal{L}_{\rm priors}(\theta; \beta_{1:4}). \label{eq:objective}
\end{align}
where and $\theta^h$ denotes the set of hyper-parameters optimized in the outer loop, including
$$(\alpha_1, \ldots, \alpha_M, \gamma_1, \ldots, \gamma_M, \beta_1, \ldots, \beta_4, \lambda),$$
and hyper-parameters of the learning process. As in stage 2, we restrict the behaviour of the football players to human-like movements, using the low-level motor module derived from human motion capture (see Section \ref{sec:Method:LowLevel}). The football policies are trained to produce control outputs in terms of the latent {\it motor intention} space defined by the low-level controller (see Section \ref{sec:MethodsPreliminaries:Architecture}).

\section{Experiments}
\label{sec:Experiments}

We study our learning framework with a series of experiments that highlight the capabilities that players can acquire and analyze the contributions of different components of the framework. In this section we describe the experimental setup as well as the evaluation framework. %

\subsection{Experimental Setup}
\label{sec:Experiments:Setup}

For experiments with the full framework we first trained an NPMP as described in Section \ref{sec:Method:LowLevel}. 
We then trained populations of drill teachers and football players as described in Sections \ref{sec:Method:Drills} and \ref{sec:Methods:MARL}. To assess the reliability of the framework we performed three independent experiments. For each experiment we trained an independent population of 16 football players, with distinct random initializations of the weights and hyper-parameters. We used the same NPMP across all experiments, but trained a separate set of drill experts and priors for each experiment.\footnote{For each of the four drills we thus trained three separately initialized populations of drill experts.}

We trained each population of drill experts for $5\times 10^9$ environment steps (with the exception of the easiest {\it follow} drill which was trained for $2.5\times 10^9$ environment steps). We then selected the best expert from each population (in terms of maximal final fitness on each drill). The selected expert was then distilled into a prior as described in Section~\ref{sec:Method:Drills}. We distilled each expert using four separate seeds to randomize initial optimization hyper-parameters, and selected the prior achieving the lowest distillation loss (i.e., KL-divergence to the expert) after $10^6$ gradient steps.
Once the drill priors were trained we proceeded with training of the populations of the football players as described in Section~\ref{sec:Methods:MARL}.

\begin{figure}
  \centering
  \includegraphics[width=0.85\textwidth]{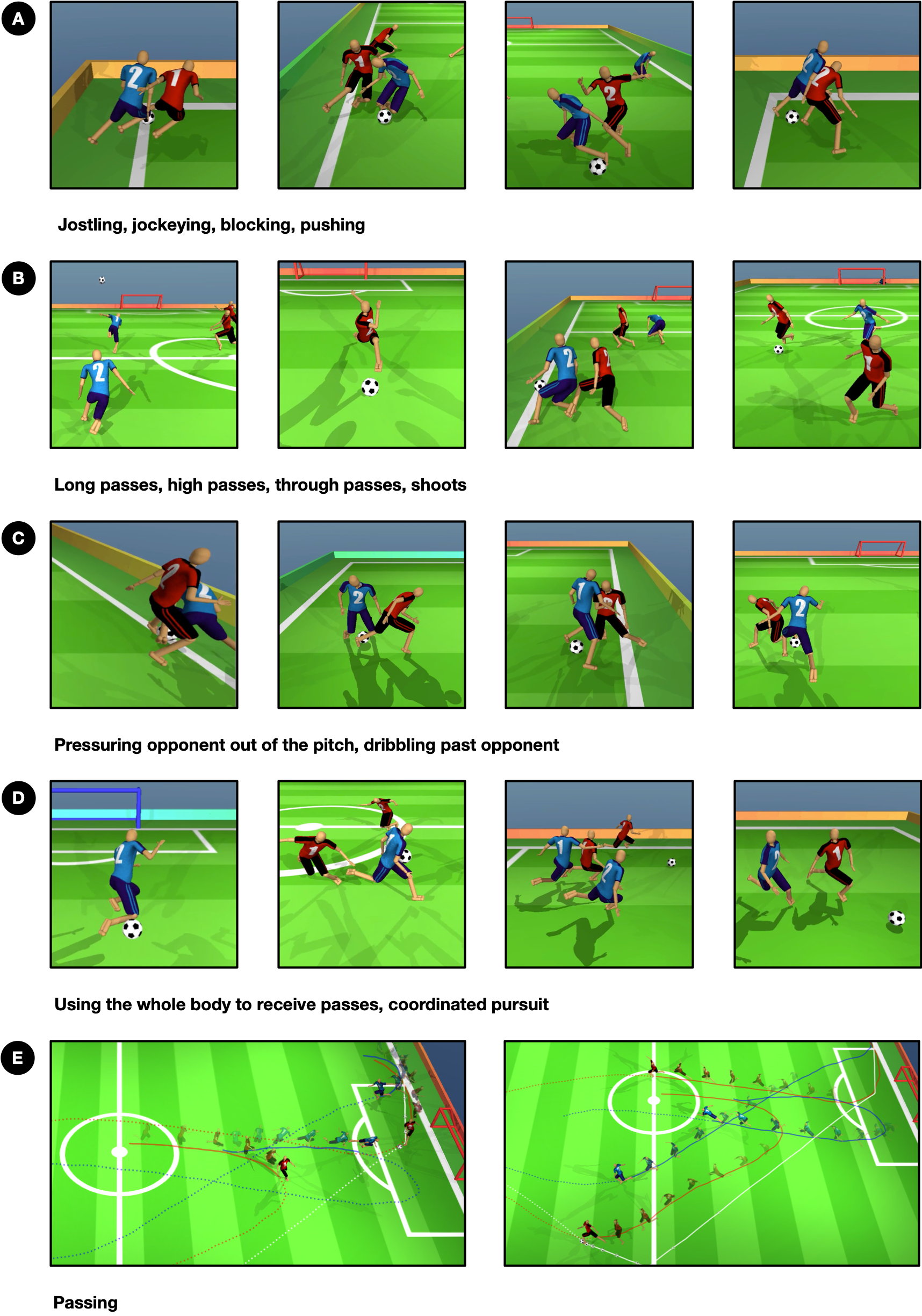}
  \caption{A selection of the learned behaviours of the agent. Through realistic physical simulation, agents acquired and optimized skills through agent-agent, agent-ball interactions in a range of scenarios.}
  \label{fig:behaviour_gallery}
\end{figure}

We trained the football players for $8 \times 10^{10}$ environment steps, corresponding to six weeks of training in wallclock time. The evolution of the performance of our full agent is shown in Figure~\ref{fig:ablation}. After about two weeks of training our best agent decisively beats all evaluation agents (defined below) and improvement begins to slow down although performance does not saturate. Over the course of training agents acquire context-dependent movement and ball-handling skills such as getting up from the ground, fast locomotion, rapid changes of direction, or dribbling around opponents to make accurate shots. Players' locomotion becomes robust to external pushes and they engage in close quarter duels with opponents. Some scenes of gameplay are shown in Figure \ref{fig:behaviour_gallery}A-D. Players combine these movement skills with cooperative play. The players' behaviour progresses from ``individualistic'' ball-chasing to more coordinated team strategies involving division of labour, near-term tactics such as passing directed towards long-term team goals. Several of these motifs are repeated reliably across games. Some examples are shown in Figure~\ref{fig:behaviour_gallery}E, and behaviours can be seen in the \textcolor{blue}{\href{\videourl}{supplementary video}} and  \textcolor{blue}{\href{\episodesurl}{videos of full episodes}}.\footnote{{\href{\videourl}{\videourltext}} and {\href{\episodesurl}{\episodesurltext}}.} The final agent shown in the videos was trained for two generations, each of $4 \times 10^{10}$ environment steps. The second generation agent was trained using regularization towards the best 1st generation agent, as well as the four drill expert priors.

\subsection{Multi-Agent Evaluation}
\label{sec:Experiments:Evaluation}

Evaluation in multi-agent domains can be a challenge since the objective is implicitly defined in terms of the (distribution of) other agents and the optimal behaviour may thus vary significantly. This is true, in particular, in non-transitive domains, where no single dominant strategy exists \cite{Czarnecki2020Spinning}. In the case of some computer games, human performance can be used to establish baselines \citep{MnihDQN,JaderbergCTF,OpenAI_dota,SilverAlphaGo, VinyalsStarCraft}, but the nature of the control problem and the lack of a natural interface for controlling the high-dimensional humanoid players renders this unfeasible in our case.

During training we require a meaningful signal of training progress.We create a set of 13 evaluation agents that play at different skill levels but also exhibit different behaviour traits. These evaluation agents differentiate between players in the training population over the course of training and highlight the differences in their behaviours. The set of evaluation agents are ``held-out'' from the training process: training agents do not optimize their performance against evaluation agents. We provide detailed statistics of the evaluators in Appendix~\ref{app:Evaluators}. For consistency, the same set of evaluators are also used as opponents for behaviour analysis in Section~\ref{sec:behaviour}.

Performance of each agent in the population in the full game of football was measured by playing 64 matches against the evaluation agents, at regular intervals during training, and computing Elo scores \cite{elo1978rating}. For each experiment, and for each measurement we select the top 3 players in the population, in terms of Elo against evaluation agents, and report that Elo score.

\subsection{Training Infrastructure}
\label{sec:Experiments:Infrastructure}
\begin{figure}[ht]
    \centering
    \includegraphics[width=\textwidth]{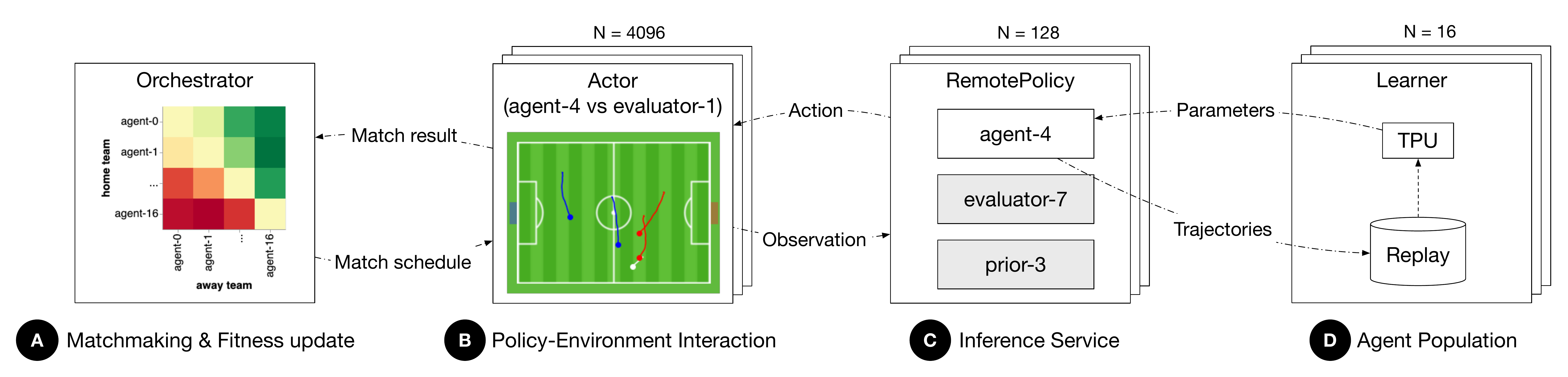}
    \caption{Distributed training infrastructure. {\bf(A)} A central orchestrator schedules agent-agent, agent-evaluator matches to be carried out by the actors. It receives match results from simulated matches and updates its payoff matrix which then informs the PBT optimization process as described in Algorithm~\ref{alg:pbt}. {\bf(B)} A large number of actors simulating matches upon receiving matchmaking schedule and connect to corresponding inference servers. Actors do not perform inference computation themselves. {\bf(C)} Inference servers receive inference requests concurrently from a large number of actors and perform inference computation for each model in batches. Depending on the inference request, the inference server may send experience trajectories to corresponding learners for learning. {\bf(D)} A set of learners host and continuously update the network parameters for the agent population, using off-policy data sampled from their respective replay buffers.}
    \label{fig:infrastructure}
\end{figure}

From a computational perspective our learning framework poses several challenges. Training and evaluation call for variable-throughput, low-latency inference over a large number of heterogeneous models. This includes high-throughput inference on the prior policies (the same prior policies are used for all players in all matches), medium-throughput inference on the training policies (the same agent participates in many matches concurrently) and low-throughput inference for the evaluation policies. 
Naively implementing the training infrastructure as described in prior work \cite{JaderbergCTF, VinyalsStarCraft} incurs significant setup costs since, for instance, the policy inference network needs to be re-created for all players for each match. This limitation has led to prior work fixing the pair of players over hundreds \cite{JaderbergCTF} or thousands of games \cite{VinyalsStarCraft} in order to amortize this setup cost.

A schematic of our infrastructure is provided in Figure~\ref{fig:infrastructure}. Learning is performed on a central 16-core TPU-v2 machine where one core is used for each player in the population.
Model inference occurs on 128 inference servers, each providing {\it inference-as-a-service} initiated by an inbound request identified by a unique model name. 
Concurrent requests for the same inference model result in automated batched inference, where an additional request incurs negligible marginal cost. 
Policy-environment interactions are executed on a large pool of 4,096 CPU actor workers. These connect to a central orchestrator machine which schedules the matches. Contrary to computation patterns commonly observed in the reinforcement learning literature, actors are light-weight workers that do not perform policy inference themselves. Inference is instead performed on the inference servers and experience data is sent directly to the relevant learner's replay buffer. Each learner samples data from its replay buffer to update its agent's policy and value functions (Section \ref{sec:Methods:InnerO10n}). We note that our proposed infrastructure offers greatly improved efficiency over previous systems \cite{JaderbergCTF, LiuSoccer} by leveraging efficient batched inference as well as improved flexibility compared to \cite{VinyalsStarCraft} by enabling per-episode re-sampling of players.

\section{How Football Agents Play} 
\label{sec:behaviour}

Gameplay of trained agents contains distinct and repeatable patterns, including agents' movements, their interactions with the ball and other players, as well as more strategic play and teamwork. We provide footage of the agents' behaviour in the \textcolor{blue}{\href{\videourl}{supplementary video}} for qualitative assessment of their skills.\footnote{{\href{\videourl}{\videourltext}}.} To understand the temporal evolution of different types of skills and behaviours, and to provide a quantitative picture of their occurrence, we perform three analysis techniques over the course of training: (a) we measure properties of the agents' behaviour during naturally occurring gameplay (\emph{behaviour statistics}); (b) we measure the sensitivity of agents to certain scene properties by selectively varying these properties in a controlled manner (\emph{counterfactual policy analysis}); and (c) we analyze the players' response to specific game situations under controlled conditions (\emph{probe tasks}).

\paragraph{Behavioural Statistics} We track statistics that measure the quality of football players' movement skills, ball handling skills, and team work during naturally occurring gameplay. Statistics were collected at regular intervals over the course of training during matches against the set of evaluation agents introduced in Section \ref{sec:Experiments:Evaluation}. We list the statistics in Table \ref{table:behaviourStatistics} and provide further details in Section \ref{sec:Appendix:Behaviors:Stats} of the appendix.

\begin{table}[h!]
    \centering
    \small
    \begin{tabular}{llp{11cm}l}
        \toprule
        \textbf{Type} & \textbf{Name} & \textbf{Description} \\ %
        \toprule
        \vspace{0.2cm}
        \parbox[t]{2mm}{\multirow{2}{*}[-0.7em]{\rotatebox[origin=c]{90}{Basic}}}
        & \emph{Speed} & Average absolute velocity of player. \\
        & \emph{Getting up} & Reliability of getting up: proportion of falls from which a player recovers before episode ends. \\ %
        \midrule
        \vspace{0.2cm}
        \parbox[t]{2mm}{\multirow{3}{*}[-0.7em]{\rotatebox[origin=c]{90}{Football}}}
        & \emph{Ball control} & Proportion of timesteps in which the closest player to the ball is a member of the team. \\ %
        \vspace{0.2cm}
        & \emph{Pass frequency} & Proportion of ball touches which are passes of range 5m or more.\footnote{A pass is defined as consecutive touches between teammates (not separated by a goal).} \\
        & \emph{Pass range} & Proportion of passes which are of range 10m or more. \\ %
        \midrule
        \vspace{0.2cm}
        \parbox[t]{2mm}{\multirow{3}{*}[-7.5em]{\rotatebox[origin=c]{90}{Team work}}}
        & 
        \emph{Division of labour} & Proportion of timesteps in which one but not multiple players in a team are within 2m the ball. Values near 1 indicate that players coordinate and do not all rush to the ball simultaneously; values close to 0 indicate the opposite. \\ %
        &\emph{Territory} & Proportion of points on the pitch to which the closest player is a member of the team. \\ %
        &\emph{Receiver OBSO} & \emph{Off-ball scoring opportunity} (OBSO) quantifies the quality of an attacking player's positioning and is used in sports analytics for the analysis of human football play \cite{spearman2018beyond}. An off-ball player creates high scoring opportunity if they control a region of the pitch to which a pass is feasible and and from which a goal is likely. We report the OBSO for the receiver at the point that a pass is received, measured at the time of the pass, averaged over passes of range 15m or more. It is high if the off-ball player positions themselves well and the passing player makes successful passes, hence it is a measure of pass quality and receiver positioning. See Appendix \ref{sec:Appendix:Behavior:OBSO} for details. \\ %
        \bottomrule
    \end{tabular}
    \caption{Behaviour statistics collected during games against evaluation agents. We consider statistics that characterize (a) basic locomotion and other movement skills; (b) football skills; and (c) team work. Further details for each statistic can be found in  Appendix \ref{sec:Appendix:Behaviors}. }
    \label{table:behaviourStatistics}
\end{table}

\paragraph{Counterfactual Policy Divergence} We use the {\it counterfactual policy divergence} (CPD) technique \cite{LiuSoccer} to measure the extent to which the behaviour of a player is influenced by different objectss in the football scene (ball, teammate, opponent). We measure the KL-divergence induced in the policy by repositioning (or changing the velocity) of a single object. For an object $b$ we define $$CPD(\pi, b) := \bE_{s_b}[\bE_{s_{b'}|s_b}[D_{KL}(\pi(\cdot|\phi^0(s_{b'})) || \pi(\cdot|\phi^0(s_b)) )]]$$ where $s_b$ is an initial game state and $s_{b'}$ is a state identical to $s_{b}$ but with the position of object $b$ re-sampled and replaced uniformly at random on the pitch, and (recalling Section~\ref{sec:Methods:Preliminaries}) where $\phi^0()$ is the observation function for football.\footnote{Our agents are recurrent but $\pi(\cdot|\phi^0(s_b))$ and $\pi(\cdot|\phi^0(s_{b'}))$ are measured on the first timestep of the game -- so that $\phi^0(s_b)$ and $\phi^0(s_{b'})$ are the observation-action histories at that point -- with the agent's LSTM in a random initial state, so that $\pi(\cdot|\phi^0(s_b))$ and $\pi(\cdot|\phi^0(s_{b'}))$ can be consistently compared. We therefore drop the conditioning on history for ease of notation.} A large CPD indicates that the action distribution of the player (i.e.\ the behaviour)  would have been very different had the object $b$ been in a different position, a small CPD indicates that the object has little influence on the player's behaviour.

\begin{figure}
  \centering
  \includegraphics[width=1.\textwidth]{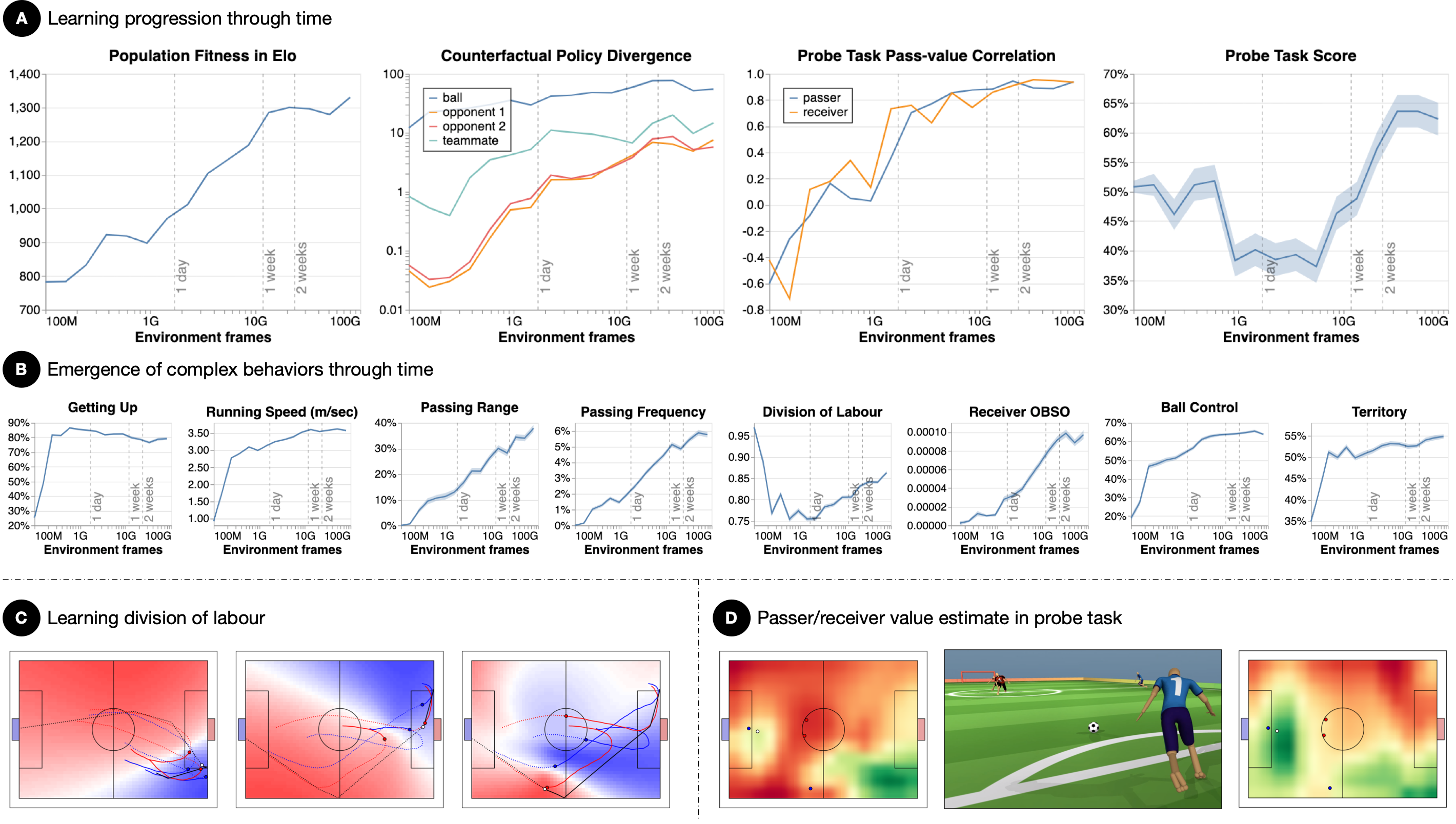}
  \caption{ {\bf (A)} {\it Agent performance}  measured by Elo against a set of pre-trained evaluation agents increases as the agents learn football behaviours. {\it Counterfactual policy divergence} by entity: early in training, the ball (blue curve) induces most divergence in the agent policy; other players have progressively more influence on the agent's policy as training progresses. {\it Pass-value-correlation} increases for both passer and receiver over training as coordination improves. Agent's {\it probe score} drops below 50\% early in training, but improves to 60\% as the agents learn coordinated strategies, and identify the value of teammate possession. {\bf (B)} Emergence of behaviours and abilities over training. Early in training (up to 1.5 billion environment steps or approximately 24 hours of training) {\it running speed} and {\it possession} increase rapidly and the ability to {\it get up} is effectively perfected. {\it Division of labour} decreases in this early phase as agents prioritize possession and learn uncoordinated ball chasing behaviours. After 1.5 billion environment steps a transition occurs in which {\it division of labour} improves and behaviour shifts from individualistic ball chasing to coordinated play. In this second phase {\it passing frequency}, {\it passing range} and {\it receiver OBSO} increase significantly. { (\bf C)} Division of Labour and passing plays: solid/dashed lines indicates past/future trajectories of the red and blue players and the ball (black line). The two left frames are at the point in time of the pass; the receiver turns to anticipate an upfield kick before the pass, leaving the teammate to control the ball. Rightmost frame is the point of reception. {\bf (D)} Typical probe task initialization with blue player 1 ("passer") initialized in its own half, and player 2 ("receiver") initialized on a wing and two defenders in the centre. Right: receiver value (scoring channel) as a function of future ball position on the pitch. Regions of high value in green and low value in red. Left: passer value function. Both receiver and passer register higher value when the ball travels to the right wing, where the receiver is positioned.} \label{fig:behaviours}
\end{figure}

\paragraph{Probe Task} We study the agent's behaviour under controlled conditions in a {\it probe task}. The probe task is a short football game, lasting 5 seconds, with a specific initial configuration designed such that kicking the ball towards a teammate should be a beneficial strategy. 
We pitch two player instances as attackers against a team of two defending agents. The defenders are randomly sampled among the evaluation agents.
The players are positioned randomly within small prescribed regions according to their respective roles: one of the attacking players takes the role of a ``passer''  and is initialized deep in its own half with the ball close by; the second player takes the role of  ``receiver'' and is initialized near the centre line but on either wing with equal probability. The  two defenders are always initialized near the centre, see Figure~\ref{fig:behaviours}D. We quantify the results using two statistics which measure (a) whether the passer tends to kick the  ball towards the teammate; and (b) whether the passer associates a higher likelihood of scoring with potential passes. We describe the statistics in Table \ref{table:ProbeStatistics} and provide further details in Appendix \ref{sec:Appendix:Behaviors:ProbeTasks}. For both statistics we report averages over multiple instances of the probe task with different initial conditions.

\begin{table}[h!]
    \centering
    \small
    \begin{tabular}{lp{11cm}l}
        \toprule
        \textbf{Name} & \textbf{Description} \\ %
        \toprule
        \vspace{0.2cm}
        \emph{Probe score} & Measures whether the passer tends to kick the ball in a direction correlated with the receiver's position. We determine whether the velocity of the ball parallel to the goal line points towards the side where the receiver is positioned or not. A score of 1 means the passer always kicks forwards and in the receiver direction, 0 means it never does.\\
        \emph{Pass-value-correlation} & Helps to determine whether the behaviour of the agent is driven by learned knowledge of the value of certain game states. We measure whether the passer's and receiver's value functions (specifically the scoring reward channel) register higher value when the ball travels towards the receiver, rather than away. Intuitively, a value close to 1 suggests that that the agent predicts a higher likelihood of scoring a goal when the ball travels towards the receiver than otherwise; values close to -1 suggest the opposite. \\ %
        \bottomrule
    \end{tabular}
    \caption{Statistics to quantify the outcome of the probe task analysis. See Appendix \ref{sec:Appendix:Behaviors:ProbeTasks} for further details.}
    \label{table:ProbeStatistics}
\end{table}

\subsection{Results}
Key results of these analyses are displayed in Figure~\ref{fig:behaviours}. They suggest that the progression and emergence of behaviours can be divided into two phases during which players first acquire basic locomotion and ball handling skills, and subsequently begin to exhibit coordinated behaviour and team work.

\paragraph{Phase 1} During approximately the first 24 hours of training ($1.5 \times 10^9$ environment steps) the players learn the basics of the game and locomotion and ball possession improve rapidly. As shown in Figure~\ref{fig:behaviours}B, the players develop a technique of {\it getting up} from the ground, and {\it speed} and {\it possession} score increase rapidly. After six hours of training the agent can recover from 80\% of falls. Play in the first phase is characterized by individualistic behaviour as revealed by the {\it division of labour} statistic, which initially decreases as players individually optimize ball possession causing both teammates to crowd around the ball often.

Consistent with this, the ball is the object with the most influence on the agent's policy (Figure~\ref{fig:behaviours}A). After about five hours of training the counterfactual policy divergence induced by the ball is 40-times greater than that induced by the teammate and 700-times greater than the divergence induced by either opponent. Finally, the score at the probe task decreases and becomes less than $50\%$ indicating that the passer first kicks the ball in a direction opposite to the receiver (Figure~\ref{fig:behaviours}A). Pass-value correlation is also negative for roughly the first ten hours of training.\footnote{One possible explanation for this is that the agents are not coordinated at this point (as revealed by the {\it division of labour} statistic), and so act to avoid obstructing each other.} 

\paragraph{Phase 2} In the second phase cooperative behaviour and team work begin to emerge. The {\it division of labour} statistic increases and after $8 \times 10^{10}$ environment steps reaches values over 0.85 (Figure~\ref{fig:behaviours}B). This indicates a more coordinated strategy with only one player aiming to get possession of the ball (while the teammate often heads up-field in anticipation of a pass). 
Passing frequency and range also increase: after $8 \times 10^{10}$ environment steps 6\% of touches are passes and approximately 40\% of passes travel more than 10m (Figure~\ref{fig:behaviours}B). 
{\it OBSO}, which is indicative of good positioning  in human football (as detailed in Appendix~\ref{sec:Appendix:Behavior:OBSO} and \autoref{fig:obso_overview}), also increases significantly. This indicates that off-ball players learn to position themselves to receive passes in positions likely to result in a goal (Figure~\ref{fig:behaviours}B);
additionally, Figure~\ref{fig:obso_results_normalized} provides an overview of the evolution of the OBSO measure throughout training, illustrating that the number of passes with high OBSO consistently increases with training time.
Importantly, our training reward does not directly incentivize behaviour that increases this statistic. This suggests that the agents' coordinated behaviour emerges from the competitive pressure to play football well.

Consistent with the above, the CPD induced in an agent's policy by the teammates and opponents increases significantly (Figure~\ref{fig:behaviours}A) and indicates that football players' policies become more sensitive to the positions of other players: after $8 \times 10^{10}$ environment steps the ball induces a CPD less than 5-times greater than the teammate and 10-times greater than either opponent, a significant increase in the relative influence of other agents.
In the probe task the pass-value correlation similarly increases significantly (Figure~\ref{fig:behaviours}A). After $8 \times 10^{10}$ environment steps it has reached values between 0.2 and 0.4 for both receiver and passer. This indicates that both passer and receiver assign higher value to situations in which the ball travels to the receiver's wing and more generally to situations in which the teammate has possession. Performance at the probe task also increases and after $8 \times 10^{10}$ environment steps the passer kicks the ball to the receiver's wing 60\% of the time. 
Taken together these observations suggests that agents understand the benefit of kicking toward a teammate and are able to act accordingly.

\begin{figure}
  \centering
  \includegraphics[width=0.96\textwidth]{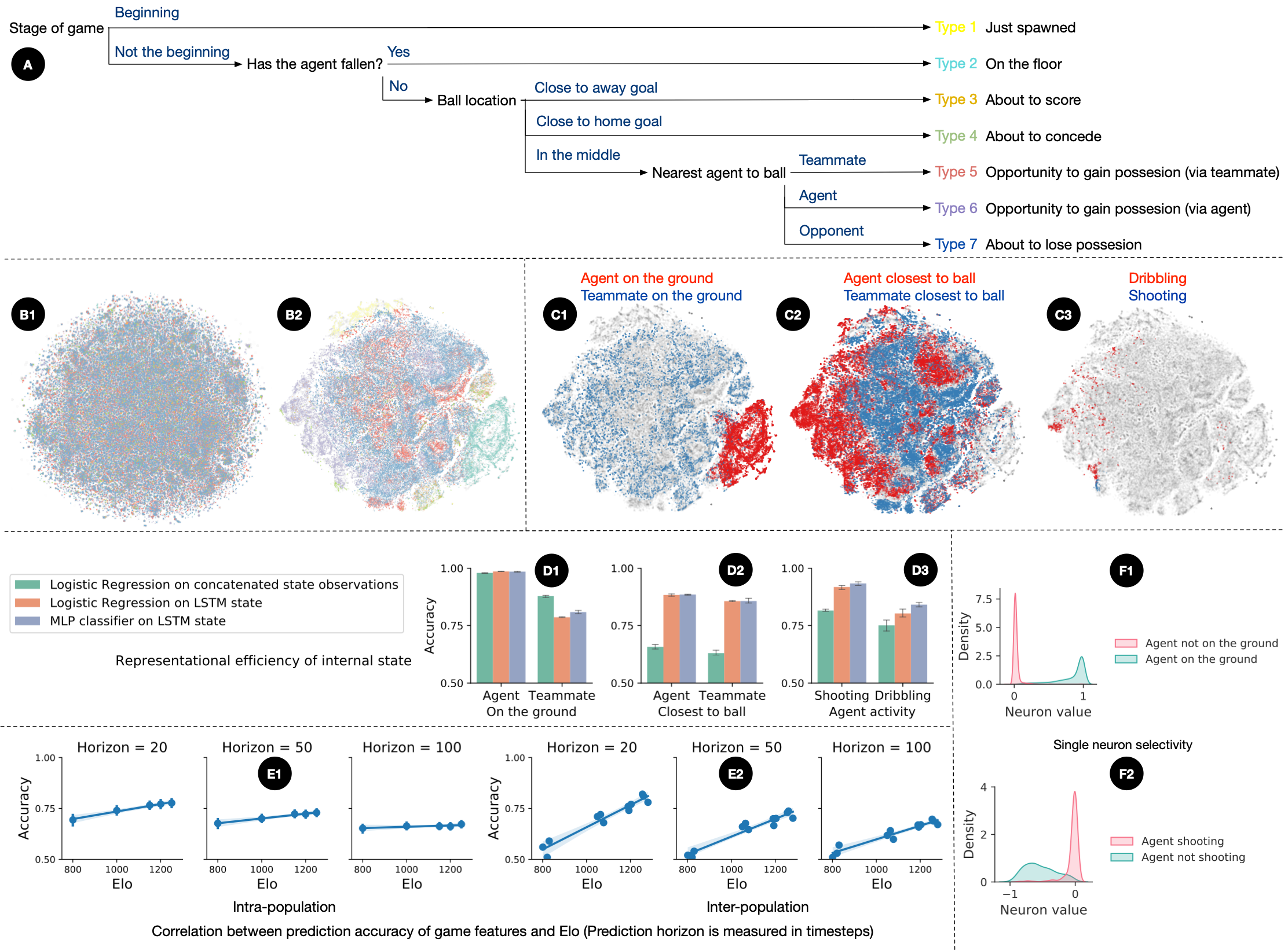}
  \caption{{\bf (A)} Hierarchical labelling of game states with high-level features from the perspective of human beings.
  {\bf (B)} t-SNE embeddings of raw observations of game state observed by our best agent {\bf (B1)} and corresponding internal states {\bf (B2)} of the agent from 200,000 timesteps, sampled from 512 gameplays consisting of about 750,000 timesteps in total. Clusters with different colours correspond to the high-level game features in A.
  {\bf (C)} Pair-wise comparison for game features: {\bf (C1)} The agent developed a highly localized cluster for \textit{agent has fallen}, while the internal states of \textit{teammate has fallen} is more distributed. This suggests that the consequent behaviour after \textit{agent has fallen}, i.e. trying to get up from the ground, is more deterministic than the action after observing its teammate has fallen; {\bf (C2)} The major clusters of \textit{agent closest to ball} being separable from those of \textit{teammate closest to ball} suggests that the agent is able to discriminate well who is in possession of the ball, which is arguably a prerequisite for dividing labour in team coordination; {\bf (C3)} Presents the t-SNE embeddings of \textit{dribbling} and \textit{shooting}. Specifically, \textit{dribbling} and \textit{shooting} are kicking with different consequences, and the fact that the clusters do not intersect suggests that the agent perceives these two types of kicking as different behaviours.
  {\bf (D)} Representational efficiency of the agent's internal state for the game features in (C). Each group of bars correspond to the performance of three classifiers on a binary classification task regarding a current game feature, e.g., classifying whether the agent is on the floor with Logistic Regression (LR) on raw observations, LR or an Multi-layer Perceptron (MLP) classifier on the agent's internal state at the same timestep. For \textit{agent on the ground}, \textit{agent closest to ball}, and \textit{teammate closest to ball}, the linear classifier on the agent's internal state performs as well as the MLP classifier. 
  {\bf (E)} Correlation between the agent's performance in Elo and the predication accuracy of future game features with LR on the agent's internal state. We investigated two cases: within the population of our best agent {\bf (E1)} and across the populations of different training regimes used in the ablation study Section~\ref{sec:Ablation} {\bf (E2)}. In both cases, the prediction accuracy is positively correlated with Elo.
  {\bf (F)} Single neuron selectivity for \textit{agent shooting} {\bf (F1)} and \textit{agent on the ground} {\bf (F2)}. For each of these game features, the agent developed a highly discriminating dimension in its internal state.}
  \label{fig:knowledge}
\end{figure}

\section{How Football Agents Work}
\label{sec:HowAgentsWork}

The analysis in  Section \ref{sec:behaviour} has shown that agents acquire a diverse set of movement and football skills, and progress from individualistic play to playing as team. To better understand what representations support the emergence of these behaviours we conduct multiple analyses. In Section \ref{sec:HowAgentsWork:Internal} we investigate how the agents represent the state of the game internally. In Section \ref{sec:HowAgentsWork:Teachers} we study how the representation of mid-level skills in the form of drill teachers is used by the agents.

\subsection{Internal Representations of the Game State}
\label{sec:HowAgentsWork:Internal}

Studies of human athletes have shown that the ability to interpret game situations and to predict how the game is going to evolve are correlated with good performance \cite{muraskin2015knowingswing}.
Analogously we hypothesize that the simulated players' 
behaviour is driven by an internal representation that emphasizes important features of the game state and allows predictions about the future.

We perform an analysis of the agents' internal representation similar to \cite{JaderbergCTF}. 
We record trajectories of full football games and label each game state with a set of binary \emph{features} that characterize important high-level properties of that state. A full list of features is provided in Table \ref{acc:knowledge_fea} in Appendix~\ref{sec:knowledge_appendix}. We further introduce a set of mutually exclusive \emph{labels}. Each label corresponds to a conjunction of game features as shown in the decision tree in Figure~\ref{fig:knowledge}, Panel A. 
The features and labels are chosen such that they are meaningfully related to desirable high-level behaviour.
For instance, successful coordination requires awareness of the teammates position; and movement skills such as getting up from the ground require an understanding of the agent's own pose.
Similarly, the ability to predict the future occurrence of events such as \emph{kick by agent has resulted in goal} or \emph{kick by agent has resulted in pass} suggests that the agent has a representation of the consequences of its own actions.

\paragraph{Qualitative Analysis} For each time step of a trajectory we gather two pieces of information: the raw observation of the game state available to the agent, and the internal state of the agent's LSTM. 
To understand how the agent's internal representation of the game state is organized we perform dimensionality reduction on the agent's recurrent state.
To assess whether similar game states are encoded in similar ways we colour code each state with one of the mutually exclusive labels listed in Figure~\ref{fig:knowledge} Panel A. Results are shown in Panels B1-B2 and C1-C3. To understand whether the internal representation increases the salience of certain high-level features of the game state we further contrast the resulting picture with a similarly labelled t-SNE embedding of the raw observations in Figure~\ref{fig:knowledge} Panel B1.

The t-SNE plots show that the internal agent states do indeed cluster in accordance with conjunctions of the high-level game state features described above.
For instance, we find that the t-SNE embeddings of shooting and dribbling form separate clusters (Panel C3) which implies that the agent is able to discriminate these two behaviours through its internal state. We also observe that similar clusters are not present in the t-SNE embedding of the raw observation (Figure~\ref{fig:knowledge} Panel B1), suggesting that the agent's internal representation transforms the raw observation to provide easy access to  high-level properties of the game state.

\paragraph{Recognition of Game Features}
Next, we identify the key game features %
that are emphasized by the agent's representation: in Figure~\ref{fig:knowledge} (Panel D).
We compare two classification methods: 1) linear classification (Logistic Regression) from the agent’s LSTM state, 2) linear classification from the agent’s raw observation.\footnote{We concatenate observations from five consecutive timesteps to ensure that information about velocities can be estimated and provide a form of short term memory.}
As in \cite{JaderbergCTF}, we say that an agent has \emph{knowledge} of a game feature, if a linear classifier on the agent's internal state accurately models the feature. Similarly, we say that the information about the game feature can be easily accessed from raw observations, if a linear classifier on the raw observation models the feature well. Improved classification accuracy by a linear classifier using the internal state of the agent, compared to using the raw observation, suggests that the agent's representation has been shaped to make the feature easily accessible.

We identify three key patterns:
a) Some features are already easily decoded from the raw observation and are preserved in the agent's internal representation.
For example, for the feature \emph{agent on the ground}, the linear classification performance from the raw observation is already high and does not improve when classification is performed from the agent's internal representation (Panel D1, left).
b) Other features are de-emphasized in the agent's internal representation. For example, for \emph{teammate has fallen}, linear classification from the agent's raw observations outperforms linear classification from the agent's internal state (Panel D1, right). This may indicate that the feature is of lower behavioural relevance to the agent.
c) Finally, a majority of game features is emphasized by the agent's internal representation. This is the case, for instance, for the features \emph{agent is closest to ball}, \emph{teammate is closest to ball} and \emph{shooting}, for which linear classification from the agent's internal state outperforms classification from the agent's raw observations (Panel D2, D3).
We summarize how the agent recognizes different game features in Table~\ref{acc:knowledge_fea} in Appendix~\ref{sec:knowledge_appendix}.

\paragraph{Representational Efficiency}
To gain a deeper understanding of the nature of the agent's internal representation we compare two additional decoding schemes. First, we compare linear decoding to nonlinear decoding with a 2-layer Multilayer Perceptron (orange vs. purple in Panel D).
A two-sided t-test does not indicate a significant difference between the two schemes with $p$-values > 0.05 for the game features \textit{agent on the ground} (0.90), \textit{agent closest to ball} (0.55), \textit{teammate closest to ball} (0.97), and \textit{shooting} (0.09). This suggests that the representation is efficient in the sense that it provides access to most information via a simple linear decoding scheme.
We summarize the representational efficiency of all game features in Table~\ref{acc:knowledge_fea} in Appendix~\ref{sec:knowledge_appendix}.

We further analyze the activation patterns of individual units (dimensions) of the internal state for different game situations. Results are shown in Figure~\ref{fig:knowledge} (Panels F1, F2). We find that several units are highly discriminatory and simply thresholding their activation can provide accurate information about the game state. This suggests that information about some features is localized and can be decoded from the internal state of the agent with particularly simple means and without reference to the full population. This bears some similarity to sparse coding schemes identified in monkey and human brains \cite{quiroga2005invariant,quiroga2008sparse}.

\paragraph{Prediction of Future Game States}
Finally, we investigate how the agent's ability to predict \emph{future} game state with its internal representation correlates with its performance (Figure~\ref{fig:knowledge}, Panels E1 and E2).
To study this question, given a snapshot of an agent, corresponding to an Elo score, we repeat the analysis above with its internal states as input to the linear classifier and use these to predict a future game feature.
We perform two analyses: 1) Intra-population anlaysis (Figure~\ref{fig:knowledge}, Panel E1): within the population of our best agent, we first sample five agents with their Elos ranging from 800 to 1250. For each agent,
we collect the prediction accuracy of all the game features listed in Table~\ref{acc:knowledge_fea}, and report the mean and standard deviation. %
Finally, we correlate the prediction accuracy with Elo with linear regression.
2) Inter-population analysis (Figure~\ref{fig:knowledge}, Panel E2): we consider the four populations in the ablation study described in Section~\ref{sec:Ablation}. For each population we take a snapshot at  $40 \times 10^9$ environment steps and select the top 3 agents (out of 16), and repeat the correlation analysis conducted in the inter-population case. For both cases, we repeat the analysis with different prediction horizons. For the intra- and inter-population analysis we find that the agent's ability to predict future game features is correlated with good performance.

\subsection{Transfer of behaviour Via Teachers and Skills} \label{sec:HowAgentsWork:Teachers}

\begin{figure}[htp]
  \centering
  \includegraphics[width=\textwidth]{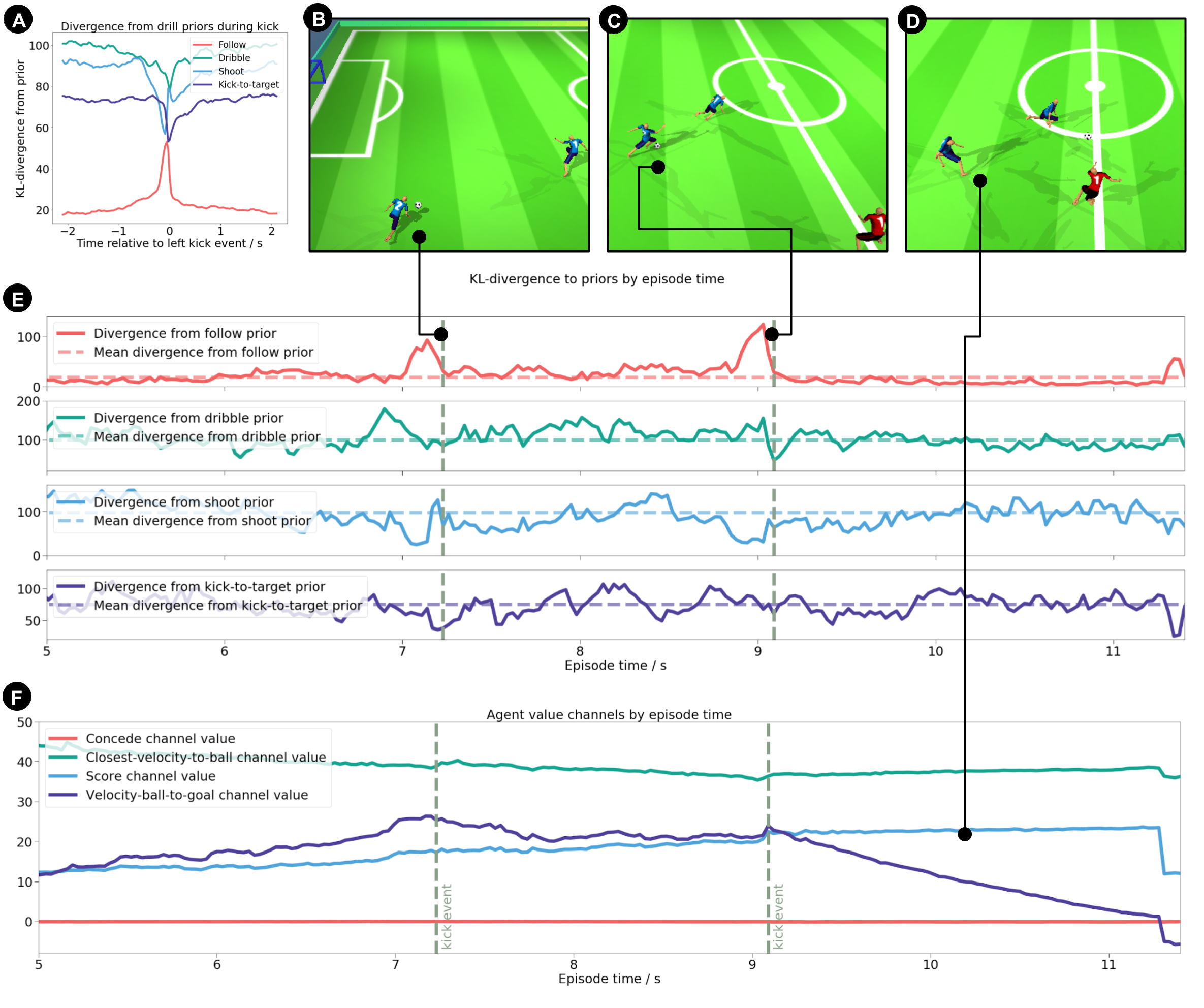} 
      \caption{{\bf (A)} Event-triggered analysis: KL-divergence to drill experts during kick events with the left foot. The football agent deviates significantly from the {\it follow} expert (red) during the kick and aligns more closely with the {\it shoot} (blue) and {\it kick-to-target} (purple) and {\it dribble} (green) experts. {\bf (E)} The KL-divergence between the blue number 2 player's policy and the four drill experts during a single specific play, in which the agent kicks the ball twice {\bf (B and C)} and ultimately scores. There is a typical deviation from the {\it follow} expert, and closer alignment with {\it shoot} expert immediately prior to the two kick events. {\bf (F)} We track the separate value function channels to gauge the drivers of the agent's behaviour. The {\it closest-velocity-to-ball} shaping reward (green) dominates the agent's internal value function overall. But at the time of the first kick, deep in the agents own half, and with control of the ball, the {\it velocity-ball-to-goal} channel makes a significant contribution to the value function. At the time of the second kick (a shot) the {\it score} channel is also significant, and rises further after the agent makes an accurate shot {\bf (D)} before finally dropping when the goal is scored.} \label{fig:HowAgentsWork}
\end{figure}

As explained in Section~\ref{sec:Method} the football agents are regularized towards four drill prior policies that represent mid-level skills: {\it follow}, {\it dribble}, {\it shoot} and {\it kick-to-target}. Unlike certain hierarchical architectures \citep[e.g.][]{heess2016learning,nachum2018data}, the drill priors are not directly reused as part of the policy. This raises the question whether and to what extent they influence the football players' final behaviour. In particular, we are interested to understand whether players make use of all skills and whether they adaptively switch between skills depending on the context.

We answer these questions by playing episodes with the trained football policy and simultaneously stepping the four drill experts at states along the trajectory generated by the football agent. By measuring the KL-divergence between the football policy and the drill priors, we can gauge how similar a player's behaviour is to each of the drill priors. Statistics for alignment with the drill experts were collected over one hour of play.

The results of this analysis are shown in Figure~\ref{fig:HowAgentsWork}. The KL-divergence between the agent's policy and the {\it follow} prior is lower than that between the football agent's policy and the other three drill priors. But Figure~\ref{fig:HowAgentsWork}A shows results of an event-triggered analysis in which we measure the KL-divergence to the priors before and after kicking events with the left foot, aggregated over one hour of play. Agents tend to deviate significantly from the {\it follow} prior during preparation for a kick, and align more closely to {\it shoot}, {\it dribble} and {\it kick-to-target} priors over the course of the kick before returning to the long-term average deviation. In addition to an aggregated analysis, in Figure~\ref{fig:HowAgentsWork}E we investigate the pattern of behaviour, and alignment with the drill priors, during a single play in a specific episode. We track the KL-divergence between one player's policy and the four drill prior policies. We see the typical divergence from the {\it follow} prior during kick preparation and alignment with the {\it shoot} prior during the two kicks. We also track the contribution of the four reward channels to the value function during the same episode, which shows that the contribution of the scoring channel increases prior to a successful shot.

\section{Ablation Study}
\label{sec:Ablation}

\begin{figure}
    \centering
    \includegraphics[width=\textwidth]{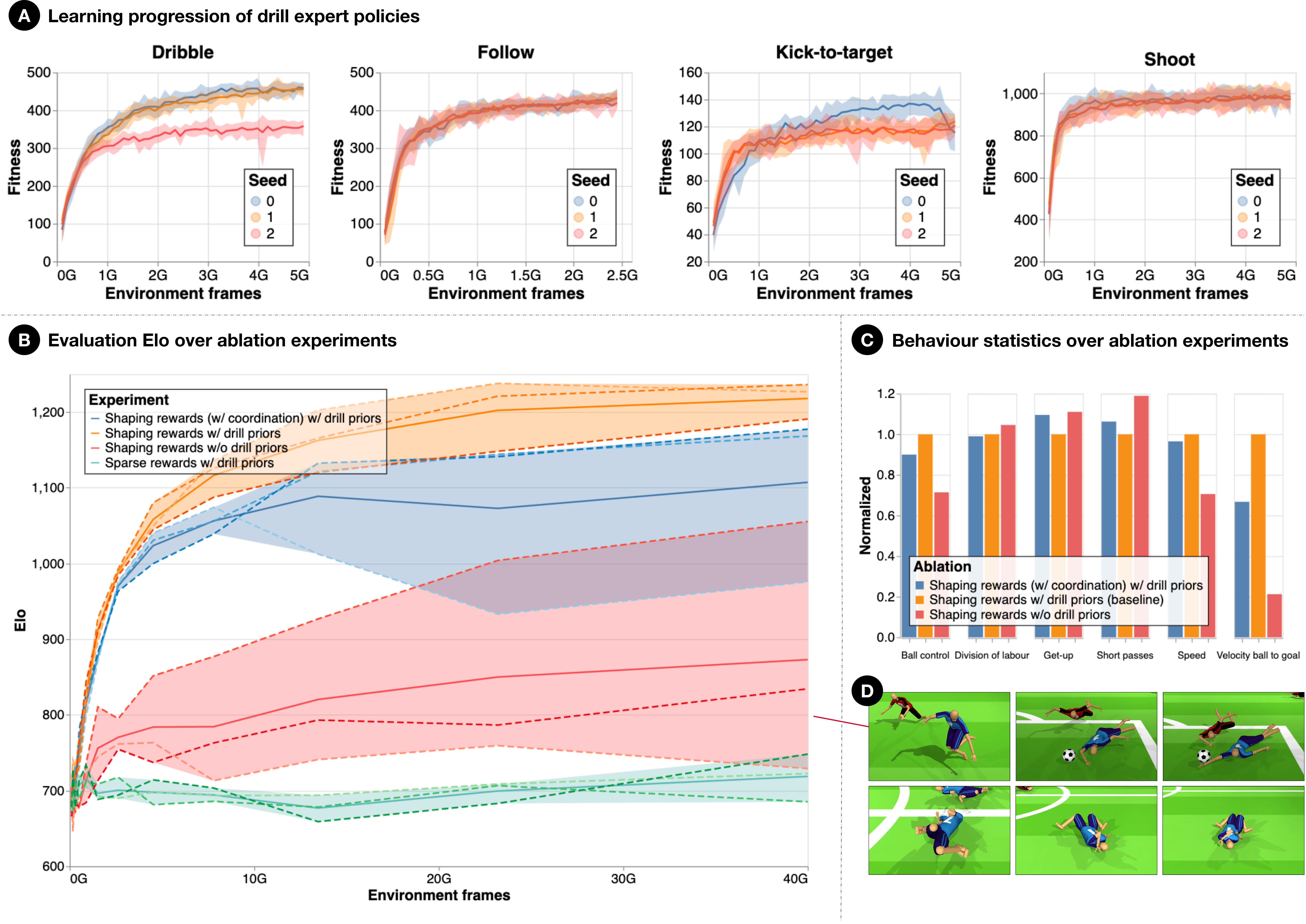}
    \caption{{\bf (A)} Training progression for the four football drills quantified by fitness (environment reward) over time. For the {\it follow} and {\it shoot} drill, performance was roughly equivalent across the three seeds, and these drills were learned quickly. The harder {\it dribble} and {\it kick-to-target} drills were learned more slowly and with some variance in performance across seeds. {\bf (B)} Football training progression for the four training methods quantified by Elo against evaluation agents by training environment step. There is a clear separation in performance across training methods. The sparse reward framework (green curve) performed consistently very poorly. The training approach without the use of drill priors (red curve) also performed poorly, but had much higher variance across seed; these agents suffered from getting stuck in several different locomotion or ball-control local optima. Our standard method with two dense shaping rewards reached the highest performance (orange curve). Including additional shaping rewards (blue curve) performed reasonably well but generally worse than the agents with fewer shaping rewards. {\bf (C)} Comparing behaviours between the best performing agent (in orange), the best agent with additional coordination shaping rewards (in blue) and the best agent without drill prior teachers (in red). We observe that the introduction of coordination rewards does not affect most statistics except that the average velocity of the ball to the goal is approximately 30\% lower than the baseline standard agent. Without drill priors, the agent's speed and velocity of the ball to the goal statistics were significantly lower than the other agents learning from drill priors. {\bf (D)} Poor ball control local optima for an agent trained without drill priors: (top) the agent dives on the ball to manoeuvre it with its hands; (bottom) the agent locomotes along the ground, which is easier to learn than upright locomotion, but is a poor local optima.}
    \label{fig:ablation}
\end{figure}

In order to assess the importance of the different components of our learning framework, we considered several variations of the training scheme outlined in Section \ref{sec:Method}:
\begin{enumerate}
\item \emph{Drill priors and basic shaping rewards}: the standard agent described in Section \ref{sec:Method} which uses low-level skills, mid-level skill priors, and simple dense shaping rewards.
\item \emph{Drill priors and sparse rewards}: the agent described in Section \ref{sec:Method} with low-level skills, and mid-level skill priors, but without dense shaping rewards. The agent is trained only with sparse rewards for scoring and conceding a goal.
\item \emph{Shaping rewards but no drill priors}: 
the agent described in Section \ref{sec:Method} with low-level skills, and shaping rewards but no mid-level skill priors.
\item \emph{Drill priors; shaping rewards; additional team-level coordination shaping rewards}:
the agent described in Section \ref{sec:Method}, with two additional team-level shaping rewards designed to encourage coordinated behaviour in an attempt to improve performance. The first, {\it vel-ball-to-teammate-after-kick}, encourages the agent to kick the ball towards a teammate. The second, {\it territory}, encourages the teammates to spread out such as to control as large a fraction of the pitch as possible.
\end{enumerate}
For each of the four training schemes we follow the procedure outlined in Section \ref{sec:Experiments} and train three separate populations of 16 agents each. For consistency we use the same low-level skill module and the same three sets of drill priors for each training scheme as explained in Section \ref{sec:Experiments}.

\paragraph{Results} 
We evaluate trained agents by playing matches against the evaluation agents as explained in Section \ref{sec:Experiments}. Results are shown in Figure~\ref{fig:ablation}. There is a clear separation in performance across training methods. Players trained with drill priors and sparse rewards only (green curve) performed very poorly. They learned to score goals, but were unable to get back up after having fallen to the ground and they generally moved slowly (perhaps in an attempt to avoid falling). This suggests that, with mid-level skill priors, it is possible to learn with sparse reward only, but movement skills still benefit from shaping rewards.

Training without the use of drill priors (red curve) also led to poor performance and much higher variance across the independent experiments. In all the three experiments the players learned degenerate movement skills. In two of the three experiments the players did not learn to run but instead moved across the pitch by rolling on the floor. In a third experiment the players did learn to run, but manoeuvred the ball by falling on it and hitting it with their hands.\footnote{This technique is not illegal in our environment but is nonetheless a poor local optima} Examples of these movement techniques are shown in Figure~\ref{fig:ablation}D. These results suggests that the low-level skill module is insufficient to adequately shape the player's movements and that the mid-level behaviour priors derived from the training drills play an important role in this regard.

The best performing players were produced by the standard training scheme described in Section \ref{sec:Method} and analyzed in Sections \ref{sec:Experiments}-\ref{sec:HowAgentsWork}. Players from all three independent experiments achieved a similarly high performance (orange curve).

The training scheme that uses additional shaping rewards to encourage more coordinated behaviour performed well (blue curve) but worse, overall, than the standard scheme detailed in Section \ref{sec:Method}. A comparison of the behaviour statistics of the players obtained with the two schemes reveals that players trained with the additional shaping rewards exhibit similar movement and ball handling skills: \emph{getting up}, \emph{speed} and \emph{ball control} are similar for the two types of players. However, the average velocity of the ball to the goal is approximately 30\% lower when training with the additional shaping rewards (Figure~\ref{fig:ablation}C), indicating that players kick less towards the opponents' goal. These results suggest that shaping rewards that reliably encourage coordinated play can be non-trivial to design, and that the need to balance incentives provided by conflicting shaping rewards can negatively affect performance. Furthermore, PBT and evolution appear to struggle to optimize the larger set of hyper-parameters that results from growing the set of shaping rewards. This may be a consequence of the relatively small population size of 16. A possible alternative approach to investigate in future work would be use of behaviour priors for cooperative play.

\section{Related Work}
\label{sec:Related}

Our work combines a range of problems into a single challenge: multi-scale behaviour; dynamic movement and control of physical embodiments; coordination and teamwork; as well as robustness to a range of adversarial opponents. These are all, separately, fundamental open-problems in AI research, each receiving significant focus. In this section, we will review the prior work on each of these topics.

\paragraph{Humanoid Control}
Human motor behaviour and the biological mechanisms that give rise to it have been widely studied in their own right in various disciplines ranging from kinesiology to motor neuroscience. The artificial reproduction of human-like behaviour in humanoid robots and virtual characters are studied predominantly within the computer graphics and robotics communities.  For computer graphics researchers, the motivation is to develop humanoid characters that produce realistic movements as well as natural interactions with physical environments, and one route towards solving this involves controlling humanoids in physics simulators \cite{Faloutsos2001, yin2007simbicon, coros2010generalized, liu2012terrain, peng2017deeploco, liu2018learning, peng2018deepmimic, chentanez2018physics, lee2019scalable, park2019learning, bergamin2019drecon}.  The approaches employed by the graphics community range from classical control approaches through to contemporary deep learning and substantially overlap with methods developed for high-dimensional and humanoid control within the AI literature \cite{tassa2012synthesis, mordatch2015interactive, schulman2015high, heess2017emergence, merel2017learning}.  A major challenge in the control of complex bodies remains the ability to compose diverse movements from a repertoire of behavioural primitives in an adaptive and goal-directed manner \citep[e.g.][]{MerelNPMP,peng2019mcp}, and to achieve object interaction \cite[e.g.][]{liu2018learning,merel2020,chao2019learning}. The present work shares motivation most closely with AI research into humanoid control, focusing on performant behaviour in a challenging physical environment but significantly increases the difficulty of the long-horizon, goal-directed nature of the task.  While Deep Reinforcement Learning (Deep-RL) has in recent years enabled rapid development of many of the humanoid control approaches for simulated environments, control of real-world humanoid robots remains difficult.  
Prominently, Boston Dynamics has made impressive advances released as video demonstrations of dynamic ``parkour" behaviours \cite{bostonDynamics}.
Though the efforts by Boston Dynamics are proprietary, they are built from considerable expertise developed by robotics researchers \cite{kuindersma2016optimization}. 
To date, these techniques appear to be rather distinct from the learning-based solutions developed in the AI community in simulation. Yet, recent partial successes transferring results from simulation to real robots \cite{hwangbo2019learning,lee2020learning,xie2019iterative,peng2020learning} suggest that learning based approaches in simulation may, in the future, play a larger role in the control of real world robots.
Importantly, most work on simulated humanoid character control, as well as on the control of real robots has so far focused on the production of high-quality movement skills rather than on the production of long-range autonomous behaviour in context rich, dynamic scenarios which is the focus of the present work.

\paragraph{Emergent Coordination}

There has been much work applying reinforcement learning to cooperative multi-agent domains %
\cite{PanaitL05,BusoniuBS08,TuylsW12,Hernandez-LealK19,yang2020overview}, and a focus on generalizing RL algorithms to the case of multiple cooperative agents. These algorithms are either limited in their applicability (e.g. \cite{LauerDistributed} which is valid for deterministic systems without function approximation) or rely on some degree of centralization, such as a shared value function during learning \cite{LoweMADDPG, FoersterCOMA}, in contrast to the setting of independent learners which we study in this work. The problem of learning coordination between independent RL agents has not been solved due to a complex joint exploration and optimization problem, which is non-stationary and non-Markovian from the perspective of any individual learner in the presence of other learning agents \cite{LaurentIndepenedentLearners, MatignonIndependentLearners, BernsteinDecPomdp, ClausBoutillierDynamics} and is particularly challenging in high-dimensional control problems with sparse, distal reward signals.

Coordinated team strategies emerged in independent RL learners in the Capture-The-Flag video game \cite{JaderbergCTF}, an environment with discrete control. Emergent cooperative behaviours such as division of labour, have recently been demonstrated in simulated physical environments but only with much simpler embodiments \cite[e.g.][]{BakerToolUse}. Achieving such behaviour in complex, articulated humanoid bodies with continuous control has not been demonstrated previously. Emergent communication between agents is a rich topic in its own right \cite{SukhbaatarComms, FoersterDIAL, MordatchEmergence}, but in this work we limit agents' abilities to rely exclusively on physically acting themselves and observing others to communicate and understand intents.

\paragraph{Multi-Agent Environments and Competition}
Competitive games have been grand challenges for artificial intelligence research since at least the 1950s \cite{SamuelCheckers, TesauroTDGammon, CampbellDeepBlue, VinyalsStarCraft}. In recent years, a number of breakthroughs in AI have been made in these domains by combining deep RL with self-play. Pitting learning agents to play against themselves (or a pool of learning agents) has achieved superhuman performance at Go and Poker \cite{SilverAlphaGo, MoravcikDeepStack}. This combination of RL and self-play provides an effective curriculum for environment complexity by automatically calibrating opponent strength to a suitable level to learn from \cite{LeiboManifesto}, and it has been speculated that intelligent life on earth has emerged during constant competitive co-adaptation \cite{DawkinsArmsRaces}. In continuous control domains in particular, complex behaviours and strategies have been shown to emerge as a result of competition between agents, rather than due to increasing difficulty of manually designed tasks \cite{sims1994evolving,BansalEmergent, BakerToolUse, AlShedivatContinuous}. In this work, we pursue this idea further by incorporating physically complex embodiments with high-dimensional control, allowing richer possibilities for agent behaviours and interactions, grounded in real physics, but otherwise not prescribed. Compared to simpler embodiments \cite{LiuSoccer,BakerToolUse} or games in environments with discrete action spaces \cite{SilverAlphaGo,JaderbergCTF} this greatly increases the difficulty, for instance of the exploration problem, and thus reduces the effectiveness of pure self-play. Some successes have been enabled by initializing agents policies via behaviour cloning from human gameplay \cite[e.g.][]{VinyalsStarCraft}, but the specific embodied nature of our domain entails that similar demonstrations are unavailable for our setting.

\paragraph{Multi-Scale Control}

The question of how to learn and reuse hierarchically structured behaviour has a long history in the reinforcement learning  \cite[e.g.][]{dayan1993feudal,Schmidhuber91neuralsequence, weiring1997hq,parr1998reinforcement,dietterich1999hierarchical,sutton1999between,barto2002} and robotics \cite[e.g.][]{brooks1986robust,albus1993reference,paraschos2013probabilistic,daniel2016hierarchical, Kambhampati, Sharir1989} literature. %
For self-learning systems highly-structured, adaptive, long-horizon behaviours pose a number of challenges including exploration, credit assignment, and model capacity. Solution strategies that have been proposed often rely on hierarchical architectures and a large number of different approaches have been pursued in the motor control and general RL literature. 

Approaches vary along a number of dimensions. For instance, a separation of concerns between different model components can be achieved either via architectural constraints \cite[e.g.][]{heess2016learning,galashov2018information} or dedicated sub-objectives \cite[e.g.][]{dayan1993feudal,vezhnevets2017feudal,nachum2018data,nachum2018nearoptimal,gregor2016variational,hausman2018learning, stone2000layered} and different architectures can be employed to model the behaviour of interest, including continuous behaviour embeddings \citep[e.g.][]{heess2016learning,wang2017robust,hausman2018learning,haarnoja2018latent,tirumala2020behavior,MerelNPMP} or discrete options \cite{sutton1999between,bacon2017option,fox2017multi,frans2018meta,merel2019hierarchical,wulfmeier2020compositional,wulfmeier2020dataefficient,merel2018hierarchical} and enforce temporally correlated behaviour \cite[e.g.][]{sutton1999between,bacon2017option,krishnan2017discovery,nachum2018data,wulfmeier2020dataefficient,tirumala2020behavior} or not \cite[e.g.][]{MerelNPMP,galashov2018information,tirumala2020behavior}. Furthermore, the behaviour to be modeled can be copied from demonstrations \cite{wang2017robust,fox2017multi,krishnan2017discovery,MerelNPMP}, learned as part of a multi-task framework \cite[e.g.][]{heess2016learning,galashov2018information,tirumala2020behavior,riedmiller2018learning}, or derived from intrinsic rewards \cite[e.g.][]{dayan1993feudal,vezhnevets2017feudal,nachum2018data,gregor2016variational,hausman2018learning,eysenbach2018diversity}; and learning can proceed either online while interacting with an environment or offline from a fixed set of data.
One fundamental problem associated with hierarchical architectures and associated training regimes is that they often impose undesirable constraints on the resulting behaviour, for instance, due to poorly chosen subgoal spaces; restrictive architectures such as enforced temporal correlations; or unsuitable decomposition of the learning scenario \citep[e.g.][]{merel2018hierarchical,nachum2018data,nachum2018nearoptimal}. 
Recently, there have been attempts to separate the modeling of hierarchical behaviour from the use of hierarchically structured architectures \cite[e.g.][]{teh2017distral,galashov2018information,tirumala2020behavior}. 

Our agent brings together several of these ideas \cite{teh2017distral,galashov2018information,MerelNPMP,tirumala2020behavior} in a multi-scale learning architecture. The behaviours that our agents exhibit originate from a mix of demonstrations, pre-training tasks and end-to-end training. Architecturally, the agent combines a skill module for low-level motor control, with a non-hierarchical behaviour prior for mid-level skills, to achieve multiple levels of control spanning core locomotion and movement skills, football-specific skills, and team level coordination. The multi-stage training scheme decomposes the learning problem and provides adequate behavioural constraints and shaping without unduly restricting the final behaviour.

\paragraph{RoboCup and Simulated Football} 

A longstanding grand challenge in AI concerns the development of autonomous robots capable of playing human-level football, as set out in the well-known RoboCup project \cite{robocup,KitanoRoboCup}. Reinforcement learning is an area that has received attention since the early days of RoboCup to tackle some of the main technical challenges, which emerge in various of its football leagues, in isolation. These works typically focus on the handling of large state-action spaces \cite{LNAI98-tpot-rl,TuylsMM02}, skill learning \cite{icra04,AAI04,LNAI2006-manish,LNAI2006-peggy,LNAI10-hausknecht,B-Human19}, the keep-away and half-field offense tasks and multi-agent coordination \cite{{AB05,LNAI2007-shivaram,LNAI09-kalyanakrishnan-1,LNAI2006-shivaram,Riedmiller09,Lauer10,Gabel10,RiedmillerHLL08,Riedmiller_karlsruhebrainstormers}}, grounded simulation learning for improved skills \cite{AAMAS13-Farchy,AAAI17-Hanna,IROS20-Karnan,IROS20-Desai} (e.g. in sim-to-real and back), skill learning in 3D humanoid football \cite{Urieli11,abreu2019learning}, and deep reinforcement learning for parameterised action spaces \cite{ICLR16-hausknecht}. %
Yet, despite these examples of applications of reinforcement learning in the RoboCup domain, many successful recent RoboCup competition entries learn or optimize only a subset of the components of the control architecture  \cite[e.g.][]{B-Human19,macalpine2019robocup}. One successful RoboCup approach related to our method is Layered Learning \cite{stone2000layered, MacAlpineOverlapping} which uses reinforcement learning at multiple levels of a pre-defined hierarchy of skills, from individual ball interaction to multi-agent behaviours such as pass selection. In Layered Learning a bottom-up hierarchical task decomposition is given and implemented architecturally, with the output of one layer feeding into the next. In contrast, in our system, the pre-learned skills are not transferred as architectural components in a hierarchy; rather, all behaviours are learned at the level of motor intention and the pre-learned skills are transferred as priors to bias soccer behaviour, rather than as parametrized skills to precisely reproduce. The mid-level individual skills and multi-agent behaviours are therefore more tightly coupled. Our NPMP motor primitive module is transferred as a low-level component but the NPMP is not optimized to perform any particular skill, but is rather a reparameterization of the action space.

Several simulated football environments have been proposed in the AI literature. These include, most prominently, the RoboCup 2D and 3D league \cite{KitanoRoboCup, RoboCupWeb}, as well as more recent additions including the Google Research Football Environment  \cite{kurach2019google}, and the immediate predecessor of this work \citep{LiuSoccer}. Both the RoboCup 2D simulation league as well as the Google Research Football Environment use abstracted action spaces and do not focus on motor control for embodied agents. The simulated RoboCup 3D humanoid league is modeled around the {\it NAO} robot used in the Standard Platform league  and emphasizes alignment with the real-world robotics platform and RoboCup rules. Teams consist of eleven robots, and the game flow can be interrupted for various reasons, such as set pieces, and the environment also allows for some non-physical actions such as blocking opponent players from approaching while a pass is being executed. As discussed in Section \ref{sec:Environment}, our environment attempts to isolate the challenge of emergent complex motor control and movement coordination in an open-ended, long-horizon task in a setting setting suitable for end-to-end learning. We thus focus on the capabilities of the simulated, human-like players and properties of the physical environment but simplify other aspects of the football game. The environment further emphasizes ease of use for learning experiments e.g.\ in terms of the stability of the underlying physics simulation, and integration with existing simulation infrastructure widely used in the literature. It conforms with a standard environment interface \cite{tassa2020dm_control} for RL environments, and it allows plug and play in the sense that different walker bodies can be used to play football, while the humanoid body used in this work can also be deployed in a range of other tasks \cite{tassa2020dm_control}.

\section{Discussion}
\label{sec:Discussion}

In this work, we have demonstrated end-to-end learning of coordinated 2v2 football gameplay of simulated humanoid players. Players learn to produce natural, human-like movements and coordinate as a team on longer timescales. They achieve integrated control in a setting where movement skills and high-level goal-directed behaviour are tightly coupled; a setting that is reflective of many challenges faced by animals and humans \citep[e.g.][]{sebanz2006joint}, and where solutions would be extremely difficult to handcraft. We have assessed the success of our approach through a number of careful analyses of the learned behaviour as well as players' internal representations. We have found, among others, that the players' behaviour improves with respect to a coordination metric from human football analytics \citep{spearman2018beyond}, and that their behaviour is driven by a representation that emphasises relevant high-level features of the game, and that they learn to make predictions about the future similar to observations in human soccer players \citep[e.g.][]{gonccalves2015anticipation,roca2012developmental}.

The behaviour emerges from a three-stage learning framework that combines low-level motor imitation, training drills for the acquisition of football skills, similar to the training of human football players, and multi-agent training with self-play for learning the full task. 
The approach implements three core ideas: Firstly, the gradual acquisition of increasingly complex skills and their subsequent reuse addresses challenges related to exploration and credit assignment that are commonly encountered in complex, long-horizon learning scenarios. Secondly, the approach relies on different types of learning signals: for instance, for human-like low-level movements good prior knowledge is available in the form of motion capture data. In contrast, for full game play similar prior knowledge would be hard to come by; the high-level game-strategy emerges from population self-play in multi-agent RL, which also helps to refine and improve the robustness of the movement skills. This provides an effective solution to the important challenge of behaviour specification in complex scenarios. Finally, the approach demonstrates how skills and behaviour priors can be used effectively to model and reuse behaviour at different levels of abstraction. It avoids common problems associated with hierarchical behaviour representations which may unduly constrain the final solution.

Overall, our study has addressed several challenges usually studied in separation: the production of naturalistic and effective human-like behaviour of humanoid players; the production of multi-scale hierarchically structured behaviour; and the emergence of coordination in challenging,  multi-agent scenarios. 
The results demonstrate that artificial agents can indeed learn to coordinate complex movements in order to interact with objects and achieve long-horizon goals in cooperation with other agents. 
The study has shown that this can be achieved by end-to-end learning methods, and how several techniques for skill transfer and self-play in multi-agent systems can effectively be combined to this end. Although our approach requires a certain level of prior knowledge of the problem, its nature did not prove particularly onerous and is readily available for many other tasks.

Obviously, the results shown in this study constitute only a small step towards human-level motor intelligence in artificial systems. %
Even though the scenario in the present paper is more challenging than many simulated environments considered in the community, it is lacking in complexity along many dimensions compared to almost all real world scenarios encountered by humans and animals. 
In particular, our work has not tackled the full football problem which is considered, in more completeness, for instance in RoboCup. In this regard, the results presented in this paper suggest several directions for future work:
Firstly, we have focused on competent 2v2 gameplay mostly for computational reasons, but it would be natural to extend to full scale football teams. Larger teams might also lead to the emergence of more sophisticated tactics. Learning in this setting may be supported, for instance, by extending our approach to include additional curricula and multi-player drills. To reduce the complexity of environment design and of the learning problem we also simplified the rules. Integrating penalties, throw-ins, or a dedicated goal-keeper role may lead to the emergence of more complex behaviours and would require a significant step in the difficulty of the movement skills (such as manipulation as in \citep[e.g.][]{merel2020}). Naturalness of individual players' movements and team tactics as well as the realism of the overall setting may be further improved by switching to egocentric vision, which would render the environment more partially observed and may favour novel movements (e.g. controlling the movement of the head or the need to run backwards) and / or team tactics (e.g. dedicated defensive roles). 
Successfully tackling these additional challenges might in the medium-term also enable an application of end-to-end learning techniques to full simulated robot football as in the RoboCup 3D league \citep{RoboCupWeb}.

Our results were obtained in a simulation environment that supports realistic physics. This has made the learning environment more open ended and allowed for the emergence of complex behavioural strategies including agile movements and physical contact between players.
Even though there have recently been some successes transferring simulation based behaviour to the real world \citep[e.g.][]{lee2020learning,openai2019solving,peng2020learning} this has not been the goal of the present experiments. Our results would currently not be suitable for direct sim-to-real transfer, nor is the developed method suitable for learning directly on robotics hardware (for a large number of reasons including the lack of data efficiency or safety considerations). 
They do demonstrate, however, the potential of learning-based approaches for generating complex movement strategies. And even though simulation is only a poor substitute for the complexity of the real world we nevertheless believe that simulation-based studies can help us understand aspects of the computational principles that may eventually enable us to generate similar behaviours in the real world. Answering the question whether and how such methods can help to achieve similar levels of sophistication in multi-scale motor intelligence for agile robotics hardware is an exciting direction for future research.

\section*{Acknowledgments}
We would like to thank: Seyhan Efendiler for his design of the players' kit; Matt Botvinick, Martin Riedmiller, Raia Hadsell, Doina Precup, Max Jaderberg, Jost Tobias Springenberg and Peter Stone for their insights and helpful comments; Tyler Liechty and Amy Merrick at DeepMind for assistance obtaining motion capture data; and Darren Carikas for assistance with the video.

\bibliographystyle{unsrtnat}

\begin{small}
\nobibliography*
\bibliography{main}

\begin{thebibliography}{162}
\providecommand{\natexlab}[1]{#1}
\providecommand{\url}[1]{\texttt{#1}}
\expandafter\ifx\csname urlstyle\endcsname\relax
  \providecommand{\doi}[1]{doi: #1}\else
  \providecommand{\doi}{doi: \begingroup \urlstyle{rm}\Url}\fi

\bibitem[Newell(1990)]{newell1990unified}
Allen Newell.
\newblock \emph{Unified Theories of Cognition}.
\newblock Harvard University Press, USA, 1990.
\newblock ISBN 0674920996.

\bibitem[Lashley(1951)]{lashley1951problem}
Karl~Spencer Lashley.
\newblock \emph{The problem of serial order in behavior}, volume~21.
\newblock Bobbs-Merrill Oxford, United Kingdom, 1951.

\bibitem[Rosenbaum(1987)]{rosenbaum1987hierarchical}
David~A Rosenbaum.
\newblock Hierarchical organization of motor programs.
\newblock 1987.

\bibitem[Schank and Abelson(1977)]{schank1977scripts}
Roger~C Schank and Robert~P Abelson.
\newblock \emph{Scripts, plans, goals, and understanding: An inquiry into human
  knowledge structures}.
\newblock Psychology Press, 1977.

\bibitem[Fuster(2001)]{fuster2001prefrontal}
JM~Fuster.
\newblock The prefrontal cortex--an update: time is of the essence.
\newblock \emph{Neuron}, 30\penalty0 (2):\penalty0 319—333, May 2001.
\newblock ISSN 0896-6273.
\newblock \doi{10.1016/s0896-6273(01)00285-9}.
\newblock URL \url{https://doi.org/10.1016/s0896-6273(01)00285-9}.

\bibitem[Merel et~al.(2019{\natexlab{a}})Merel, Botvinick, and
  Wayne]{merel2019hierarchical}
Josh Merel, Matthew Botvinick, and Greg Wayne.
\newblock Hierarchical motor control in mammals and machines.
\newblock \emph{Nature Communications}, 10\penalty0 (1):\penalty0 1--12,
  2019{\natexlab{a}}.

\bibitem[Smith et~al.(2012)Smith, Swanson, Reed, and
  Holekamp]{smith2012evolution}
Jennifer~E. Smith, Eli~M. Swanson, Daphna Reed, and Kay~E. Holekamp.
\newblock Evolution of cooperation among mammalian carnivores and its relevance
  to hominin evolution.
\newblock \emph{Current Anthropology}, 53\penalty0 (S6):\penalty0 S436--S452,
  2012.
\newblock ISSN 00113204, 15375382.

\bibitem[Sebanz et~al.(2006)Sebanz, Bekkering, and Knoblich]{sebanz2006joint}
Natalie Sebanz, Harold Bekkering, and G{\"u}nther Knoblich.
\newblock Joint action: bodies and minds moving together.
\newblock \emph{Trends in Cognitive Sciences}, 10\penalty0 (2):\penalty0
  70--76, 2006.

\bibitem[Raibert(1986)]{raibert1986legged}
Marc~H Raibert.
\newblock \emph{Legged robots that balance}.
\newblock MIT press, 1986.

\bibitem[Kuindersma et~al.(2016)Kuindersma, Deits, Fallon, Valenzuela, Dai,
  Permenter, Koolen, Marion, and Tedrake]{kuindersma2016optimization}
Scott Kuindersma, Robin Deits, Maurice Fallon, Andr{\'e}s Valenzuela, Hongkai
  Dai, Frank Permenter, Twan Koolen, Pat Marion, and Russ Tedrake.
\newblock Optimization-based locomotion planning, estimation, and control
  design for the atlas humanoid robot.
\newblock \emph{Autonomous Robots}, 40\penalty0 (3):\penalty0 429--455, 2016.

\bibitem[Sims(1994)]{sims1994evolving}
Karl Sims.
\newblock Evolving virtual creatures.
\newblock In \emph{Proceedings of the 21st Annual Conference on Computer
  Graphics and Interactive Techniques}, SIGGRAPH '94, page 15–22, New York,
  NY, USA, 1994. Association for Computing Machinery.
\newblock ISBN 0897916670.
\newblock \doi{10.1145/192161.192167}.
\newblock URL \url{https://doi.org/10.1145/192161.192167}.

\bibitem[Faloutsos et~al.(2001)Faloutsos, van~de Panne, and
  Terzopoulos]{Faloutsos2001}
Petros Faloutsos, Michiel van~de Panne, and Demetri Terzopoulos.
\newblock Composable controllers for physics-based character animation.
\newblock In \emph{Proceedings of the 28th Annual Conference on Computer
  Graphics and Interactive Techniques}. Association for Computing Machinery,
  2001.

\bibitem[Yin et~al.(2007)Yin, Loken, and Van~de Panne]{yin2007simbicon}
KangKang Yin, Kevin Loken, and Michiel Van~de Panne.
\newblock Simbicon: Simple biped locomotion control.
\newblock In \emph{ACM Transactions on Graphics (TOG)}, volume~26, page 105.
  ACM, 2007.

\bibitem[Coros et~al.(2010)Coros, Beaudoin, and Van~de
  Panne]{coros2010generalized}
Stelian Coros, Philippe Beaudoin, and Michiel Van~de Panne.
\newblock Generalized biped walking control.
\newblock In \emph{ACM Transactions on Graphics (TOG)}, volume~29, page 130.
  ACM, 2010.

\bibitem[Liu et~al.(2012)Liu, Yin, van~de Panne, and Guo]{liu2012terrain}
Libin Liu, KangKang Yin, Michiel van~de Panne, and Baining Guo.
\newblock Terrain runner: control, parameterization, composition, and planning
  for highly dynamic motions.
\newblock \emph{ACM Transactions on Graphics (TOG)}, 31\penalty0 (6):\penalty0
  154, 2012.

\bibitem[OpenAI(2018)]{OpenAI_dota}
OpenAI.
\newblock Open{AI} {F}ive.
\newblock \url{https://blog.openai.com/openai-five/}, 2018.

\bibitem[Vinyals et~al.(2019)Vinyals, Babuschkin, Czarnecki, Mathieu, Dudzik,
  Chung, Choi, Powell, Ewalds, Georgiev, Oh, Horgan, Kroiss, Danihelka, Huang,
  Sifre, Cai, Agapiou, Jaderberg, Vezhnevets, Leblond, Pohlen, Dalibard,
  Budden, Sulsky, Molloy, Paine, G{\"{u}}l{\c{c}}ehre, Wang, Pfaff, Wu, Ring,
  Yogatama, W{\"{u}}nsch, McKinney, Smith, Schaul, Lillicrap, Kavukcuoglu,
  Hassabis, Apps, and Silver]{VinyalsStarCraft}
Oriol Vinyals, Igor Babuschkin, Wojciech~M. Czarnecki, Micha{\"{e}}l Mathieu,
  Andrew Dudzik, Junyoung Chung, David~H. Choi, Richard Powell, Timo Ewalds,
  Petko Georgiev, Junhyuk Oh, Dan Horgan, Manuel Kroiss, Ivo Danihelka, Aja
  Huang, Laurent Sifre, Trevor Cai, John~P. Agapiou, Max Jaderberg,
  Alexander~Sasha Vezhnevets, R{\'{e}}mi Leblond, Tobias Pohlen, Valentin
  Dalibard, David Budden, Yury Sulsky, James Molloy, Tom~L. Paine, {\c{C}}aglar
  G{\"{u}}l{\c{c}}ehre, Ziyu Wang, Tobias Pfaff, Yuhuai Wu, Roman Ring, Dani
  Yogatama, Dario W{\"{u}}nsch, Katrina McKinney, Oliver Smith, Tom Schaul,
  Timothy~P. Lillicrap, Koray Kavukcuoglu, Demis Hassabis, Chris Apps, and
  David Silver.
\newblock Grandmaster level in {StarCraft} {II} using multi-agent reinforcement
  learning.
\newblock \emph{Nature}, 575\penalty0 (7782):\penalty0 350--354, 2019.
\newblock \doi{10.1038/s41586-019-1724-z}.
\newblock URL \url{https://doi.org/10.1038/s41586-019-1724-z}.

\bibitem[Jaderberg et~al.(2019)Jaderberg, Czarnecki, Dunning, Marris, Lever,
  Casta{\~n}eda, Beattie, Rabinowitz, Morcos, Ruderman, Sonnerat, Green,
  Deason, Leibo, Silver, Hassabis, Kavukcuoglu, and Graepel]{JaderbergCTF}
Max Jaderberg, Wojciech~M. Czarnecki, Iain Dunning, Luke Marris, Guy Lever,
  Antonio~Garcia Casta{\~n}eda, Charles Beattie, Neil~C. Rabinowitz, Ari~S.
  Morcos, Avraham Ruderman, Nicolas Sonnerat, Tim Green, Louise Deason, Joel~Z.
  Leibo, David Silver, Demis Hassabis, Koray Kavukcuoglu, and Thore Graepel.
\newblock Human-level performance in 3d multiplayer games with population-based
  reinforcement learning.
\newblock \emph{Science}, 364\penalty0 (6443):\penalty0 859--865, 2019.
\newblock ISSN 0036-8075.
\newblock \doi{10.1126/science.aau6249}.
\newblock URL \url{https://science.sciencemag.org/content/364/6443/859}.

\bibitem[Mordatch and Abbeel(2018)]{MordatchEmergence}
Igor Mordatch and Pieter Abbeel.
\newblock Emergence of grounded compositional language in multi-agent
  populations.
\newblock In Sheila~A. McIlraith and Kilian~Q. Weinberger, editors,
  \emph{Proceedings of the Thirty-Second {AAAI} Conference on Artificial
  Intelligence, (AAAI-18), the 30th innovative Applications of Artificial
  Intelligence (IAAI-18), and the 8th {AAAI} Symposium on Educational Advances
  in Artificial Intelligence (EAAI-18), New Orleans, Louisiana, USA, February
  2-7, 2018}, pages 1495--1502. {AAAI} Press, 2018.
\newblock URL
  \url{https://www.aaai.org/ocs/index.php/AAAI/AAAI18/paper/view/17007}.

\bibitem[Heess et~al.(2017)Heess, Tirumala, Sriram, Lemmon, Merel, Wayne,
  Tassa, Erez, Wang, Eslami, Riedmiller, et~al.]{heess2017emergence}
Nicolas Heess, Dhruva Tirumala, Srinivasan Sriram, Jay Lemmon, Josh Merel, Greg
  Wayne, Yuval Tassa, Tom Erez, Ziyu Wang, Ali Eslami, Martin Riedmiller,
  et~al.
\newblock Emergence of locomotion behaviours in rich environments.
\newblock \emph{arXiv preprint arXiv:1707.02286}, 2017.

\bibitem[Bansal et~al.(2018)Bansal, Pachocki, Sidor, Sutskever, and
  Mordatch]{BansalEmergent}
Trapit Bansal, Jakub Pachocki, Szymon Sidor, Ilya Sutskever, and Igor Mordatch.
\newblock Emergent complexity via multi-agent competition.
\newblock In \emph{6th International Conference on Learning Representations,
  {ICLR} 2018, Vancouver, BC, Canada, April 30 - May 3, 2018, Conference Track
  Proceedings}. OpenReview.net, 2018.
\newblock URL \url{https://openreview.net/forum?id=Sy0GnUxCb}.

\bibitem[Peng et~al.(2018)Peng, Abbeel, Levine, and van~de
  Panne]{peng2018deepmimic}
Xue~Bin Peng, Pieter Abbeel, Sergey Levine, and Michiel van~de Panne.
\newblock {DeepMimic: Example-guided deep reinforcement learning of
  physics-based character skills}.
\newblock \emph{ACM Transactions on Graphics (TOG)}, 37\penalty0 (4):\penalty0
  143, 2018.

\bibitem[Merel et~al.(2020)Merel, Tunyasuvunakool, Ahuja, Tassa, Hasenclever,
  Pham, Erez, Wayne, and Heess]{merel2020}
Josh Merel, Saran Tunyasuvunakool, Arun Ahuja, Yuval Tassa, Leonard
  Hasenclever, Vu~Pham, Tom Erez, Greg Wayne, and Nicolas Heess.
\newblock {Catch \& Carry: Reusable Neural Controllers for Vision-Guided
  Whole-Body Tasks}.
\newblock \emph{ACM Transactions on Graphics (TOG)}, 39\penalty0 (4), 2020.

\bibitem[Lee et~al.(2020)Lee, Hwangbo, Wellhausen, Koltun, and
  Hutter]{lee2020learning}
Joonho Lee, Jemin Hwangbo, Lorenz Wellhausen, Vladlen Koltun, and Marco Hutter.
\newblock Learning quadrupedal locomotion over challenging terrain.
\newblock \emph{Science Robotics}, 5\penalty0 (47), 2020.
\newblock \doi{10.1126/scirobotics.abc5986}.
\newblock URL \url{https://robotics.sciencemag.org/content/5/47/eabc5986}.

\bibitem[OpenAI et~al.(2019)OpenAI, Akkaya, Andrychowicz, Chociej, Litwin,
  McGrew, Petron, Paino, Plappert, Powell, Ribas, Schneider, Tezak, Tworek,
  Welinder, Weng, Yuan, Zaremba, and Zhang]{openai2019solving}
OpenAI, Ilge Akkaya, Marcin Andrychowicz, Maciek Chociej, Mateusz Litwin, Bob
  McGrew, Arthur Petron, Alex Paino, Matthias Plappert, Glenn Powell, Raphael
  Ribas, Jonas Schneider, Nikolas Tezak, Jerry Tworek, Peter Welinder, Lilian
  Weng, Qiming Yuan, Wojciech Zaremba, and Lei Zhang.
\newblock Solving {R}ubik's cube with a robot hand.
\newblock \emph{CoRR}, abs/1910.07113, 2019.
\newblock URL \url{http://arxiv.org/abs/1910.07113}.

\bibitem[Peng et~al.(2020)Peng, Coumans, Zhang, Lee, Tan, and
  Levine]{peng2020learning}
Xue~Bin Peng, Erwin Coumans, Tingnan Zhang, Tsang-Wei Lee, Jie Tan, and Sergey
  Levine.
\newblock Learning agile robotic locomotion skills by imitating animals, 2020.

\bibitem[Brooks(1986)]{brooks1986robust}
Rodney Brooks.
\newblock A robust layered control system for a mobile robot.
\newblock \emph{IEEE Journal on Robotics and Automation}, 2\penalty0
  (1):\penalty0 14--23, 1986.

\bibitem[Albus(1993)]{albus1993reference}
James~S. Albus.
\newblock \emph{A Reference Model Architecture for Intelligent Systems Design},
  page 27–56.
\newblock Kluwer Academic Publishers, USA, 1993.
\newblock ISBN 0792392671.

\bibitem[Stone(2000)]{stone2000layered}
Peter Stone.
\newblock \emph{Layered learning in multiagent systems - a winning approach to
  robotic soccer}.
\newblock Intelligent robotics and autonomous agents. {MIT} Press, 2000.

\bibitem[Gupta et~al.(2019)Gupta, Kumar, Lynch, Levine, and
  Hausman]{gupta2019relay}
Abhishek Gupta, Vikash Kumar, Corey Lynch, Sergey Levine, and Karol Hausman.
\newblock Relay policy learning: Solving long-horizon tasks via imitation and
  reinforcement learning.
\newblock \emph{arXiv preprint arXiv:1910.11956}, 2019.

\bibitem[Merel et~al.(2019{\natexlab{b}})Merel, Ahuja, Pham, Tunyasuvunakool,
  Liu, Tirumala, Heess, and Wayne]{merel2018hierarchical}
Josh Merel, Arun Ahuja, Vu~Pham, Saran Tunyasuvunakool, Siqi Liu, Dhruva
  Tirumala, Nicolas Heess, and Greg Wayne.
\newblock Hierarchical visuomotor control of humanoids.
\newblock In \emph{7th International Conference on Learning Representations,
  {ICLR} 2019, New Orleans, LA, USA, May 6-9, 2019}. OpenReview.net,
  2019{\natexlab{b}}.

\bibitem[Peng et~al.(2019)Peng, Chang, Zhang, Abbeel, and Levine]{peng2019mcp}
Xue~Bin Peng, Michael Chang, Grace Zhang, Pieter Abbeel, and Sergey Levine.
\newblock {MCP:} learning composable hierarchical control with multiplicative
  compositional policies.
\newblock In Hanna~M. Wallach, Hugo Larochelle, Alina Beygelzimer, Florence
  d'Alch{\'{e}}{-}Buc, Emily~B. Fox, and Roman Garnett, editors, \emph{Advances
  in Neural Information Processing Systems 32: Annual Conference on Neural
  Information Processing Systems 2019, NeurIPS 2019, December 8-14, 2019,
  Vancouver, BC, Canada}, pages 3681--3692, 2019.

\bibitem[Merel et~al.(2019{\natexlab{c}})Merel, Hasenclever, Galashov, Ahuja,
  Pham, Wayne, Teh, and Heess]{MerelNPMP}
Josh Merel, Leonard Hasenclever, Alexandre Galashov, Arun Ahuja, Vu~Pham, Greg
  Wayne, Yee~Whye Teh, and Nicolas Heess.
\newblock Neural probabilistic motor primitives for humanoid control.
\newblock In \emph{7th International Conference on Learning Representations,
  {ICLR} 2019, New Orleans, LA, USA, May 6-9, 2019}. OpenReview.net,
  2019{\natexlab{c}}.
\newblock URL \url{https://openreview.net/forum?id=BJl6TjRcY7}.

\bibitem[Mnih et~al.(2015)Mnih, Kavukcuoglu, Silver, Rusu, Veness, Bellemare,
  Graves, Riedmiller, Fidjeland, Ostrovski, Petersen, Beattie, Sadik,
  Antonoglou, King, Kumaran, Wierstra, Legg, and Hassabis]{MnihDQN}
Volodymyr Mnih, Koray Kavukcuoglu, David Silver, Andrei~A. Rusu, Joel Veness,
  Marc~G. Bellemare, Alex Graves, Martin~A. Riedmiller, Andreas Fidjeland,
  Georg Ostrovski, Stig Petersen, Charles Beattie, Amir Sadik, Ioannis
  Antonoglou, Helen King, Dharshan Kumaran, Daan Wierstra, Shane Legg, and
  Demis Hassabis.
\newblock Human-level control through deep reinforcement learning.
\newblock \emph{Nature}, 518\penalty0 (7540):\penalty0 529--533, 2015.

\bibitem[Mnih et~al.(2016)Mnih, Badia, Mirza, Graves, Lillicrap, Harley,
  Silver, and Kavukcuoglu]{MnihA3C}
Volodymyr Mnih, Adri{\`{a}}~Puigdom{\`{e}}nech Badia, Mehdi Mirza, Alex Graves,
  Timothy~P. Lillicrap, Tim Harley, David Silver, and Koray Kavukcuoglu.
\newblock Asynchronous methods for deep reinforcement learning.
\newblock In \emph{Proceedings of the 33nd International Conference on Machine
  Learning, {ICML} 2016, New York City, NY, USA, June 19-24, 2016}, pages
  1928--1937, 2016.

\bibitem[Schulman et~al.(2017)Schulman, Wolski, Dhariwal, Radford, and
  Klimov]{SchulmanPPO}
John Schulman, Filip Wolski, Prafulla Dhariwal, Alec Radford, and Oleg Klimov.
\newblock Proximal policy optimization algorithms.
\newblock \emph{CoRR}, abs/1707.06347, 2017.

\bibitem[Heess et~al.(2015)Heess, Wayne, Silver, Lillicrap, Erez, and
  Tassa]{heess2015learning}
Nicolas Heess, Gregory Wayne, David Silver, Tim Lillicrap, Tom Erez, and Yuval
  Tassa.
\newblock Learning continuous control policies by stochastic value gradients.
\newblock In \emph{Advances in Neural Information Processing Systems}, 2015.

\bibitem[Lillicrap et~al.(2015)Lillicrap, Hunt, Pritzel, Heess, Erez, Tassa,
  Silver, and Wierstra]{LillicrapDDPG}
Timothy~P. Lillicrap, Jonathan~J. Hunt, Alexander Pritzel, Nicolas Heess, Tom
  Erez, Yuval Tassa, David Silver, and Daan Wierstra.
\newblock Continuous control with deep reinforcement learning.
\newblock \emph{CoRR}, abs/1509.02971, 2015.
\newblock URL \url{http://arxiv.org/abs/1509.02971}.

\bibitem[Abdolmaleki et~al.(2018)Abdolmaleki, Springenberg, Tassa, Munos,
  Heess, and Riedmiller]{AbdolmalekiMPO}
Abbas Abdolmaleki, Jost~Tobias Springenberg, Yuval Tassa, R{\'{e}}mi Munos,
  Nicolas Heess, and Martin~A. Riedmiller.
\newblock Maximum a posteriori policy optimisation.
\newblock In \emph{6th International Conference on Learning Representations,
  {ICLR} 2018, Vancouver, BC, Canada, April 30 - May 3, 2018, Conference Track
  Proceedings}, 2018.

\bibitem[Kitano et~al.(1997)Kitano, Asada, Kuniyoshi, Noda, and
  Osawa]{KitanoRoboCup}
Hiroaki Kitano, Minoru Asada, Yasuo Kuniyoshi, Itsuki Noda, and Eiichi Osawa.
\newblock Robo{C}up: The robot {W}orld {C}up initiative.
\newblock In \emph{Agents}, pages 340--347, 1997.

\bibitem[{RoboCup Federation}()]{RoboCupWeb}
{RoboCup Federation}.
\newblock https://www.robocup.org/.

\bibitem[Tassa et~al.(2020)Tassa, Tunyasuvunakool, Muldal, Doron, Liu, Bohez,
  Merel, Erez, Lillicrap, and Heess]{tassa2020dm_control}
Yuval Tassa, Saran Tunyasuvunakool, Alistair Muldal, Yotam Doron, Siqi Liu,
  Steven Bohez, Josh Merel, Tom Erez, Timothy Lillicrap, and Nicolas Heess.
\newblock dm\_control: Software and tasks for continuous control.
\newblock \emph{arXiv preprint arXiv:2006.12983}, 2020.

\bibitem[Ericsson et~al.(1993)Ericsson, Krampe, and
  Tesch-R{\"o}mer]{ericsson1993role}
K~Anders Ericsson, Ralf~T Krampe, and Clemens Tesch-R{\"o}mer.
\newblock The role of deliberate practice in the acquisition of expert
  performance.
\newblock \emph{Psychological review}, 100\penalty0 (3):\penalty0 363, 1993.

\bibitem[Baker and Young(2014)]{baker201420years}
J.~Baker and B.~Young.
\newblock 20 years later: deliberate practice and the development of expertise
  in sport.
\newblock \emph{International Review of Sport and Exercise Psychology},
  7:\penalty0 135 -- 157, 2014.

\bibitem[Ashford et~al.(2006)Ashford, Bennett, and
  Davids]{ashford2006observational}
Derek Ashford, Simon~J. Bennett, and Keith Davids.
\newblock Observational modeling effects for movement dynamics and movement
  outcome measures across differing task constraints: A meta-analysis.
\newblock \emph{Journal of Motor Behavior}, 38\penalty0 (3):\penalty0 185--205,
  2006.
\newblock \doi{10.3200/JMBR.38.3.185-205}.
\newblock URL \url{https://doi.org/10.3200/JMBR.38.3.185-205}.
\newblock PMID: 16709559.

\bibitem[Diedrichsen and Kornysheva(2015)]{diedrichsen2015motor}
J{\"o}rn Diedrichsen and Katja Kornysheva.
\newblock Motor skill learning between selection and execution.
\newblock \emph{Trends in cognitive sciences}, 19\penalty0 (4):\penalty0
  227--233, 2015.

\bibitem[Liu et~al.(2019)Liu, Lever, Merel, Tunyasuvunakool, Heess, and
  Graepel]{LiuSoccer}
Siqi Liu, Guy Lever, Josh Merel, Saran Tunyasuvunakool, Nicolas Heess, and
  Thore Graepel.
\newblock Emergent coordination through competition.
\newblock \emph{ICLR}, 2019.

\bibitem[Tuyls et~al.(2020)Tuyls, Omidshafiei, Muller, Wang, Connor, Hennes,
  Graham, Spearman, Waskett, Steele, Luc, Recasens, Galashov, Thornton, Elie,
  Sprechmann, Moreno, Cao, Garnelo, Dutta, Valko, Heess, Bridgland, Perolat,
  Vylder, Eslami, Rowland, Jaegle, Munos, Back, Ahamed, Bouton, Beauguerlange,
  Broshear, Graepel, and Hassabis]{tuyls2020game}
Karl Tuyls, Shayegan Omidshafiei, Paul Muller, Zhe Wang, Jerome Connor, Daniel
  Hennes, Ian Graham, William Spearman, Tim Waskett, Dafydd Steele, Pauline
  Luc, Adria Recasens, Alexandre Galashov, Gregory Thornton, Romuald Elie,
  Pablo Sprechmann, Pol Moreno, Kris Cao, Marta Garnelo, Praneet Dutta, Michal
  Valko, Nicolas Heess, Alex Bridgland, Julien Perolat, Bart~De Vylder, Ali
  Eslami, Mark Rowland, Andrew Jaegle, Remi Munos, Trevor Back, Razia Ahamed,
  Simon Bouton, Nathalie Beauguerlange, Jackson Broshear, Thore Graepel, and
  Demis Hassabis.
\newblock Game plan: What {AI} can do for football, and what football can do
  for {AI}, 2020.

\bibitem[Spearman(2018)]{spearman2018beyond}
William Spearman.
\newblock Beyond expected goals.
\newblock In \emph{Proceeding of the 12th {MIT} {S}loan {S}ports {A}nalytics
  {C}onference}, 2018.

\bibitem[Gon{\c{c}}alves et~al.(2015)Gon{\c{c}}alves, Gonzaga, Cardoso, and
  Teoldo]{gonccalves2015anticipation}
Eder Gon{\c{c}}alves, Adeilton dos~Santos Gonzaga, Felippe da Silva~Leite
  Cardoso, and Israel Teoldo.
\newblock Anticipation in soccer: a systematic review.
\newblock 2015.

\bibitem[Roca et~al.(2012)Roca, Williams, and Ford]{roca2012developmental}
Andr{\'e} Roca, A~Mark Williams, and Paul~R Ford.
\newblock Developmental activities and the acquisition of superior anticipation
  and decision making in soccer players.
\newblock \emph{Journal of sports sciences}, 30\penalty0 (15):\penalty0
  1643--1652, 2012.

\bibitem[Todorov et~al.(2012)Todorov, Erez, and Tassa]{TodorovMujoco}
Emanuel Todorov, Tom Erez, and Yuval Tassa.
\newblock Mujoco: {A} physics engine for model-based control.
\newblock In \emph{2012 {IEEE/RSJ} International Conference on Intelligent
  Robots and Systems, {IROS} 2012, Vilamoura, Algarve, Portugal, October 7-12,
  2012}, pages 5026--5033, 2012.

\bibitem[Heess et~al.(2016)Heess, Wayne, Tassa, Lillicrap, Riedmiller, and
  Silver]{heess2016learning}
Nicolas Heess, Greg Wayne, Yuval Tassa, Timothy Lillicrap, Martin Riedmiller,
  and David Silver.
\newblock Learning and transfer of modulated locomotor controllers.
\newblock \emph{arXiv preprint arXiv:1610.05182}, 2016.

\bibitem[Brockman et~al.(2016)Brockman, Cheung, Pettersson, Schneider,
  Schulman, Tang, and Zaremba]{OpenAIGym}
Greg Brockman, Vicki Cheung, Ludwig Pettersson, Jonas Schneider, John Schulman,
  Jie Tang, and Wojciech Zaremba.
\newblock Open{AI} gym.
\newblock \emph{CoRR}, abs/1606.01540, 2016.

\bibitem[Tassa et~al.(2018)Tassa, Doron, Muldal, Erez, Li, de~Las~Casas,
  Budden, Abdolmaleki, Merel, Lefrancq, Lillicrap, and
  Riedmiller]{TassaControlSuite}
Yuval Tassa, Yotam Doron, Alistair Muldal, Tom Erez, Yazhe Li, Diego
  de~Las~Casas, David Budden, Abbas Abdolmaleki, Josh Merel, Andrew Lefrancq,
  Timothy~P. Lillicrap, and Martin~A. Riedmiller.
\newblock Deepmind control suite.
\newblock \emph{CoRR}, abs/1801.00690, 2018.

\bibitem[Riedmiller et~al.(2018)Riedmiller, Hafner, Lampe, Neunert, Degrave,
  van~de Wiele, Mnih, Heess, and Springenberg]{riedmiller2018learning}
Martin Riedmiller, Roland Hafner, Thomas Lampe, Michael Neunert, Jonas Degrave,
  Tom van~de Wiele, Vlad Mnih, Nicolas Heess, and Jost~Tobias Springenberg.
\newblock Learning by playing - solving sparse reward tasks from scratch.
\newblock In \emph{Proceedings of the 35th International Conference on Machine
  Learning}, 2018.

\bibitem[Kurach et~al.(2019)Kurach, Raichuk, Stanczyk, Zajac, Bachem, Espeholt,
  Riquelme, Vincent, Michalski, Bousquet, and Gelly]{kurach2019google}
Karol Kurach, Anton Raichuk, Piotr Stanczyk, Michal Zajac, Olivier Bachem,
  Lasse Espeholt, Carlos Riquelme, Damien Vincent, Marcin Michalski, Olivier
  Bousquet, and Sylvain Gelly.
\newblock Google research football: {A} novel reinforcement learning
  environment.
\newblock \emph{CoRR}, abs/1907.11180, 2019.
\newblock URL \url{http://arxiv.org/abs/1907.11180}.

\bibitem[Shapley(1953)]{shapley1953stochastic}
Lloyd~S Shapley.
\newblock Stochastic games.
\newblock \emph{Proceedings of the National Academy of Sciences}, 39\penalty0
  (10):\penalty0 1095--1100, 1953.

\bibitem[Jaderberg et~al.(2017)Jaderberg, Dalibard, Osindero, Czarnecki,
  Donahue, Razavi, Vinyals, Green, Dunning, Simonyan,
  et~al.]{jaderberg2017population}
Max Jaderberg, Valentin Dalibard, Simon Osindero, Wojciech~M Czarnecki, Jeff
  Donahue, Ali Razavi, Oriol Vinyals, Tim Green, Iain Dunning, Karen Simonyan,
  et~al.
\newblock Population based training of neural networks.
\newblock \emph{arXiv preprint arXiv:1711.09846}, 2017.

\bibitem[Hochreiter and Schmidhuber(1997)]{HochreiterLSTM}
Sepp Hochreiter and J{\"{u}}rgen Schmidhuber.
\newblock Long short-term memory.
\newblock \emph{Neural Comput.}, 9\penalty0 (8):\penalty0 1735--1780, 1997.

\bibitem[Hasenclever et~al.(2020)Hasenclever, Pardo, Hadsell, Heess, and
  Merel]{Hasenclever2020}
Leonard Hasenclever, Fabio Pardo, Raia Hadsell, Nicolas Heess, and Josh Merel.
\newblock {CoMic: C}omplementary task learning \& mimicry for reusable skills.
\newblock In \emph{Proceedings of the International Conference on Machine
  Learning (ICML)}, 2020.

\bibitem[Galashov et~al.(2019)Galashov, Jayakumar, Hasenclever, Tirumala,
  Schwarz, Desjardins, Czarnecki, Teh, Pascanu, and
  Heess]{galashov2018information}
Alexandre Galashov, Siddhant Jayakumar, Leonard Hasenclever, Dhruva Tirumala,
  Jonathan Schwarz, Guillaume Desjardins, Wojtek~M. Czarnecki, Yee~Whye Teh,
  Razvan Pascanu, and Nicolas Heess.
\newblock Information asymmetry in {KL}-regularized {RL}.
\newblock In \emph{International Conference on Learning Representations}, 2019.

\bibitem[Tirumala et~al.(2020)Tirumala, Galashov, Noh, Hasenclever, Pascanu,
  Schwarz, Desjardins, Czarnecki, Ahuja, Teh, and Heess]{tirumala2020behavior}
Dhruva Tirumala, Alexandre Galashov, Hyeonwoo Noh, Leonard Hasenclever, Razvan
  Pascanu, Jonathan Schwarz, Guillaume Desjardins, Wojciech~Marian Czarnecki,
  Arun Ahuja, Yee~Whye Teh, and Nicolas Heess.
\newblock Behavior priors for efficient reinforcement learning, 2020.

\bibitem[Leibo et~al.(2019)Leibo, Hughes, Lanctot, and Graepel]{LeiboManifesto}
Joel~Z. Leibo, Edward Hughes, Marc Lanctot, and Thore Graepel.
\newblock Autocurricula and the emergence of innovation from social
  interaction: {A} manifesto for multi-agent intelligence research.
\newblock \emph{CoRR}, abs/1903.00742, 2019.
\newblock URL \url{http://arxiv.org/abs/1903.00742}.

\bibitem[Balduzzi et~al.(2018)Balduzzi, Tuyls, P{\'{e}}rolat, and
  Graepel]{balduzzi2018re}
David Balduzzi, Karl Tuyls, Julien P{\'{e}}rolat, and Thore Graepel.
\newblock Re-evaluating evaluation.
\newblock In Samy Bengio, Hanna~M. Wallach, Hugo Larochelle, Kristen Grauman,
  Nicol{\`{o}} Cesa{-}Bianchi, and Roman Garnett, editors, \emph{Advances in
  Neural Information Processing Systems 31: Annual Conference on Neural
  Information Processing Systems 2018, NeurIPS 2018, December 3-8, 2018,
  Montr{\'{e}}al, Canada}, pages 3272--3283, 2018.
\newblock URL
  \url{https://proceedings.neurips.cc/paper/2018/hash/cdf1035c34ec380218a8cc9a43d438f9-Abstract.html}.

\bibitem[Teh et~al.(2017)Teh, Bapst, Czarnecki, Quan, Kirkpatrick, Hadsell,
  Heess, and Pascanu]{teh2017distral}
Yee Teh, Victor Bapst, Wojciech~M Czarnecki, John Quan, James Kirkpatrick, Raia
  Hadsell, Nicolas Heess, and Razvan Pascanu.
\newblock Distral: Robust multitask reinforcement learning.
\newblock In \emph{Advances in Neural Information Processing Systems}, 2017.

\bibitem[Czarnecki et~al.(2020)Czarnecki, Gidel, Tracey, Tuyls, Omidshafiei,
  Balduzzi, and Jaderberg]{Czarnecki2020Spinning}
Wojciech~M. Czarnecki, Gauthier Gidel, Brendan Tracey, Karl Tuyls, Shayegan
  Omidshafiei, David Balduzzi, and Max Jaderberg.
\newblock Real world games look like spinning tops.
\newblock In Hugo Larochelle, Marc'Aurelio Ranzato, Raia Hadsell,
  Maria{-}Florina Balcan, and Hsuan{-}Tien Lin, editors, \emph{Advances in
  Neural Information Processing Systems 33: Annual Conference on Neural
  Information Processing Systems 2020, NeurIPS 2020, December 6-12, 2020,
  virtual}, 2020.

\bibitem[Silver et~al.(2016)Silver, Huang, Maddison, Guez, Sifre, van~den
  Driessche, Schrittwieser, Antonoglou, Panneershelvam, Lanctot, Dieleman,
  Grewe, Nham, Kalchbrenner, Sutskever, Lillicrap, Leach, Kavukcuoglu, Graepel,
  and Hassabis]{SilverAlphaGo}
David Silver, Aja Huang, Chris~J. Maddison, Arthur Guez, Laurent Sifre, George
  van~den Driessche, Julian Schrittwieser, Ioannis Antonoglou, Vedavyas
  Panneershelvam, Marc Lanctot, Sander Dieleman, Dominik Grewe, John Nham, Nal
  Kalchbrenner, Ilya Sutskever, Timothy~P. Lillicrap, Madeleine Leach, Koray
  Kavukcuoglu, Thore Graepel, and Demis Hassabis.
\newblock Mastering the game of go with deep neural networks and tree search.
\newblock \emph{Nature}, 529\penalty0 (7587):\penalty0 484--489, 2016.

\bibitem[Elo(1978)]{elo1978rating}
Arpad~E. Elo.
\newblock \emph{The Rating of Chessplayers, Past and Present}.
\newblock Arco, New York, 1978.

\bibitem[Muraskin and Sherwin(2015)]{muraskin2015knowingswing}
Jordan Muraskin and Jason Sherwin.
\newblock Knowing when not to swing: Eeg evidence that enhanced
  perception-action coupling underlies baseball batter expertise.
\newblock \emph{Neuroimage}, 123:1-10, 2015.
\newblock URL \url{https://doi.org/10.1016/j.neuroimage.2015.08.028}.

\bibitem[Quiroga et~al.(2005)Quiroga, Reddy, Kreiman, Koch, and
  Fried]{quiroga2005invariant}
Quian~R. Quiroga, L.~Reddy, G.~Kreiman, C.~Koch, and I.~Fried.
\newblock Invariant visual representation by single neurons in the human brain.
\newblock \emph{Nature}, pages 1102--1107, 2005.

\bibitem[Quiroga et~al.(2008)Quiroga, Kreiman, Koch, and
  Fried]{quiroga2008sparse}
R.~Quian Quiroga, G.~Kreiman, C.~Koch, and I.~Fried.
\newblock Sparse but not ‘grandmother-cell’ coding in the medial temporal
  lobe.
\newblock \emph{Trends in Cognitive Sciences}, 12\penalty0 (3):\penalty0
  87--91, 2008.
\newblock ISSN 1364-6613.
\newblock \doi{https://doi.org/10.1016/j.tics.2007.12.003}.
\newblock URL
  \url{https://www.sciencedirect.com/science/article/pii/S1364661308000235}.

\bibitem[Nachum et~al.(2018)Nachum, Gu, Lee, and Levine]{nachum2018data}
Ofir Nachum, Shixiang~(Shane) Gu, Honglak Lee, and Sergey Levine.
\newblock Data-efficient hierarchical reinforcement learning.
\newblock In S.~Bengio, H.~Wallach, H.~Larochelle, K.~Grauman, N.~Cesa-Bianchi,
  and R.~Garnett, editors, \emph{Advances in Neural Information Processing
  Systems 31}, pages 3303--3313. Curran Associates, Inc., 2018.
\newblock URL
  \url{http://papers.nips.cc/paper/7591-data-efficient-hierarchical-reinforcement-learning.pdf}.

\bibitem[Peng et~al.(2017)Peng, Berseth, Yin, and Van
  De~Panne]{peng2017deeploco}
Xue~Bin Peng, Glen Berseth, KangKang Yin, and Michiel Van De~Panne.
\newblock {DeepLoco: Dynamic locomotion skills using hierarchical deep
  reinforcement learning}.
\newblock \emph{ACM Transactions on Graphics (TOG)}, 36\penalty0 (4):\penalty0
  41, 2017.

\bibitem[Liu and Hodgins(2018)]{liu2018learning}
Libin Liu and Jessica Hodgins.
\newblock Learning basketball dribbling skills using trajectory optimization
  and deep reinforcement learning.
\newblock \emph{ACM Transactions on Graphics (TOG)}, 37\penalty0 (4):\penalty0
  142, 2018.

\bibitem[Chentanez et~al.(2018)Chentanez, M{\"u}ller, Macklin, Makoviychuk, and
  Jeschke]{chentanez2018physics}
Nuttapong Chentanez, Matthias M{\"u}ller, Miles Macklin, Viktor Makoviychuk,
  and Stefan Jeschke.
\newblock Physics-based motion capture imitation with deep reinforcement
  learning.
\newblock In \emph{Proceedings of the 11th Annual International Conference on
  Motion, Interaction, and Games}, pages 1--10, 2018.

\bibitem[Lee et~al.(2019)Lee, Park, Lee, and Lee]{lee2019scalable}
Seunghwan Lee, Moonseok Park, Kyoungmin Lee, and Jehee Lee.
\newblock Scalable muscle-actuated human simulation and control.
\newblock \emph{ACM Transactions on Graphics (TOG)}, 38\penalty0 (4):\penalty0
  73, 2019.

\bibitem[Park et~al.(2019)Park, Ryu, Lee, Lee, and Lee]{park2019learning}
Soohwan Park, Hoseok Ryu, Seyoung Lee, Sunmin Lee, and Jehee Lee.
\newblock Learning predict-and-simulate policies from unorganized human motion
  data.
\newblock \emph{ACM Transactions on Graphics (TOG)}, 38\penalty0 (6):\penalty0
  205, 2019.

\bibitem[Bergamin et~al.(2019)Bergamin, Clavet, Holden, and
  Forbes]{bergamin2019drecon}
Kevin Bergamin, Simon Clavet, Daniel Holden, and James~Richard Forbes.
\newblock Drecon: data-driven responsive control of physics-based characters.
\newblock \emph{ACM Transactions on Graphics (TOG)}, 38\penalty0 (6):\penalty0
  206, 2019.

\bibitem[Tassa et~al.(2012)Tassa, Erez, and Todorov]{tassa2012synthesis}
Yuval Tassa, Tom Erez, and Emanuel Todorov.
\newblock Synthesis and stabilization of complex behaviors through online
  trajectory optimization.
\newblock In \emph{2012 IEEE/RSJ International Conference on Intelligent Robots
  and Systems}, pages 4906--4913. IEEE, 2012.

\bibitem[Mordatch et~al.(2015)Mordatch, Lowrey, Andrew, Popovic, and
  Todorov]{mordatch2015interactive}
Igor Mordatch, Kendall Lowrey, Galen Andrew, Zoran Popovic, and Emanuel~V
  Todorov.
\newblock Interactive control of diverse complex characters with neural
  networks.
\newblock In \emph{Advances in Neural Information Processing Systems}, pages
  3132--3140, 2015.

\bibitem[Schulman et~al.(2015)Schulman, Moritz, Levine, Jordan, and
  Abbeel]{schulman2015high}
John Schulman, Philipp Moritz, Sergey Levine, Michael Jordan, and Pieter
  Abbeel.
\newblock High-dimensional continuous control using generalized advantage
  estimation.
\newblock \emph{arXiv preprint arXiv:1506.02438}, 2015.

\bibitem[Merel et~al.(2017)Merel, Tassa, TB, Srinivasan, Lemmon, Wang, Wayne,
  and Heess]{merel2017learning}
Josh Merel, Yuval Tassa, Dhruva TB, Sriram Srinivasan, Jay Lemmon, Ziyu Wang,
  Greg Wayne, and Nicolas Heess.
\newblock Learning human behaviors from motion capture by adversarial
  imitation.
\newblock \emph{arXiv preprint arXiv:1707.02201}, 2017.

\bibitem[Chao et~al.(2019)Chao, Yang, Chen, and Deng]{chao2019learning}
Yu-Wei Chao, Jimei Yang, Weifeng Chen, and Jia Deng.
\newblock Learning to sit: Synthesizing human-chair interactions via
  hierarchical control.
\newblock \emph{arXiv preprint arXiv:1908.07423}, 2019.

\bibitem[{Boston Dynamics}(2019)]{bostonDynamics}
{Boston Dynamics}.
\newblock {More Parkour Atlas}, 2019.
\newblock Available at \url{https://www.youtube.com/watch?v=_sBBaNYex3E}.

\bibitem[Hwangbo et~al.(2019)Hwangbo, Lee, Dosovitskiy, Bellicoso, Tsounis,
  Koltun, and Hutter]{hwangbo2019learning}
Jemin Hwangbo, Joonho Lee, Alexey Dosovitskiy, Dario Bellicoso, Vassilios
  Tsounis, Vladlen Koltun, and Marco Hutter.
\newblock Learning agile and dynamic motor skills for legged robots.
\newblock \emph{Science Robotics}, 4\penalty0 (26), 2019.
\newblock \doi{10.1126/scirobotics.aau5872}.
\newblock URL \url{https://robotics.sciencemag.org/content/4/26/eaau5872}.

\bibitem[Xie et~al.(2019)Xie, Clary, Dao, Morais, Hurst, and van~de
  Panne]{xie2019iterative}
Zhaoming Xie, Patrick Clary, Jeremy Dao, Pedro Morais, Jonathan~W. Hurst, and
  Michiel van~de Panne.
\newblock Iterative reinforcement learning based design of dynamic locomotion
  skills for {Cassie}.
\newblock \emph{CoRR}, abs/1903.09537, 2019.
\newblock URL \url{http://arxiv.org/abs/1903.09537}.

\bibitem[Panait and Luke(2005)]{PanaitL05}
Liviu Panait and Sean Luke.
\newblock Cooperative multi-agent learning: The state of the art.
\newblock \emph{Autonomous Agents and Multi-Agent Systems}, 11\penalty0
  (3):\penalty0 387--434, 2005.

\bibitem[Busoniu et~al.(2008)Busoniu, Babuska, and Schutter]{BusoniuBS08}
Lucian Busoniu, Robert Babuska, and Bart~De Schutter.
\newblock A comprehensive survey of multiagent reinforcement learning.
\newblock \emph{{IEEE} Trans. Systems, Man, and Cybernetics, Part {C}},
  38\penalty0 (2):\penalty0 156--172, 2008.

\bibitem[Tuyls and Weiss(2012)]{TuylsW12}
Karl Tuyls and Gerhard Weiss.
\newblock Multiagent learning: Basics, challenges, and prospects.
\newblock \emph{{AI} Magazine}, 33\penalty0 (3):\penalty0 41--52, 2012.

\bibitem[Hernandez{-}Leal et~al.(2019)Hernandez{-}Leal, Kartal, and
  Taylor]{Hernandez-LealK19}
Pablo Hernandez{-}Leal, Bilal Kartal, and Matthew~E. Taylor.
\newblock A survey and critique of multiagent deep reinforcement learning.
\newblock \emph{Autonomous Agents and Multi-Agent Systems}, 33\penalty0
  (6):\penalty0 750--797, 2019.

\bibitem[Yang and Wang(2020)]{yang2020overview}
Yaodong Yang and Jun Wang.
\newblock An overview of multi-agent reinforcement learning from game
  theoretical perspective, 2020.

\bibitem[Lauer and Riedmiller(2000)]{LauerDistributed}
Martin Lauer and Martin~A. Riedmiller.
\newblock An algorithm for distributed reinforcement learning in cooperative
  multi-agent systems.
\newblock In \emph{Proceedings of the Seventeenth International Conference on
  Machine Learning {(ICML} 2000), Stanford University, Stanford, CA, USA, June
  29 - July 2, 2000}, pages 535--542, 2000.

\bibitem[Lowe et~al.(2017)Lowe, Wu, Tamar, Harb, Abbeel, and
  Mordatch]{LoweMADDPG}
Ryan Lowe, Yi~Wu, Aviv Tamar, Jean Harb, Pieter Abbeel, and Igor Mordatch.
\newblock Multi-agent actor-critic for mixed cooperative-competitive
  environments.
\newblock In Isabelle Guyon, Ulrike von Luxburg, Samy Bengio, Hanna~M. Wallach,
  Rob Fergus, S.~V.~N. Vishwanathan, and Roman Garnett, editors, \emph{Advances
  in Neural Information Processing Systems 30: Annual Conference on Neural
  Information Processing Systems 2017, 4-9 December 2017, Long Beach, CA,
  {USA}}, pages 6379--6390, 2017.
\newblock URL
  \url{http://papers.nips.cc/paper/7217-multi-agent-actor-critic-for-mixed-cooperative-competitive-environments}.

\bibitem[Foerster et~al.(2018)Foerster, Farquhar, Afouras, Nardelli, and
  Whiteson]{FoersterCOMA}
Jakob~N. Foerster, Gregory Farquhar, Triantafyllos Afouras, Nantas Nardelli,
  and Shimon Whiteson.
\newblock Counterfactual multi-agent policy gradients.
\newblock In \emph{Proceedings of the Thirty-Second {AAAI} Conference on
  Artificial Intelligence, (AAAI-18), the 30th innovative Applications of
  Artificial Intelligence (IAAI-18), and the 8th {AAAI} Symposium on
  Educational Advances in Artificial Intelligence (EAAI-18), New Orleans,
  Louisiana, USA, February 2-7, 2018}, pages 2974--2982, 2018.
\newblock URL
  \url{https://www.aaai.org/ocs/index.php/AAAI/AAAI18/paper/view/17193}.

\bibitem[Laurent et~al.(2011)Laurent, Matignon, and
  Le~Fort-Piat]{LaurentIndepenedentLearners}
Guillaume~J. Laurent, La{\"e}titia Matignon, and Nadine Le~Fort-Piat.
\newblock {The world of Independent learners is not Markovian.}
\newblock \emph{{International Journal of Knowledge-Based and Intelligent
  Engineering Systems}}, 15\penalty0 (1):\penalty0 55--64, March 2011.
\newblock \doi{10.3233/KES-2010-0206}.

\bibitem[Matignon et~al.(2012)Matignon, Laurent, and
  Fort{-}Piat]{MatignonIndependentLearners}
La{\"{e}}titia Matignon, Guillaume~J. Laurent, and Nadine~Le Fort{-}Piat.
\newblock Independent reinforcement learners in cooperative {M}arkov games: a
  survey regarding coordination problems.
\newblock \emph{Knowledge Eng. Review}, 27\penalty0 (1):\penalty0 1--31, 2012.

\bibitem[Bernstein et~al.(2000)Bernstein, Zilberstein, and
  Immerman]{BernsteinDecPomdp}
Daniel~S. Bernstein, Shlomo Zilberstein, and Neil Immerman.
\newblock The complexity of decentralized control of {Markov Decision
  Processes}.
\newblock In \emph{{UAI} '00: Proceedings of the 16th Conference in Uncertainty
  in Artificial Intelligence, Stanford University, Stanford, California, USA,
  June 30 - July 3, 2000}, pages 32--37, 2000.

\bibitem[Claus and Boutilier(1998)]{ClausBoutillierDynamics}
Caroline Claus and Craig Boutilier.
\newblock The dynamics of reinforcement learning in cooperative multiagent
  systems.
\newblock In \emph{Proceedings of the Fifteenth National Conference on
  Artificial Intelligence and Tenth Innovative Applications of Artificial
  Intelligence Conference, {AAAI} 98, {IAAI} 98, July 26-30, 1998, Madison,
  Wisconsin, {USA.}}, pages 746--752, 1998.

\bibitem[Baker et~al.(2020)Baker, Kanitscheider, Markov, Wu, Powell, McGrew,
  and Mordatch]{BakerToolUse}
Bowen Baker, Ingmar Kanitscheider, Todor Markov, Yi~Wu, Glenn Powell, Bob
  McGrew, and Igor Mordatch.
\newblock Emergent tool use from multi-agent autocurricula.
\newblock In \emph{8th International Conference on Learning Representations,
  {ICLR} 2020, Addis Ababa, Ethiopia, April 26-30, 2020}. OpenReview.net, 2020.
\newblock URL \url{https://openreview.net/forum?id=SkxpxJBKwS}.

\bibitem[Sukhbaatar et~al.(2016)Sukhbaatar, Szlam, and Fergus]{SukhbaatarComms}
Sainbayar Sukhbaatar, Arthur Szlam, and Rob Fergus.
\newblock Learning multiagent communication with backpropagation.
\newblock In \emph{Advances in Neural Information Processing Systems 29: Annual
  Conference on Neural Information Processing Systems 2016, December 5-10,
  2016, Barcelona, Spain}, pages 2244--2252, 2016.
\newblock URL
  \url{http://papers.nips.cc/paper/6398-learning-multiagent-communication-with-backpropagation}.

\bibitem[Foerster et~al.(2016)Foerster, Assael, de~Freitas, and
  Whiteson]{FoersterDIAL}
Jakob Foerster, Ioannis~Alexandros Assael, Nando de~Freitas, and Shimon
  Whiteson.
\newblock Learning to communicate with deep multi-agent reinforcement learning.
\newblock In \emph{Advances in Neural Information Processing Systems}, pages
  2137--2145, 2016.

\bibitem[Samuel(1959)]{SamuelCheckers}
A.~L. Samuel.
\newblock Some studies in machine learning using the game of checkers.
\newblock \emph{IBM J. Res. Dev.}, 3\penalty0 (3):\penalty0 210--229, July
  1959.
\newblock ISSN 0018-8646.

\bibitem[Tesauro(1995)]{TesauroTDGammon}
G.~Tesauro.
\newblock Temporal difference learning and {TD}-gammon.
\newblock \emph{Commun. ACM}, 38\penalty0 (3):\penalty0 58--68, March 1995.

\bibitem[Campbell et~al.(2002)Campbell, Jr., and Hsu]{CampbellDeepBlue}
Murray Campbell, A.~Joseph~Hoane Jr., and Feng{-}hsiung Hsu.
\newblock Deep {B}lue.
\newblock \emph{Artif. Intell.}, 134\penalty0 (1-2):\penalty0 57--83, 2002.

\bibitem[Moravčík et~al.(2017)Moravčík, Schmid, Burch, Lisý, Morrill,
  Bard, Davis, Waugh, Johanson, and Bowling]{MoravcikDeepStack}
Matej Moravčík, Martin Schmid, Neil Burch, Viliam Lisý, Dustin Morrill,
  Nolan Bard, Trevor Davis, Kevin Waugh, Michael Johanson, and Michael Bowling.
\newblock Deepstack: Expert-level artificial intelligence in no-limit poker.
\newblock 356, 01 2017.

\bibitem[Dawkins et~al.(1979)Dawkins, Krebs, Maynard~Smith, and
  Holliday]{DawkinsArmsRaces}
Richard Dawkins, John~Richard Krebs, J.~Maynard~Smith, and Robin Holliday.
\newblock Arms races between and within species.
\newblock \emph{Proceedings of the Royal Society of London. Series B.
  Biological Sciences}, 205\penalty0 (1161):\penalty0 489--511, 1979.
\newblock \doi{10.1098/rspb.1979.0081}.
\newblock URL
  \url{https://royalsocietypublishing.org/doi/abs/10.1098/rspb.1979.0081}.

\bibitem[Al{-}Shedivat et~al.(2018)Al{-}Shedivat, Bansal, Burda, Sutskever,
  Mordatch, and Abbeel]{AlShedivatContinuous}
Maruan Al{-}Shedivat, Trapit Bansal, Yura Burda, Ilya Sutskever, Igor Mordatch,
  and Pieter Abbeel.
\newblock Continuous adaptation via meta-learning in nonstationary and
  competitive environments.
\newblock In \emph{6th International Conference on Learning Representations,
  {ICLR} 2018, Vancouver, BC, Canada, April 30 - May 3, 2018, Conference Track
  Proceedings}. OpenReview.net, 2018.
\newblock URL \url{https://openreview.net/forum?id=Sk2u1g-0-}.

\bibitem[Dayan and Hinton(1993)]{dayan1993feudal}
Peter Dayan and Geoffrey~E Hinton.
\newblock Feudal reinforcement learning.
\newblock In \emph{Advances in Neural Information Processing Systems}, 1993.

\bibitem[Schmidhuber(1991)]{Schmidhuber91neuralsequence}
Jürgen Schmidhuber.
\newblock Neural sequence chunkers.
\newblock Technical report, Institut für Informatik, Technische Universität
  München, 1991.

\bibitem[Wiering and Schmidhuber(1997)]{weiring1997hq}
Marco Wiering and Jürgen Schmidhuber.
\newblock Hq-learning.
\newblock \emph{Adaptive Behavior}, 6\penalty0 (2):\penalty0 219--246, 1997.
\newblock \doi{10.1177/105971239700600202}.
\newblock URL \url{https://doi.org/10.1177/105971239700600202}.

\bibitem[Parr and Russell(1998)]{parr1998reinforcement}
Ronald Parr and Stuart~J Russell.
\newblock Reinforcement learning with hierarchies of machines.
\newblock In \emph{Advances in Neural Information Processing Systems}, 1998.

\bibitem[Dietterich(1999)]{dietterich1999hierarchical}
Thomas~G. Dietterich.
\newblock Hierarchical reinforcement learning with the {MAXQ} value function
  decomposition, 1999.

\bibitem[Sutton et~al.(1999)Sutton, Precup, and Singh]{sutton1999between}
Richard~S. Sutton, Doina Precup, and Satinder Singh.
\newblock Between {MDP}s and semi-{MDP}s: A framework for temporal abstraction
  in reinforcement learning.
\newblock \emph{Artif. Intell.}, 112\penalty0 (1–2):\penalty0 181–211,
  August 1999.
\newblock ISSN 0004-3702.
\newblock \doi{10.1016/S0004-3702(99)00052-1}.
\newblock URL \url{https://doi.org/10.1016/S0004-3702(99)00052-1}.

\bibitem[Barto and Mahadevan(2002)]{barto2002}
Andrew Barto and Sridhar Mahadevan.
\newblock Recent advances in hierarchical reinforcement learning.
\newblock \emph{Discrete Event Dynamic Systems: Theory and Applications}, 13,
  12 2002.
\newblock \doi{10.1023/A:1025696116075}.

\bibitem[Paraschos et~al.(2013)Paraschos, Daniel, Peters, and
  Neumann]{paraschos2013probabilistic}
Alexandros Paraschos, Christian Daniel, Jan~R Peters, and Gerhard Neumann.
\newblock Probabilistic movement primitives.
\newblock In C.~J.~C. Burges, L.~Bottou, M.~Welling, Z.~Ghahramani, and K.~Q.
  Weinberger, editors, \emph{Advances in Neural Information Processing Systems
  26}, pages 2616--2624. Curran Associates, Inc., 2013.
\newblock URL
  \url{http://papers.nips.cc/paper/5177-probabilistic-movement-primitives.pdf}.

\bibitem[Daniel et~al.(2016)Daniel, Neumann, Kroemer, and
  Peters]{daniel2016hierarchical}
Christian Daniel, Gerhard Neumann, Oliver Kroemer, and Jan Peters.
\newblock Hierarchical relative entropy policy search.
\newblock \emph{Journal of Machine Learning Research}, 17\penalty0
  (93):\penalty0 1--50, 2016.
\newblock URL \url{http://jmlr.org/papers/v17/15-188.html}.

\bibitem[Kambhampati et~al.(1991)Kambhampati, Cutkosky, Tenenbaum, and
  Lee]{Kambhampati}
Subbarao Kambhampati, Mark Cutkosky, Marty Tenenbaum, and Soo~Hong Lee.
\newblock Combining specialized reasoners and general purpose planners: A case
  study.
\newblock In \emph{Proceedings of the Ninth National Conference on Artificial
  Intelligence - Volume 1}, AAAI'91, page 199–205. AAAI Press, 1991.
\newblock ISBN 0262510596.

\bibitem[Sharir(1989)]{Sharir1989}
Micha Sharir.
\newblock Algorithmic motion planning in robotics.
\newblock \emph{Computer}, 22\penalty0 (3):\penalty0 9–20, March 1989.
\newblock ISSN 0018-9162.
\newblock \doi{10.1109/2.16221}.
\newblock URL \url{https://doi.org/10.1109/2.16221}.

\bibitem[Vezhnevets et~al.(2017)Vezhnevets, Osindero, Schaul, Heess, Jaderberg,
  Silver, and Kavukcuoglu]{vezhnevets2017feudal}
Alexander~Sasha Vezhnevets, Simon Osindero, Tom Schaul, Nicolas Heess, Max
  Jaderberg, David Silver, and Koray Kavukcuoglu.
\newblock Feudal networks for hierarchical reinforcement learning.
\newblock In \emph{Proceedings of the 34th International Conference on Machine
  Learning}, 2017.

\bibitem[Nachum et~al.(2019)Nachum, Gu, Lee, and Levine]{nachum2018nearoptimal}
Ofir Nachum, Shixiang Gu, Honglak Lee, and Sergey Levine.
\newblock Near-optimal representation learning for hierarchical reinforcement
  learning.
\newblock In \emph{International Conference on Learning Representations}, 2019.

\bibitem[Gregor et~al.(2017)Gregor, Rezende, and
  Wierstra]{gregor2016variational}
Karol Gregor, Danilo~Jimenez Rezende, and Daan Wierstra.
\newblock Variational intrinsic control.
\newblock In \emph{International Conference on Learning Representations}, 2017.

\bibitem[Hausman et~al.(2018)Hausman, Springenberg, Wang, Heess, and
  Riedmiller]{hausman2018learning}
Karol Hausman, Jost~Tobias Springenberg, Ziyu Wang, Nicolas Heess, and Martin
  Riedmiller.
\newblock Learning an embedding space for transferable robot skills.
\newblock In \emph{International Conference on Learning Representations}, 2018.

\bibitem[Wang et~al.(2017)Wang, Merel, Reed, Wayne, de~Freitas, and
  Heess]{wang2017robust}
Ziyu Wang, Josh Merel, Scott~E. Reed, Greg Wayne, Nando de~Freitas, and Nicolas
  Heess.
\newblock Robust imitation of diverse behaviors.
\newblock \emph{CoRR}, abs/1707.02747, 2017.

\bibitem[Haarnoja et~al.(2018)Haarnoja, Hartikainen, Abbeel, and
  Levine]{haarnoja2018latent}
Tuomas Haarnoja, Kristian Hartikainen, Pieter Abbeel, and Sergey Levine.
\newblock Latent space policies for hierarchical reinforcement learning.
\newblock In \emph{Proceedings of the 35th International Conference on Machine
  Learning}, 2018.

\bibitem[Bacon et~al.(2017)Bacon, Harb, and Precup]{bacon2017option}
Pierre-Luc Bacon, Jean Harb, and Doina Precup.
\newblock The option-critic architecture.
\newblock In \emph{Thirty-First AAAI Conference on Artificial Intelligence},
  2017.

\bibitem[Fox et~al.(2017)Fox, Krishnan, Stoica, and Goldberg]{fox2017multi}
Roy Fox, Sanjay Krishnan, Ion Stoica, and Ken Goldberg.
\newblock Multi-level discovery of deep options.
\newblock \emph{CoRR}, abs/1703.08294, 2017.
\newblock URL \url{http://arxiv.org/abs/1703.08294}.

\bibitem[Frans et~al.(2018)Frans, Ho, Chen, Abbeel, and
  Schulman]{frans2018meta}
Kevin Frans, Jonathan Ho, Xi~Chen, Pieter Abbeel, and John Schulman.
\newblock Meta learning shared hierarchies.
\newblock In \emph{International Conference on Learning Representations}, 2018.

\bibitem[Wulfmeier et~al.(2020{\natexlab{a}})Wulfmeier, Abdolmaleki, Hafner,
  Tobias~Springenberg, Neunert, Siegel, Hertweck, Lampe, Heess, and
  Riedmiller]{wulfmeier2020compositional}
Markus Wulfmeier, Abbas Abdolmaleki, Roland Hafner, Jost Tobias~Springenberg,
  Michael Neunert, Noah Siegel, Tim Hertweck, Thomas Lampe, Nicolas Heess, and
  Martin Riedmiller.
\newblock Compositional transfer in hierarchical reinforcement learning.
\newblock \emph{Robotics: Science and Systems XVI}, Jul 2020{\natexlab{a}}.
\newblock \doi{10.15607/rss.2020.xvi.054}.
\newblock URL \url{http://dx.doi.org/10.15607/rss.2020.xvi.054}.

\bibitem[Wulfmeier et~al.(2020{\natexlab{b}})Wulfmeier, Rao, Hafner, Lampe,
  Abdolmaleki, Hertweck, Neunert, Tirumala, Siegel, Heess, and
  Riedmiller]{wulfmeier2020dataefficient}
Markus Wulfmeier, Dushyant Rao, Roland Hafner, Thomas Lampe, Abbas Abdolmaleki,
  Tim Hertweck, Michael Neunert, Dhruva Tirumala, Noah Siegel, Nicolas Heess,
  and Martin Riedmiller.
\newblock Data-efficient hindsight off-policy option learning,
  2020{\natexlab{b}}.

\bibitem[Krishnan et~al.(2017)Krishnan, Fox, Stoica, and
  Goldberg]{krishnan2017discovery}
Sanjay Krishnan, Roy Fox, Ion Stoica, and Ken Goldberg.
\newblock {DDCO:} discovery of deep continuous options forrobot learning from
  demonstrations.
\newblock \emph{CoRR}, abs/1710.05421, 2017.
\newblock URL \url{http://arxiv.org/abs/1710.05421}.

\bibitem[Eysenbach et~al.(2019)Eysenbach, Gupta, Ibarz, and
  Levine]{eysenbach2018diversity}
Benjamin Eysenbach, Abhishek Gupta, Julian Ibarz, and Sergey Levine.
\newblock Diversity is all you need: Learning skills without a reward function.
\newblock In \emph{International Conference on Learning Representations}, 2019.

\bibitem[rob()]{robocup}
{RoboCup project}.
\newblock URL \url{https://www.robocup.org}.
\newblock Available at \url{https://www.robocup.org}, accessed 09.09.2020.

\bibitem[Stone and Veloso(1999)]{LNAI98-tpot-rl}
Peter Stone and Manuela Veloso.
\newblock Team-partitioned, opaque-transition reinforcement learning.
\newblock In Minoru Asada and Hiroaki Kitano, editors, \emph{{R}obo{C}up-98:
  Robot Soccer World Cup {II}}, volume 1604 of \emph{Lecture Notes in
  Artificial Intelligence}, pages 261--72. Springer Verlag, Berlin, 1999.
\newblock Also in {\it Proceedings of the Third International Conference on
  Autonomous Agents}, 1999.

\bibitem[Tuyls et~al.(2002)Tuyls, Maes, and Manderick]{TuylsMM02}
Karl Tuyls, Sam Maes, and Bernard Manderick.
\newblock Reinforcement learning in large state spaces.
\newblock In Gal~A. Kaminka, Pedro~U. Lima, and Ra{\'{u}}l Rojas, editors,
  \emph{RoboCup 2002: Robot Soccer World Cup {VI}}, volume 2752 of
  \emph{Lecture Notes in Computer Science}, pages 319--326. Springer, 2002.

\bibitem[Kohl and Stone(2004{\natexlab{a}})]{icra04}
Nate Kohl and Peter Stone.
\newblock Policy gradient reinforcement learning for fast quadrupedal
  locomotion.
\newblock In \emph{Proceedings of the {IEEE} International Conference on
  Robotics and Automation}, May 2004{\natexlab{a}}.

\bibitem[Kohl and Stone(2004{\natexlab{b}})]{AAI04}
Nate Kohl and Peter Stone.
\newblock Machine learning for fast quadrupedal locomotion.
\newblock In \emph{The Nineteenth National Conference on Artificial
  Intelligence}, pages 611--616, July 2004{\natexlab{b}}.

\bibitem[Saggar et~al.(2007)Saggar, D'Silva, Kohl, and Stone]{LNAI2006-manish}
Manish Saggar, Thomas D'Silva, Nate Kohl, and Peter Stone.
\newblock Autonomous learning of stable quadruped locomotion.
\newblock In Gerhard Lakemeyer, Elizabeth Sklar, Domenico Sorenti, and Tomoichi
  Takahashi, editors, \emph{{R}obo{C}up-2006: Robot Soccer World Cup {X}},
  volume 4434 of \emph{Lecture Notes in Artificial Intelligence}, pages
  98--109. Springer Verlag, Berlin, 2007.
\newblock ISBN 978-3-540-74023-0.

\bibitem[Fidelman and Stone(2007)]{LNAI2006-peggy}
Peggy Fidelman and Peter Stone.
\newblock The chin pinch: A case study in skill learning on a legged robot.
\newblock In Gerhard Lakemeyer, Elizabeth Sklar, Domenico Sorenti, and Tomoichi
  Takahashi, editors, \emph{{R}obo{C}up-2006: Robot Soccer World Cup {X}},
  volume 4434 of \emph{Lecture Notes in Artificial Intelligence}, pages 59--71.
  Springer Verlag, Berlin, 2007.
\newblock ISBN 978-3-540-74023-0.

\bibitem[Hausknecht and Stone(2011)]{LNAI10-hausknecht}
Matthew Hausknecht and Peter Stone.
\newblock Learning powerful kicks on the {A}ibo {ERS}-7: The quest for a
  striker.
\newblock In Javier~Ruiz del Solar, Eric Chown, and Paul~G. Pl\"oger, editors,
  \emph{{R}obo{C}up-2010: Robot Soccer World Cup {XIV}}, volume 6556 of
  \emph{Lecture Notes in Artificial Intelligence}, pages 254--65. Springer
  Verlag, Berlin, 2011.

\bibitem[R{\"o}fer et~al.(2019)R{\"o}fer, Laue, Felsch, Hasselbring, Ha{\ss},
  Oppermann, Reichenberg, and Schrader]{B-Human19}
Thomas R{\"o}fer, Tim Laue, Gerrit Felsch, Arne Hasselbring, Tim Ha{\ss}, Jan
  Oppermann, Philip Reichenberg, and Nicole Schrader.
\newblock B-{H}uman 2019 -- complex team play under natural lighting
  conditions.
\newblock In Stephan Chalup, Tim Niemueller, Jackrit Suthakorn, and Mary-Anne
  Williams, editors, \emph{RoboCup 2019: Robot World Cup XXIII}, pages
  646--657, Cham, 2019. Springer International Publishing.

\bibitem[Stone et~al.(2005)Stone, Sutton, and Kuhlmann]{AB05}
Peter Stone, Richard~S. Sutton, and Gregory Kuhlmann.
\newblock Reinforcement learning for {R}obo{C}up-soccer keepaway.
\newblock \emph{Adaptive Behavior}, 13\penalty0 (3):\penalty0 165--188, 2005.

\bibitem[Kalyanakrishnan et~al.(2008)Kalyanakrishnan, Stone, and
  Liu]{LNAI2007-shivaram}
Shivaram Kalyanakrishnan, Peter Stone, and Yaxin Liu.
\newblock Model-based reinforcement learning in a complex domain.
\newblock In Ubbo Visser, Fernando Ribeiro, Takeshi Ohashi, and Frank Dellaert,
  editors, \emph{{R}obo{C}up-2007: Robot Soccer World Cup {XI}}, volume 5001 of
  \emph{Lecture Notes in Artificial Intelligence}, pages 171--83. Springer
  Verlag, Berlin, 2008.

\bibitem[Kalyanakrishnan and Stone(2010)]{LNAI09-kalyanakrishnan-1}
Shivaram Kalyanakrishnan and Peter Stone.
\newblock Learning complementary multiagent behaviors: A case study.
\newblock In Jacky Baltes, Michail~G. Lagoudakis, Tadashi Naruse, and
  Saeed~Shiry Ghidary, editors, \emph{{R}obo{C}up 2009: Robot Soccer World Cup
  {XIII}}, pages 153--165. Springer Verlag, 2010.

\bibitem[Kalyanakrishnan et~al.(2007)Kalyanakrishnan, Liu, and
  Stone]{LNAI2006-shivaram}
Shivaram Kalyanakrishnan, Yaxin Liu, and Peter Stone.
\newblock Half field offense in {R}obo{C}up soccer: A multiagent reinforcement
  learning case study.
\newblock In Gerhard Lakemeyer, Elizabeth Sklar, Domenico Sorenti, and Tomoichi
  Takahashi, editors, \emph{{R}obo{C}up-2006: {R}obot {S}occer {W}orld {C}up
  {X}}, volume 4434 of \emph{Lecture Notes in Artificial Intelligence}, pages
  72--85. Springer Verlag, Berlin, 2007.
\newblock ISBN 978-3-540-74023-0.

\bibitem[Riedmiller et~al.(2009)Riedmiller, Gabel, Hafner, and
  Lange]{Riedmiller09}
Martin Riedmiller, Thomas Gabel, Roland Hafner, and Sascha Lange.
\newblock Reinforcement learning for robot soccer.
\newblock \emph{Autonomous Robots}, 27\penalty0 (1):\penalty0 55--73, 2009.

\bibitem[Lauer et~al.(2010)Lauer, Hafner, Lange, and Riedmiller]{Lauer10}
Martin Lauer, Roland Hafner, Sascha Lange, and Martin Riedmiller.
\newblock Cognitive concepts in autonomous soccer playing robots.
\newblock \emph{Cognitive Systems Research}, 11:\penalty0 287--309, 09 2010.

\bibitem[Gabel and Riedmiller(2010)]{Gabel10}
Thomas Gabel and Martin Riedmiller.
\newblock On progress in {R}obo{C}up: The simulation league showcase.
\newblock pages 36--47, 01 2010.

\bibitem[Riedmiller et~al.(2008)Riedmiller, Hafner, Lange, and
  Lauer]{RiedmillerHLL08}
Martin~A. Riedmiller, Roland Hafner, Sascha Lange, and Martin Lauer.
\newblock Learning to dribble on a real robot by success and failure.
\newblock In \emph{2008 {IEEE} International Conference on Robotics and
  Automation, {ICRA} 2008, May 19-23, 2008, Pasadena, California, {USA}}, pages
  2207--2208. {IEEE}, 2008.

\bibitem[Riedmiller et~al.(2000)Riedmiller, Merke, Meier, Hoffmann, Sinner,
  Thate, and Ehrmann]{Riedmiller_karlsruhebrainstormers}
M.~Riedmiller, A.~Merke, D.~Meier, A.~Hoffmann, A.~Sinner, O.~Thate, and
  R.~Ehrmann.
\newblock Karlsruhe {B}rainstormers - a reinforcement learning approach to
  robotic soccer.
\newblock In \emph{RoboCup-2000: Robot Soccer World Cup IV, LNCS}, pages
  367--372. Springer, 2000.

\bibitem[Farchy et~al.(2013)Farchy, Barrett, MacAlpine, and
  Stone]{AAMAS13-Farchy}
Alon Farchy, Samuel Barrett, Patrick MacAlpine, and Peter Stone.
\newblock Humanoid robots learning to walk faster: From the real world to
  simulation and back.
\newblock In \emph{Proc. of 12th Int. Conf. on Autonomous Agents and Multiagent
  Systems (AAMAS)}, May 2013.

\bibitem[Hanna and Stone(2017)]{AAAI17-Hanna}
Josiah Hanna and Peter Stone.
\newblock Grounded action transformation for robot learning in simulation.
\newblock In \emph{Proceedings of the 31st AAAI Conference on Artificial
  Intelligence (AAAI)}, February 2017.

\bibitem[Karnan et~al.(2020)Karnan, Desai, Hanna, Warnell, and
  Stone]{IROS20-Karnan}
Haresh Karnan, Siddharth Desai, Josiah~P. Hanna, Garrett Warnell, and Peter
  Stone.
\newblock Reinforced grounded action transformation for sim-to-real transfer.
\newblock In \emph{IEEE/RSJ International Conference on Intelligent Robots and
  Systems (IROS 2020)}, October 2020.

\bibitem[Desai et~al.(2020)Desai, Karnan, Hanna, Warnell, and
  Stone]{IROS20-Desai}
Siddharth Desai, Haresh Karnan, Josiah~P. Hanna, Garrett Warnell, and Peter
  Stone.
\newblock Stochastic grounded action transformation for robot learning in
  simulation.
\newblock In \emph{IEEE/RSJ International Conference on Intelligent Robots and
  Systems(IROS 2020)}, October 2020.

\bibitem[Urieli et~al.(2011)Urieli, MacAlpine, Kalyanakrishnan, Bentor, and
  Stone]{Urieli11}
Daniel Urieli, Patrick MacAlpine, Shivaram Kalyanakrishnan, Yinon Bentor, and
  Peter Stone.
\newblock On optimizing interdependent skills: A case study in simulated 3d
  humanoid robot soccer.
\newblock In Kagan Tumer, Pinar Yolum, Liz Sonenberg, and Peter Stone, editors,
  \emph{Proc. of 10th Int. Conf. on Autonomous Agents and Multiagent Systems
  (AAMAS)}, volume~2, pages 769--776. IFAAMAS, May 2011.
\newblock ISBN 978-0-9826571-5-7.

\bibitem[Abreu et~al.(2019)Abreu, Reis, and Lau]{abreu2019learning}
Miguel Abreu, Luis~Paulo Reis, and Nuno Lau.
\newblock Learning to run faster in a humanoid robot soccer environment through
  reinforcement learning.
\newblock In \emph{Robot World Cup}, pages 3--15. Springer, 2019.

\bibitem[Hausknecht and Stone(2016)]{ICLR16-hausknecht}
Matthew Hausknecht and Peter Stone.
\newblock Deep reinforcement learning in parameterized action space.
\newblock In \emph{Proceedings of the International Conference on Learning
  Representations (ICLR)}, May 2016.

\bibitem[MacAlpine et~al.(2019)MacAlpine, Torabi, Pavse, and
  Stone]{macalpine2019robocup}
Patrick MacAlpine, Faraz Torabi, Brahma Pavse, and Peter Stone.
\newblock {UT} {A}ustin {V}illa: {R}obo{C}up 2019 3{D} simulation league
  competition and technical challenge champions.
\newblock In Stephan Chalup, Tim Niemueller, Jackrit Suthakorn, and Mary-Anne
  Williams, editors, \emph{{R}obo{C}up 2019: Robot World Cup {XXIII}}, Lecture
  Notes in Artificial Intelligence, pages 540--52. Springer, 2019.

\bibitem[MacAlpine and Stone(2018)]{MacAlpineOverlapping}
Patrick MacAlpine and Peter Stone.
\newblock Overlapping layered learning.
\newblock \emph{Artif. Intell.}, 254:\penalty0 21--43, 2018.

\bibitem[Munos et~al.(2016)Munos, Stepleton, Harutyunyan, and
  Bellemare]{munos2016safe}
R{\'e}mi Munos, Tom Stepleton, Anna Harutyunyan, and Marc Bellemare.
\newblock Safe and efficient off-policy reinforcement learning.
\newblock In \emph{Advances in Neural Information Processing Systems}, 2016.

\bibitem[Wu et~al.(2013)Wu, Tassa, Kumar, Movellan, and Todorov]{wu2013stac}
Tingfan Wu, Yuval Tassa, Vikash Kumar, Javier Movellan, and Emanuel Todorov.
\newblock {STAC: Simultaneous tracking and calibration}.
\newblock In \emph{2013 13th IEEE-RAS International Conference on Humanoid
  Robots (Humanoids)}, pages 469--476. IEEE, 2013.

\bibitem[Balduzzi et~al.(2019)Balduzzi, Garnelo, Bachrach, Czarnecki,
  P{\'{e}}rolat, Jaderberg, and Graepel]{balduzzi2019open}
David Balduzzi, Marta Garnelo, Yoram Bachrach, Wojciech Czarnecki, Julien
  P{\'{e}}rolat, Max Jaderberg, and Thore Graepel.
\newblock Open-ended learning in symmetric zero-sum games.
\newblock In Kamalika Chaudhuri and Ruslan Salakhutdinov, editors,
  \emph{Proceedings of the 36th International Conference on Machine Learning,
  {ICML} 2019, 9-15 June 2019, Long Beach, California, {USA}}, volume~97 of
  \emph{Proceedings of Machine Learning Research}, pages 434--443. {PMLR},
  2019.
\newblock URL \url{http://proceedings.mlr.press/v97/balduzzi19a.html}.

\end{thebibliography}
\end{small}

\appendix

\section{Appendix: Environment}

\subsection{Evaluators}
\label{app:Evaluators}

\begin{figure}
    \centering
    \includegraphics[width=\textwidth]{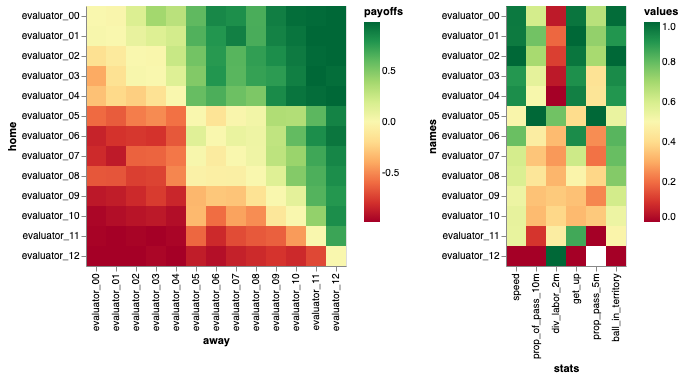}
    \caption{{\bf Left:} Pairwise evaluator payoff matrix. {\bf Right:} Behaviour statistics of the evaluator agents. Blank cells indicate that the statistic could not be calculated for the given agent, for example, because it made no passes.}
    \label{fig:evaluators_payoff}
\end{figure}

The set of 13 evaluators that our agents are continually evaluated against exhibit a range of levels of skills as shown in Figure~\ref{fig:evaluators_payoff}. In particular, \textit{evaluator\_12}, the weakest agent acts randomly and does not interact with the ball (resulting in undefined \textit{prop\_pass\_5m}).

\subsection{Drills}
\label{sec:Appendix:Drills}

\begin{figure}[htp]
  \centering
  \includegraphics[width=\textwidth]{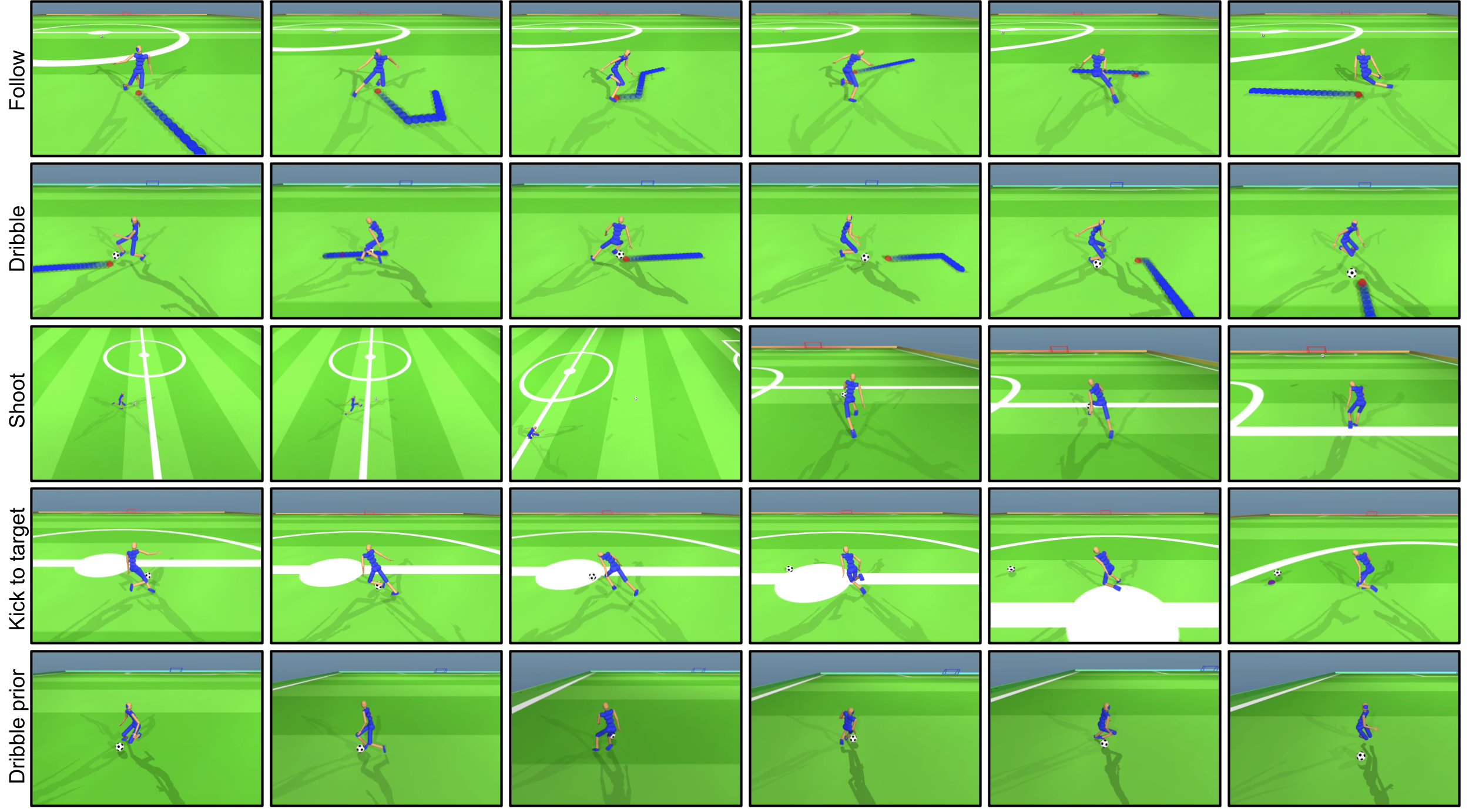}
\caption{\captionsize{Top row: a \emph{follow} expert performing the \emph{follow} drill, remaining close to the moving target (red spot) and observing the future position of the target (blue trace). 2nd row: \emph{dribble} expert performing the \emph{dribble} drill. 3rd row: the \emph{shoot} drill. 4th row: the \emph{kick-to-target} drill. 5th row: the distilled \emph{dribble} drill prior performing the \emph{dribble} drill. The agent dribbles the ball in a similar manner to the \emph{dribble} expert but does not observe the target and so does not remain close to it. See also \textcolor{blue}{\href{\drillepisodesurl}{videos of full drill episodes at \drillepisodesurltext}}.}}
  \label{fig:DrillScreenshots}
\end{figure}

See \autoref{fig:DrillScreenshots} for visualizations of the drill tasks.

\subsection{Drill Expert and Prior Policy Observation}
\label{sec:Appendix:DrillObservations}

Recall that the observation space for the football task, $\cO^0$, is composed of proprioceptive observations $\cX$ and the football-specific context observations $\cC^0$. Similarly, the observation spaces $\cO^k$ for the drill experts are comprised of $\cX$ and the drill-specific context observations $\cC^k$. We can then express the relevant contexts for the transferable behaviour priors as the intersection of context observation feature sets $\tilde \cC^k = \cC^k \cap \cC^0$.

\section{Appendix: Training}

\subsection{Task Abstraction Using Stochastic Games}
\label{sec:Appendix:Training:Abstraction}

We model each of the $K$ tasks using the framework of \textit{Stochastic Games} \cite{shapley1953stochastic}, which generalizes the concept of a \textit{Markov Decision Process} (MDP) to $n\ge 1$ decision makers. The $k$-th task is modelled as a stochastic game $\cG_k = (\cS^k, \cO^k, \cA, \phi^k_{1:n}, r^k_{1:n}, P^k, p^k_0)$, with state space $\cS^k$, observation space $\cO^k$, action  space $\cA$, agent-specific observation functions $\{\phi^k_i : \cS^k \rightarrow \cO^k \mid i \in [n]\}$, and reward functions $\{r^k_i : \cS^k \rightarrow \bR \mid i \in [n]\}$. The Markovian transition function $P^k$  defines the conditional distribution over successor states given previous state-actions $P^k(s_{t+1}|s_t, a^1_t, \ldots, a^n_t)$, and $p^k_0$ defines the distribution of initial states over $\cS^k$. 
The embodiment of the humanoid football players is identical in all tasks, and the agents act by providing a 56-dimensional continuous vector to the environment, so that the action set is consistent across tasks and players. The observation sets are consistent across players within each task. Observations are further partially consistent \emph{across} tasks: proprioceptive observations $x\in\cX$ are present for all players in all tasks; task-specific observations $c\in \mathcal{C}^k$, including features with information of the ball, goal-posts, the moving target and other players recur in a subset of tasks. This partial consistency enables skill transfer across the family of tasks.

\subsection{Policy Optimization with Maximum a Posteriori Policy Optimization}
\label{sec:Appendix:Training:MPO}

For each reward channel, $\ell \in [M]$, we define action-value functions $$Q_{ \pi, \coplayers}^\ell(s_\tau, a_\tau; \alpha_\ell, \gamma_\ell) := \alpha_\ell \mathbb{E}\Bigl[\sum^\infty_{t=\tau} \gamma_\ell^{t-\tau} \hat{r}_{i}^\ell(s_t) \mid a^{i}_\tau = a_\tau ~;~ \pi^i=\pi, \pi^{\noti} = \coplayers \Bigr],$$ and define $Q_{\pi, \coplayers}(s_t, a_t; \myvec{\alpha}, \myvec{\gamma}):= \sum_{\ell=1}^M Q_{\pi, \coplayers}^\ell(s_t, a_t;\alpha_\ell, \gamma_\ell)$. Each agent maintains action-value function approximations for each channel, parametrized by network parameters $\theta_{\ell}^Q$,
\begin{equation}
\hat Q^\ell_{\theta_{\ell}^Q}(h_t, a_t; \alpha_\ell, \gamma_\ell) \approx Q_{\pi_{\theta}, \coplayers}^\ell(s_t, a_t;\alpha_\ell, \gamma_\ell) \nonumber
\end{equation}
so that $\hat Q_{\theta^Q}(h_t, a_t; \myvec{\alpha}, \myvec{\gamma}) := \sum_{\ell=1}^M \hat Q^\ell_{\theta_\ell^Q}(h_t, a_t;\alpha_\ell, \gamma_\ell) \approx Q_{\pi_{\theta}, \coplayers}(s_t, a_t; \myvec{\alpha}, \myvec{\gamma})$. Note that agents do not observe the identity of their opponents and therefore rely upon the history to infer the dependence of the action-value function on the coplayers $\coplayers$ in any particular episode. For this reason it is important that the $\hat Q^\ell_{\theta_\ell^Q}$ are parametrized using recurrent neural networks.

We consider the paradigm of independent-learning and independent-execution and optimize, for each agent its parameters $(\theta^Q_{1},\ldots,\theta^Q_{M}, \theta)$ of the $M$ action-value functions $\{\hat Q_{\theta_ \ell^Q}(h_t, a_t; \alpha_\ell, \gamma_\ell)\}^M_{\ell=1}$, and policy $\pi_{\theta}(\cdot| h_t)$. As described in Section \ref{sec:Experiments:Infrastructure}, policy and value function updates are computed from off-policy data sampled from an agent-specific replay buffer.

\paragraph{Policy Evaluation}
We employ the \textit{retrace} off-policy correction \cite{munos2016safe} to compute $M$ target Q-values $\{\hat{Q}_\ell^{ret} : \ell\in[M]\}$ and, for each agent in the population, minimize the loss:

\begin{equation}
    \min_{\theta_1^Q, \cdots, \theta_M^Q}\bE\left[\sum^M_{\ell=1} \sum_t (\hat{Q}_\ell^{ret}(h_t, a_t) - \hat Q_{\theta_\ell^Q}(h_t, a_t; \alpha_\ell, \gamma_\ell))^2 \right]
\end{equation}
where expectation is over trajectories sampled from the agent's replay buffer.

\paragraph{Policy Improvement}
We follow \cite{AbdolmalekiMPO} and compute policy updates with a two-step procedure.
In the first step we compute an improved policy according to the constrained optimization problem
\begin{equation}
\begin{split}
&\max_{q} %
\int_{a_t} q(a_t|h_t) \sum_\ell \hat Q_{\theta_\ell^Q}(h_t,a_t;\alpha_\ell,\gamma_\ell)\,da_t~~~~~~~\forall h_t \\
&{\rm s.t.}\,\bE \left[D_{KL}(q(a_t|h_t) \lVert \pi_{old}(a_t|h_t)) \right] < \epsilon,
\end{split}
\end{equation}
where $\epsilon$ specifies the maximum allowed change of $q$ relative to the current policy $\pi_{old}$.\footnote{We note that, since $\hat Q_{\theta^Q}$ is a generalized action-value function, with a separate discount factor per channel, there is no guarantee of policy improvement when choosing actions greedily to maximize $\hat Q_{\theta^Q}$, per state.}
We solve this problem in closed form $$q^*(a_t|h_t; \myvec{\alpha}, \myvec{\gamma}) \propto \pi_{old}(a_t|h_t){\rm exp}\left(\frac{\sum^M_\ell Q_{\theta_\ell^Q}(h_t,a_t;\alpha_\ell, \gamma_\ell)}{\eta}\right)$$ with $\eta$ the temperature computed based on the constraint $\epsilon$ and $\pi_{old}$ the current policy. 
We represent the optimal $q^*$  non-parametrically through a set of weighted action samples

In the second step we update the parametric policy $\pi_{\theta}$ through a projection step that minimizes the KL-divergence to the non-parametric policy, $q(a_t|h_t)$, subject to a constraint which implements a trust region of size $\beta$ and  improves stability of learning,
\begin{equation}
\begin{split}
&\mathcal{L}_{reward}(\theta ; \myvec{\alpha}, \myvec{\gamma}) := \bE\left[ \int_{a_t} q(a_t|h_t; \myvec{\alpha}, \myvec{\gamma}) \log \pi_{\theta}(a_t|h_t) \right] \\
&{\rm s.t.}\,\bE \left[D_{KL}(\pi_{old}(a_t|h_t) \lVert \pi_\theta(a_t|h_t))  \right] < \beta.
\end{split}
\end{equation}

\subsection{Motion Capture and NPMP Training Details}
\label{sec:Appendix:NPMP}

Motion capture data of the football vignettes was licensed from Audiomotion Studios, where this football data had been commissioned for an unrelated project by other clients.  The football vignettes consisted of different numbers of players per scene, and the motions of each individual player was tracked.  For our purposes we treated player motions individually (ignoring the interactions between players).  The dataset amounted to  560 player-clips, with this total coming from multiple players in each of a smaller number of vignettes.  In total, this was roughly 1 hour and 45 minutes of player-time.  As noted in the main text, we cut the 560 player-clips into snippets that were 4-8 seconds in duration.  More specifically, the original 560 player trajectories were cut into 1245 shorter snippets, and these snippets were then %
imitated by separate tracking policies. 

Although most clips did not involve interactions between the humanoid and the ground, excepting foot contacts, a meaningful subset of the data (100/1245 snippets) displayed behaviours such as the goalie diving for the ball, players performing a slide tackle/kick, and players standing back up after aggressive plays.  We anticipated that the inclusion of this class of behaviours would enable the low-level motor controller to support richer behaviours including recovery from falls.  As noted in the main text, the ball was not tracked and was not included in the motion reconstructions. 

As a first processing step, we converted the motion capture point-cloud data into joint angles for our humanoid body model, using an implementation of STAC \cite{wu2013stac}. Since different human subjects had substantially different body proportions, we resized the humanoid body model to the proportions of each unique human subject and performed STAC for these per-subject body variants.  Once the joint angles were identified for all clips, we re-targeted the joint angles to our standard proportion body model.

For the tracking stage, the motion capture snippets were resampled to a coarser temporal resolution consistent with the control timestep employed for this domain (30 ms/timestep). Our motion capture tracking infrastructure has been open-sourced at {\color{blue} \href{https://github.com/deepmind/dm_control/tree/master/dm_control/locomotion}{dm\_control/locomotion}} \cite{Hasenclever2020}.

The second stage, involving sampling trajectories from the tracking policies and distilling them into an NPMP architecture, is detailed in the main text.  While many possible specific neural network architectures are possible for the encoder and decoder ($q$ and $\npmp$ in \autoref{eqn:ELBO}), in the present work we use an encoder that is a MLP with two hidden layers of 1024 units each, followed by an output (the latent embedding space) of 60 latent variables. The decoder, is also an MLP with three hidden layers of 1024 units each. This was not something we systematically explored for the present work, however this choice was informed by other exploration of network sizes and architectures that we have performed in other work \cite{MerelNPMP,Hasenclever2020}. The latent prior $p_z$ is a simple auto-regressive process as described in \cite{MerelNPMP}.

\subsection{Agent Architecture}
\label{app:AgentArchitecture}

\paragraph{Scene Embedding}

Each agent observes the state of the environment through a set of egocentric features.
These features always include proprioceptive information $x\in\cX$ about the player's pose and velocities. For task $k$ the feature set further includes information about objects that are part of the game context $c^k_c\in \cC^k_{c}$ (including ball, goal posts, moving targets, etc). For the full game of football ($k=0$), each agent additionally observes other players (teammates and opponents) with the observation for each player denoted $c_p \in \cC^0_{p}$. Agents learn a proprioceptive encoder $\psi_{proprio}(x)$ as well as a task context encoder $\psi_{context}(c^k_c)$. In the football game observations of each teammate and opponent are processed by teammate and opponent encoders $\psi_{teammate}(c_p)$ and $\psi_{opponent}(c_p)$ respectively. All feature encoders are implemented as 2-layer MLP networks.

\paragraph{Player-Player Relationship Representation}

For the football task, the structure of the observations offers an opportunity to introduce inductive biases that facilitate representation learning. Recognizing that teammates and opponents have no natural order, we use an order-invariant representation that treats observations of teammates and opponents as unordered sets, so that the number of learned parameters is independent of the number of players.
We first compute \emph{pairwise} encodings for all ${n \choose 2}$ pairs of players. These are then processed by a multi-headed attention module that uses the proprioceptive and context encodings as input to compute queries.

\begin{figure}
  \centering
  \includegraphics[width=\textwidth]{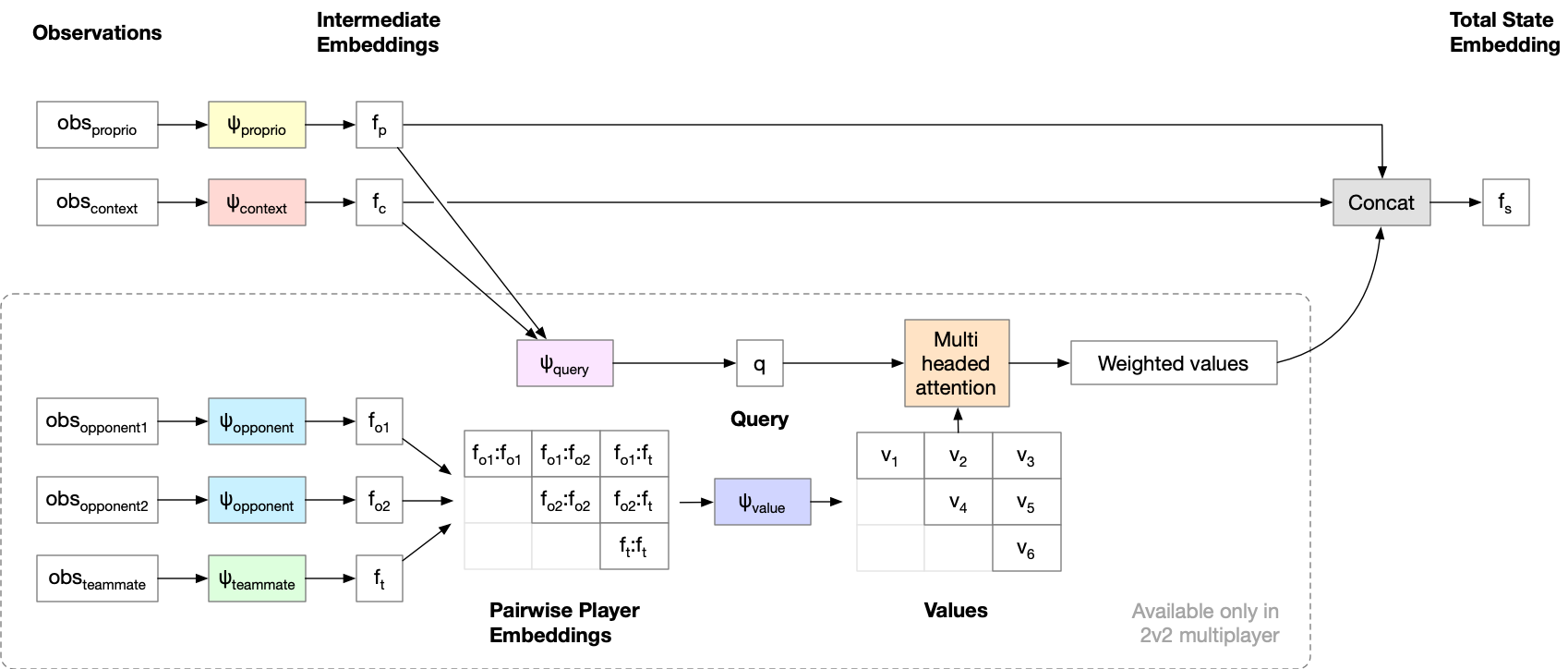}
  \caption{The network architecture of the environment state representation.}
  \label{fig:architecture}
\end{figure}

The multi-headed attention module consists of a query encoder $\psi_{query}$ taking as its input the concatenation of one's proprioceptive encoding $\psi_{proprio}(x)$ and the context encoding $\psi_{context}(c^0_c)$. The encoded query is then concatenated with all $3+2-1 \choose 2$ pairwise peer encodings chosen with replacement and fed to the value encoder $\psi_{value}$. The combination is with replacement which allows for the representation of relationships between an agent itself and any one of its peers (the diagonal of the pairwise player embeddings matrix in Figure~\ref{fig:architecture}). Given separately parameterized attention heads, we compute 6 attention masks over the resulting value encodings, resulting in 6 weighted pairwise relationship embeddings. The representation of players is therefore order-invariant thanks to the unordered sum via the attention mechanism and decouples the number of parameters from the number of peers. The number of attention heads can be loosely interpreted as the number of different pairwise relationships that could be relevant for downstream decision making.

\paragraph{Environment State Embedding} The final state representation is obtained as the concatenation of the proprioceptive encoding and the context encoding and, in the case of the football task, further concatenated with player-player interaction encodings.

\paragraph{Policy/Action-Value Function Parameterization} Many multi-agent environments are inherently partially observed, for instance because the identity or internal state of the opponent is unknown \cite{LiuSoccer}. In this case, the optimal policy is a function of the interaction history. %
Following the state embedding, LSTM modules \cite{HochreiterLSTM} process history in the policy and action value functions. The agent samples latent actions $z$ from a Gaussian policy which are translated to actions $a \in \cA$ using the fixed low-level motor controller described in Section~\ref{sec:Method:LowLevel}. Reinforcement learning is performed over the latent motor intention space. Thus action-value functions are defined over the motor intention space. Similar to \cite{LiuSoccer}, the action-value function outputs a vectorised return estimates where each dimension corresponds to a reward component.

\subsection{Shaping Rewards}

\label{app:ShapingRewards}

We describe the shaping rewards introduced to each task, including the full task of football and the mid-level drills in Table~\ref{table:ShapingRewards}.

\begin{table}[h!]
    \centering
    \small
    \begin{tabular}{llp{11cm}l}
        \toprule
        \textbf{Task} & \textbf{Reward} & \textbf{Description} \\ %
        \toprule
        \vspace{0.2cm}
        \parbox[t]{2mm}{\multirow{6}{*}{\rotatebox[origin=c]{90}{Football}}}
        & \emph{Scoring} & Returns +1.0 when the home team scores a goal and 0.0 otherwise. \\
        & \emph{Conceding} & Returns -1.0 when the away team scores a goal and 0.0 otherwise. \\ %
        & \emph{Closest Velocity to Ball} & Returns the magnitude of the player's velocity to ball if the player is the closest player to the ball in its team. \\ %
        & \emph{Velocity Ball to Goal} & Returns the magnitude of the velocity of the ball towards the center of the opposing team's goal. \\ %
        \midrule
        \vspace{0.2cm}
        \parbox[t]{2mm}{\multirow{4}{*}[-1.3em]{\rotatebox[origin=c]{90}{Dribble}}}
        & \emph{Ball Close to Target} & Returns $e^{-\frac{1}{2}||\myvec{x}_{ball} - \myvec{x}_{target}||}$ where $\myvec{x}_{ball}$, $\myvec{x}_{target}$ denote the cordinates of the ball and the target. This defines the environment reward and fitness measure for the task. \\ %
        \vspace{0.2cm}
        & \emph{Velocity Player to Ball} & Returns the magnitude of the player's velocity to ball. \\
        \vspace{0.2cm}
        & \emph{Velocity Ball to Target} & Returns the magnitude of the velocity of the ball towards the target. \\
        \midrule
        \vspace{0.2cm}
        \parbox[t]{2mm}{\multirow{1}{*}[-0.6em]{\rotatebox[origin=c]{90}{Follow}}}
        & \emph{Close to Target} & Returns $e^{-\frac{1}{2}||\myvec{x}_{player} - \myvec{x}_{target}||}$ where $\myvec{x}_{player}$, $\myvec{x}_{target}$ denote the coordinates of the player and the target. This defines the environment reward and fitness measure for the task. \\ %
        \midrule
        \vspace{0.2cm}
        \parbox[t]{2mm}{\multirow{3}{*}[-1.5em]{\rotatebox[origin=c]{90}{Shoot}}}
        & \emph{Velocity Ball to Goal} & See above. This is included in the environment reward and fitness measure for the task. \\ %
        \vspace{0.2cm}
        & \emph{Scoring} & See above. This is included in the environment reward and fitness measure for the task. \\
        \vspace{0.2cm}
        & \emph{Velocity Player to Ball} & Returns the magnitude of the player's velocity to ball. \\
        \midrule
        \vspace{0.2cm}
        \parbox[t]{2mm}{\multirow{3}{*}[-1.0em]{\rotatebox[origin=c]{90}{Kick-to-target}}}
        & \emph{Ball Close to Target} & Returns $e^{-\frac{1}{5}||\myvec{x}_{ball} - \myvec{x}_{target}||}$ where $\myvec{x}_{ball}$, $\myvec{x}_{target}$ denote the coordinates of the ball and the target. This defines the environment reward and fitness measure for the task. \\ %
        \vspace{0.2cm}
        & \emph{Velocity Player to Ball} & Returns the magnitude of the player's velocity to ball. \\
        \vspace{0.2cm}
        & \emph{Velocity Ball to Target} & Returns the magnitude of the velocity of the ball towards the target. \\
        \bottomrule
    \end{tabular}
    \caption{Shaping rewards used for each task and their descriptions.}
    \label{table:ShapingRewards}
\end{table}

\subsection{Behaviour Shaping with Behaviour Priors}
\label{app:MixtureKL}

The KL between a distribution $p$ and a mixture distribution with components $\{q_i\}$ is upper bounded, up to a constant defined in terms of the mixture weights, by the minimum KL between $p$ and any of the components:
\begin{align}
\textstyle
D_{KL}\left [p||\sum_i \alpha_i q_i\right ] &= \int p(x)\log \frac{p(x)}{\sum_i\alpha_i q_i(x)} dx \nonumber \\
&\le {\rm min}_i \int p(x)\log \frac{p(x)}{\alpha_i q_i(x)} dx \nonumber \\
&=  {\rm min}_i \left( D_{KL}(p||q_i)  - \log \alpha_i  \right) \nonumber
\end{align} 

\subsection{Multi-Agent Population Fitness Measure}  
\label{app:FitnessMeasure}

We continuously keep track of each agent's fitness, as outlined by the Fitness Update in Algorithm~\ref{alg:pbt}. In contrast to the single-agent setting, where the reward function offers a direct performance ranking of population members, the stochastic game does not by itself offer a well-defined complete ordering. Specifically, the terminal reward of win, loss or draw from a match (corresponding to a terminal reward of +1, -1 and 0) between agents $(i, j)$ is only informative insofar as to compare the relative strength of the pair of agents involved. Relative performance between agents also need not be transitive \cite{balduzzi2019open}.

We continuously keep track of an empirical payoff matrix $\cM : \cW \times \cW \rightarrow \bR$ between all pairs of continuously learning agents within a finite population of size 16. We model the ternary terminal rewards as a beta distribution $B(\alpha, \beta)$ where $\alpha$ ($\beta$) corresponds to the counts of wins (losses) from sampled match-ups. A result of draw counts towards both $\alpha$ and $\beta$. To account for the continual learning of agents, we exponentially decay the counts of pairwise win/loss results throughout training. Based on the empirical payoff matrix $\cM$, we define a fitness vector $f \in \bR^{|\cW|}$ measuring the empirical performance of agents in the population. While popular fitness measure such as \textit{Elo} have been used extensively in the MARL literature \cite{JaderbergCTF, LiuSoccer}, we adopted \textit{Nash Averaging} \cite{balduzzi2018re} which is invariant to the introduction of agents that win or lose to the same set of opponents. Treating the task as a two-player zero-sum game, we can solve for the unique mixed Nash strategy represented by $\myvec{p}$ the \textit{Nash} distribution over the population $\cW$. The fitness vector of agents is defined as $\myvec{f} = \cM \cdot \myvec{p}$ or their expected payoff when playing against the mixed Nash strategy. Note that agents with non-zero support under the \textit{Nash} equilibrium by definition have the maximum fitness of 0.5, representing a 50\% win-rate against the \textit{Nash} mixture. 

In short, our population-based training scheme amounts to evaluating agents against the {\it Nash} mixture player, while improving over the {\it average} player of the population.

\section{Appendix: Behaviors}
\label{sec:Appendix:Behaviors}

\subsection{Behavior Statistics}
\label{sec:Appendix:Behaviors:Stats}

{\bf Basic locomotion skills}: we measure the agent's \emph{speed}, and ability to \emph{get up} -- defined as the proportion of times an agent is able to recover from being fallen.

{\bf Football skills}: we measure (1) the proportion of timesteps in which the closest player to the ball is a member of the team, \emph{ball control}; (2) the proportion of ball touches which are passes of range 5m or more (\emph{pass frequency}), and (3) the proportion of passes which are of range 10m or more (\emph{pass range}).

{\bf Teamwork statistics}: we firstly design a metric to measure \emph{division of labour}, which we define as $$DOL(\pi):= 1 - \frac{\bE[\frac{1}{T}\sum_{t=1}^T \bI_{crowding}(s_t)]}{\bE[\frac{1}{T}\sum_{t=1}^T \bI_{close-to-ball}(s_t)]},$$ 
where $\bI_{close-to-ball}$ is an indicator for the event that one player on the team is within 2m of the ball and $\bI_{crowding}$ is an indicator for the event that both players on the team are within 2m of the ball. Expectation is over trajectories encountered by playing policy $\pi$ against a randomly sampled evaluation agent. Thus \emph{division of labour} is close to 0 if, whenever one player is close to the ball, so is it's teammate, and close to 1 if, whenever one player is close to the ball, it's teammate does not try to possess the ball but adopts an alternative behaviour (typically turning and heading up-field in anticipation of an up-field kick, pass or shot). We secondly consider \emph{Receiver off-ball scoring opportunity} (OBSO), which is described in detail in Section~\ref{sec:Appendix:Behavior:OBSO}. We do not directly encourage agents to optimize this quantity, but we are interested in whether our agents improve useful metrics known to the sports analytics community.

\begin{figure}
  \centering
  \includegraphics[width=\textwidth]{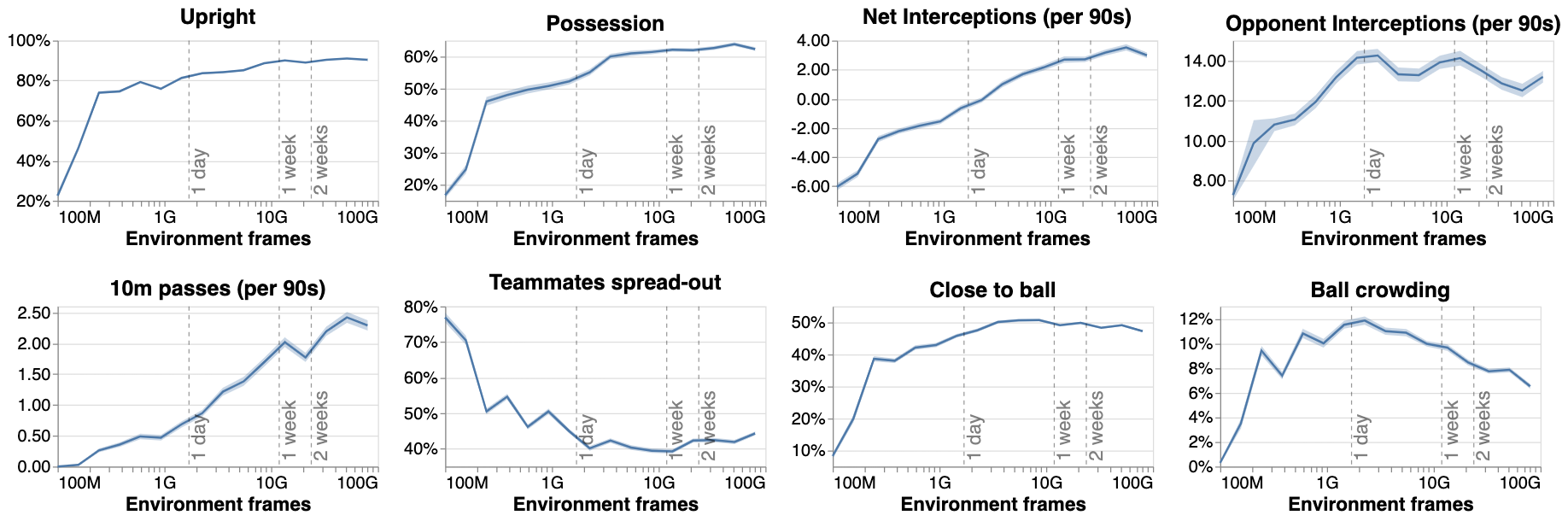}
  \caption{
    Locomotion skills, such as the ability to remain {\bf upright}, are learned quickly, with the majority of improvement occurring within 6 hours of training. The agent is able to stay upright 75\% of the time after 10 hours of training but continues to gradually improve its robustness in locomotion for at least 1 week of further refinement. Ball {\bf possession} is also learned early on but continues to improve for at least 1 week. We also see that the agents improve their ability to intercept, with {\bf net interceptions} still significantly increasing in phase 2, as well as their ability to avoid being {\bf intercepted by opponent}, as the agents also improve their awareness of opponents. Passing occurs early in training but passes are initially infrequent and short-range - after one day of training less than 20\% of passes are over 10m in range. Long-range pass frequency, reflected by the absolute frequency of {\bf 10m passes} over a 90 second game, continuously improve for at least 2 weeks of training. Long-range passes eventually represent 40\% of passes. Behaviours related to the division of labour are learned more slowly. For 1 day of training the {\bf teammates spread out} metric significantly decrease, as agents prioritize ball possession, and behaviours are characterized by individualistic ball chasing. After 1 day of training there is a phase shift and the off-ball player starts to reduce {\bf ball crowding} and cease competing for possession with a teammate. {\bf Ball crowding} decreases significantly, despite {\bf close to ball} remaining high, indicating that at least one teammate remains close to the ball and agents have learnt division of labour. After 1 week of training there is also small increase in the average spread of agents measured by the {\bf teammates spread out} statistic.
  } 
\label{ExtraBehaviors}
\end{figure}

In addition to the emergence of behaviours reported in Section~\ref{sec:behaviour}, we analyze emergence of additional behaviours in Figure~\ref{ExtraBehaviors}. Additional behaviours, not detailed in Section~\ref{sec:behaviour} are detailed in Table~\ref{table:ExtraBehaviorStatistics}.

\begin{table}[h!]
    \centering
    \small
    \begin{tabular}{llp{11cm}l}
        \toprule
        \textbf{Type} & \textbf{Name} & \textbf{Description} \\ %
        \toprule
        \vspace{0.0cm}
        \parbox[t]{2mm}{\multirow{1}{*}{\rotatebox[origin=c]{90}{Basic}}}\vspace{0.4cm}
        & \emph{Upright} & Proportion of timesteps in which the agent was not fallen.\footnote{An agent is considered fallen if a body part above the tibia makes contact with the ground, and is considered upright again after 10 consecutive timesteps not fallen.} \\
        \midrule
        \vspace{0.2cm}
        \parbox[t]{2mm}{\multirow{5}{*}[-2.1em]{\rotatebox[origin=c]{90}{Football}}}
        & \emph{10m passes} & Number of passes of range 10m or greater per 90s episode. \\ %
        &\emph{Close to ball} & Proportion of timesteps in which at least one teammate is within 2m of the ball. \\ %
        & 
        \emph{Possession} & Proportion of timesteps that an team had possession of the ball. A team possess the ball if the last player to make contact with the ball was a member of the team. \\ %
        & 
        \emph{Net interceptions} & Number of net interceptions (interceptions for - interceptions against) per 90s episode. \\ %
        &\emph{Interception by opponent} & Number of interceptions conceded to opponent per 90s episode. \\ %
        \midrule
        \vspace{0.5cm}
        \parbox[t]{2mm}{\multirow{2}{*}[0em]{\rotatebox[origin=c]{90}{Team work}}}
        & 
        \emph{Ball crowding} & Proportion of timesteps in which both teammates are within 2m of the ball. \\ %
        &\emph{Teammates spread-out} & Proportion of timesteps in which two teammates are at least 5m apart. \\ %
        \bottomrule
    \end{tabular}
    \caption{Additional behaviour statistics collected during games against evaluation agents.}
    \label{table:ExtraBehaviorStatistics}
\end{table}

\subsection{Collecting Behavioral Statistics}\label{sec:Appendix:Tournament}
To collect the behavioural statistics of a given snapshot of an agent, it played 100 episodes against the set of evaluation agents described in \ref{app:Evaluators}.
The length of each episode is 3000 environment steps (90 seconds), and the behavioural statistics are averaged over environment steps episode-wise.
Finally, for a population, for each behavioural statistic, we report corresponding average over the 300 episodes played by the top 3 agents in the population, measured in terms of Elo against evaluation bots, together with standard error of the mean, i.e. $\frac{\sigma_x}{\sqrt{300}}$ where $\sigma_x$ is the sample standard deviation of the 300 episode means. Error therefore does not relate to variance across independent experiments, but the across episode variance of statistics for the best agents in a single population.

To plot the progression of the agent's behavioural statistics with respect to environment steps, we take snapshots of the agent at 16 intervals in training time up to $80 \times 10^9$ environment steps.
For each snapshot, we repeat the aforementioned procedure to collect the behavioural statistics.

\subsection{Probe Tasks}
\label{sec:Appendix:Behaviors:ProbeTasks}

To test whether the passer kicks the ball in a direction correlated with the receiver position we measure the $y$ component (the direction parallel to the goal-line) of the ball velocity when the passer first touches the ball and moves it forwards. The {\it probe score} measures the performance of a policy at the probe task. On any epsiode of the probe task an agent scores $\frac{1}{2} + \frac{1}{2}{\rm sign}(ball\_vel_y(s_\tau)) {\rm sign}(receiver\_pos_y(s_0))$ where the random variable $\tau$ is first time-point in which the passer touches the ball, or scores $0.5$ if the ball is first touched by a defender or if the ball is first kicked backwards. The probe score is averaged over episodes against evaluation agents: $$PS(\pi):=\frac{1}{2} + \frac{1}{2}\bE_{\traj\sim\cP_{\pi, \pi_{eval}}}[{\rm sign}(ball\_vel_y(s_\tau)) {\rm sign}(receiver\_pos_y(s_0))]$$
where $\cP_{\pi, \pi_{eval}}$ is the distribution over trajectories $\traj = \left((s_t, a_t^1, \ldots , a_t^n, r_t^1, \ldots , r_t^n)\right)_{t\in[T]}$ encountered by playing policy $\pi$ against a randomly sampled evaluation agent in the probe task. Thus a score of 1 means the passer always kicks in the receiver direction, 0 means it never does. 

To understand whether the behaviour of the agent is driven by learned knowledge of the value of certain game states we also measure whether the passer's and receiver's value functions register higher value when the ball travels towards the receiver, rather than away: we analyze the {\it pass-value correlation}\alternative{\footnote{\guycom{There are alternatives which exhibit greater correlation e.g. $V(s_0|{ball\ moves\ left}) - V(s_0|{ball\ moves\ right})$ correlates very highly with $\bI_{receiver\ on\ left}(s_0)$ as it removes a lot of randomness.}}} (PVC) statistic defined on any particular episode via $$PVC(\hat Q^\pi):= \mathrm{corr}(\hat Q^{\pi}_{ scoring}(s_0, a_0), \bI_{ball-to-receiver}(s_0))$$ where $\hat Q^{\pi}_{scoring}$ is the agent's Q-function for the scoring reward channel, and $$\bI_{ ball-to-receiver}(s):= \begin{cases}
  1 & \text{if } {\rm sign}({ball\_vel_y}(s)) = {\rm sign}({receiver\_pos_y}(s)) \\    
  0 & \text{otherwise}    
\end{cases}$$ is a function indicating whether the ball travels towards the receiver wing. We average this quantity over %
all initial configurations of the probe task and over all possible ball velocities which are such that the ball will reach the forward right or forward left quadrant (from the passer's perspective) in 1 second, and over $a\sim\pi(\cdot|s)$.

\subsection{Off-Ball Scoring Opportunity}
\label{sec:Appendix:Behavior:OBSO}

Off-ball scoring opportunities (OBSO) \cite{spearman2018beyond} quantify the quality of an attacker's positions. Specifically, OBSO models the probability of an off-ball player (i.e., a player currently not in possession of the ball) scoring in the immediate future: in order to convert an opportunity into a goal, the ball first needs to be passed to the off-ball player, the player needs to successfully control the ball, and finally score. Importantly, OBSO provides a spatiotemporally dense measure of performance, allowing one to give each player credit for creating opportunities even if they do not yield a goal. We measure an OBSO statistic at regular snapshots as agent training progresses, by playing matches against a fixed set of opponent evaluation agents. %
This section provides a high-level description of the OBSO model used for agent evaluation in the main text, with technical details provided in \cite{spearman2018beyond}.

\subsubsection{Model Overview}

\begin{figure}[t]
    \centering
    \includegraphics[width=0.24\textwidth]{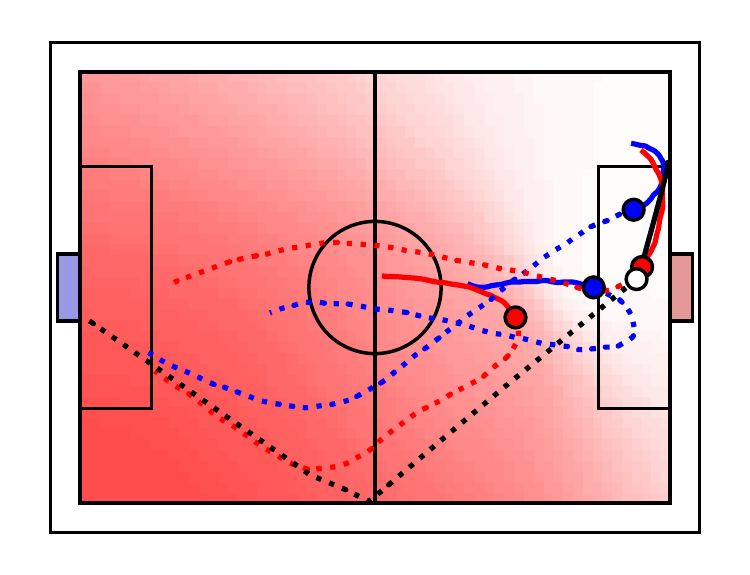}%
    \hfill
    \includegraphics[width=0.24\textwidth]{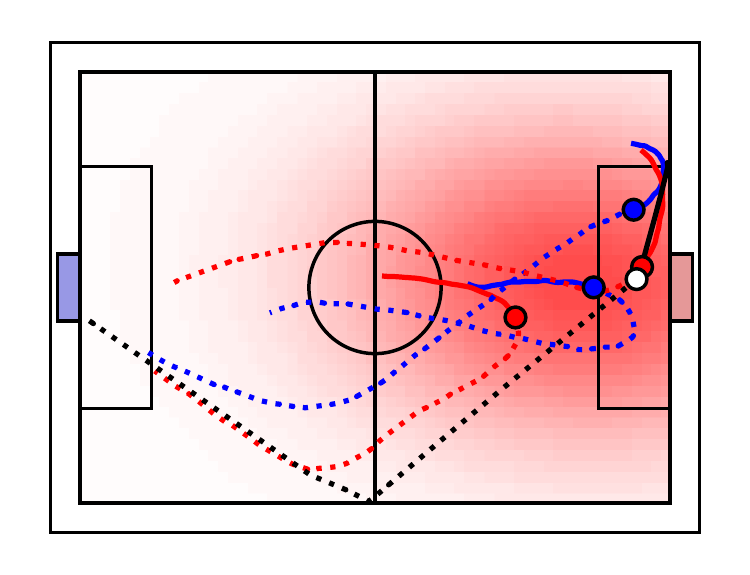}%
    \hfill
    \includegraphics[width=0.24\textwidth]{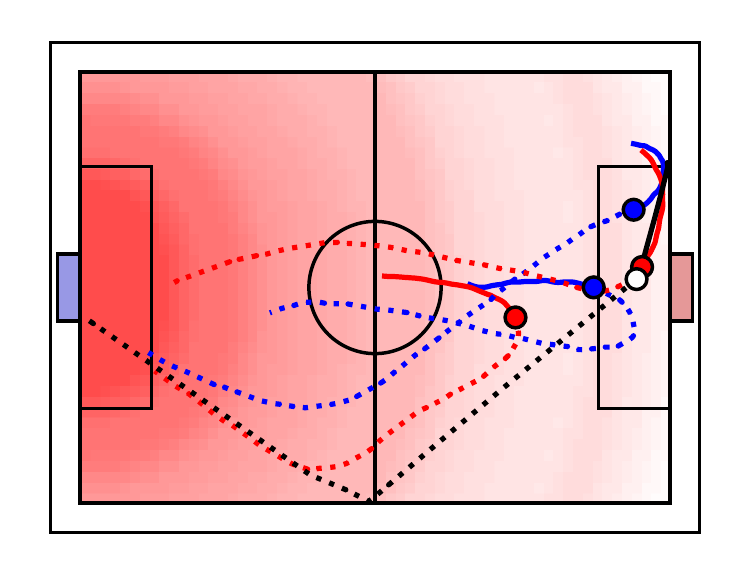}%
    \hfill
    \includegraphics[width=0.24\textwidth]{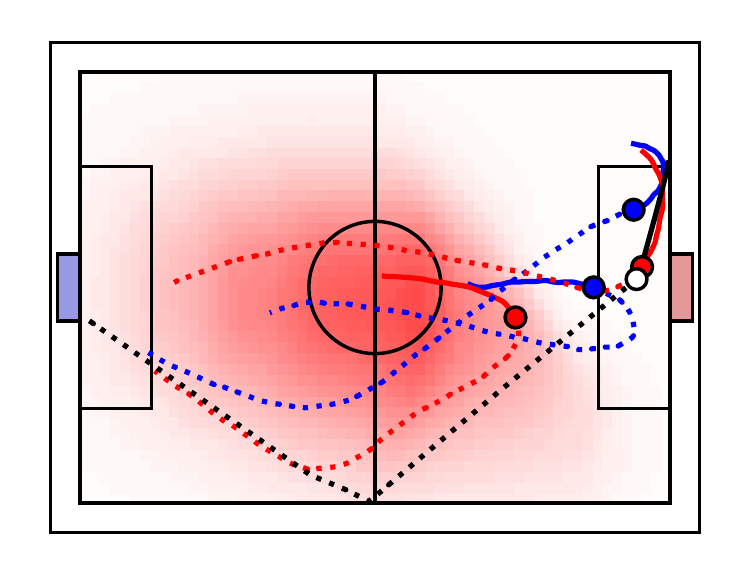}%
    \caption{Visualization of the underlying components in the OBSO model for an example play. From left to right: 
    potential Pitch Control Field (PPCF) model, capturing the probability of successful passes from the player in control to off-ball players; 
    ball transition model, capturing the ball transition probability (from its original position to a location on the pitch); 
    score model, capturing the probability of scoring from any position on the pitch; receiver OBSO scalar field, computed from the three preceding model probabilities. %
    }
    \label{fig:obso_overview}
\end{figure}

OBSO is based on three underlying models that prescribe the probability of the events necessary for a goal to be scored (see Figure~\ref{fig:obso_overview} for an illustrative example): 
i) the Potential Pitch Control Field model (capturing the probability of a pass being controlled by an off-ball player); 
ii) the ball transition model (capturing the dynamics of the ball);
and iii) the score model (capturing the probability of scoring from a point on the pitch). 
For each agent, we build each of these three models, fitting model parameters using data collected from the individual agent's play.
We next provide an overview of these underlying models at a high level, subsequently detailing how parameters are fit to more accurately reflect the physical characteristics of our particular agents and environment. 

\paragraph{Potential Pitch Control Field}

Potential Pitch Control Fields (PPCF) model probabilities of successful passes using physical concepts such as interception time, ball flight time, and player reaction time, to quantify the spatial control of the football pitch by individual players or their associated teams.
Here we use PPCF with a simplified motion model, wherein at the point of passing, we assume the ball travels with a constant reference speed $v_{b}$ to the target destination.
Simultaneously, the off-ball receiving player, $p$, is assumed to continue travelling in a straight line (at their current velocity) during reaction time $t_{r}$.
Following this reaction period, the off-ball player is assumed to travel to the ball's target destination at their reference top speed, $v_{p}$, in a straight line.
Under these simplifying assumptions, we use the PPCF model as detailed in \cite{spearman2018beyond} to compute the probability of the ball being received by a given off-ball player at all possible ball target locations on pitch. 

\paragraph{Ball Transition Model}
The OBSO model assumes that the motion model underlying the spatial movement of the ball, on average across all games and plays, is normally distributed due to the aggregate passes, collisions, interceptions, and other interactions (see second panel of Figure~\ref{fig:obso_overview} for an example).
To account for the agency of the players (i.e., the fact that the passing player will choose destinations with high pitch control for their own team), \cite{spearman2018beyond} modulates the ball transition probability by the pitch control probability.

\paragraph{Score Model}
The scoring model used in OBSO assumes that the probability of scoring is a function of the distance to the goal (see the third panel of Figure~\ref{fig:obso_overview}).

\paragraph{Overall OBSO Model}
Given the above three models, given any instantaneous game state, the overall OBSO probability for a giving receiving player is simply computed as the product of the above probabilities at any receiving position on the pitch (as exemplified in the final panel of Figure~\ref{fig:obso_overview}).

\paragraph{Data Fitting}\label{sec:obso_datafitting}

The OBSO model requires fitting of the various parameters to ensure a reasonable approximation of scoring opportunities.
We fit the OBSO parameters independently for agents and evaluation agents in our evaluation benchmarks;
i.e., a distinct set of parameters (e.g., reference velocities, scoring model parameters) are fit independently at each considered point in training.
Unlike the work of \cite{spearman2018beyond}, which used a shared set of parameters across all  real-world players considered in their analysis, our implementation uses independent parameter fitting to reflect the progression of performance throughout training and the distinction between learning agents and fixed evaluation agents.

In more detail, we conduct parameter fitting for each player type over the tournament dataset discussed in Appendix~\ref{sec:Appendix:Tournament}.
For the PPCF model, we aggregate player speeds across all episodes, and use the mean as the reference, $v_{p}$.
Likewise, the reference ball speed, $v_{b}$, corresponds to the mean ball speed observed across all passes in the dataset.
Finally, we found the specific value of the reaction time ($t_{r} = 0.5 \mbox{sec}$ for all reported figures) to have little impact on the results, compared to the other parameters.

For the raw ball transition model, we filter all consecutive touches by pairs of unique players (i.e., excluding dribbling), using these aggregate transitions to fit the parameters of a 2D Gaussian. 

Finally, for the scoring model, we aggregate all touches immediately resulting in a goal and subsequently use the empirical distribution of goal probabilities as a function of distance to goal in the OBSO model. %

\subsubsection{Additional OBSO Results}
\begin{figure}[t]
    \centering
    \includegraphics[height=0.3\textwidth]{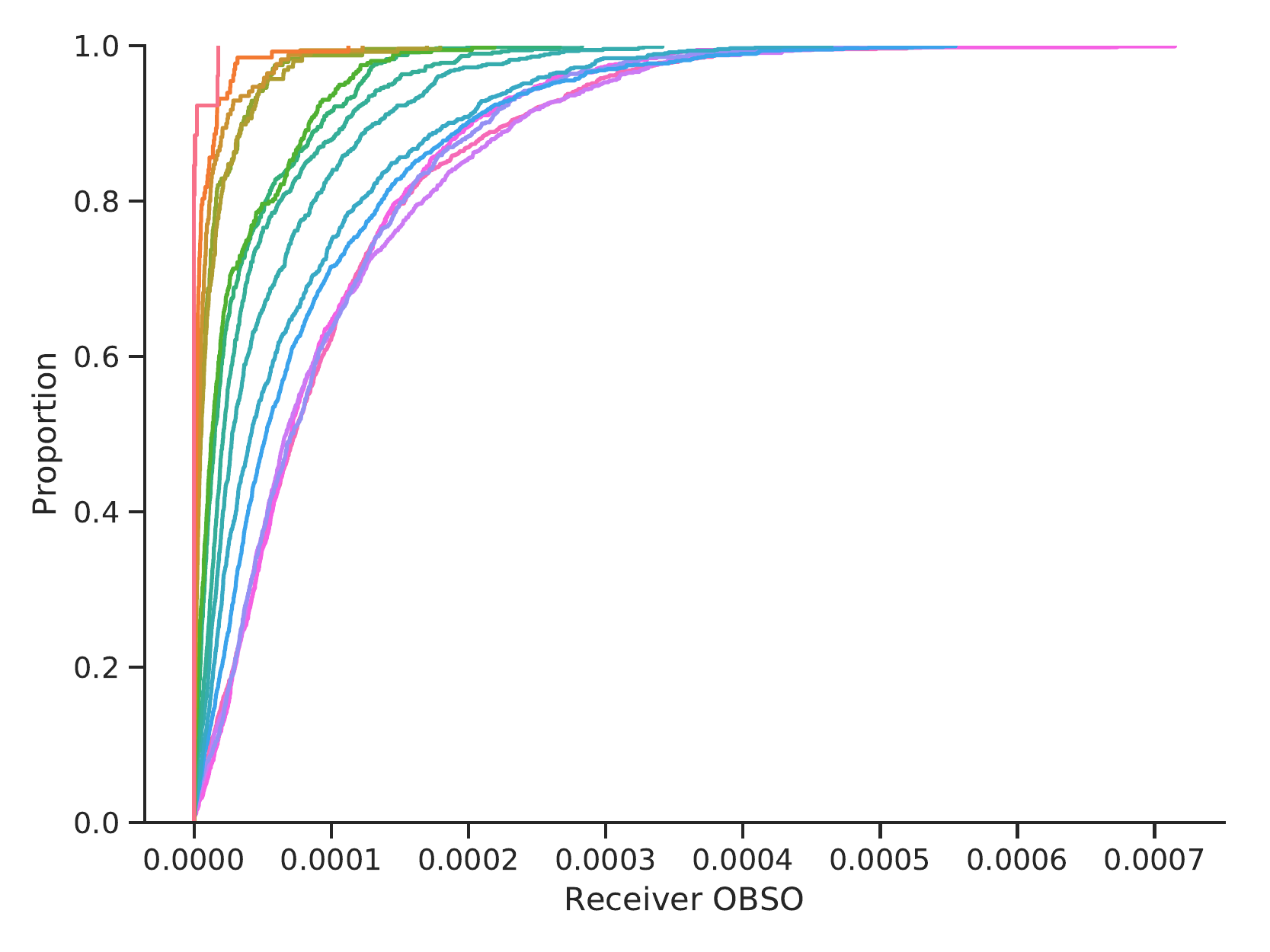}
    \includegraphics[height=0.3\textwidth]{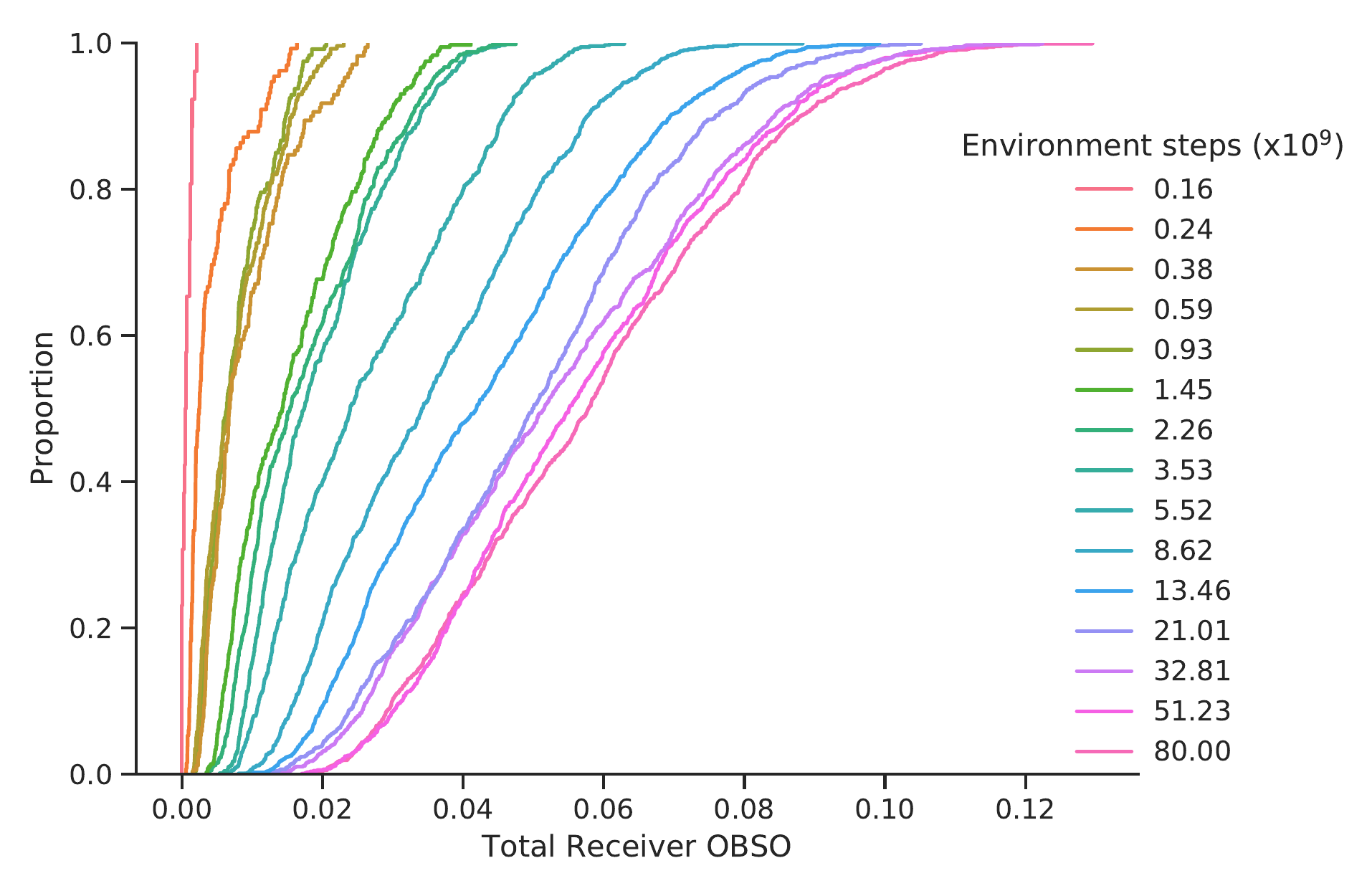}
    \caption{Empirical cumulative distribution functions of (left) receiver OBSO and (right) total receiver OBSO.
    The environment steps for each of the series in these plots correspond to those also used for performance reporting in Figure~\ref{fig:behaviours}.
    }
    \label{fig:obso_results_normalized}
\end{figure}

Analysis of OBSO in the trained agents is informative from an evaluation perspective, as i) OBSO is an externally-defined performance measure used in real-world football analytics, and ii) is not a measure that is explicitly optimized for by the agents in the training pipeline (i.e., our agents are unaware of OBSO as a measure to maximize during training).
Figure~\ref{fig:obso_results_normalized} provides an overview of the evolution of the OBSO measure throughout training. We differentiate between \emph{Receiver OBSO} and \emph{Total Receiver OBSO}; 
For both, we compute the OBSO measure over the entire pitch at the moment of pass initiation for all passes with range 5m or more. 
For \emph{Receiver OBSO}, we use the point evaluation of the OBSO measure for the receiver's position at the moment of pass reception;
this captures team coordination, as it emphasizes both the receiver's current location at the time the pass was initiated.
For \emph{Total Receiver OBSO}, we simply spatially integrate the OBSO scalar field of the receiving player over the entire pitch; 
this measures the overall scoring opportunity of the receiver at the moment of pass initiation.
Each series in this plot corresponds to a snapshot of the trained agents, evaluated against a static set of pre-trained evaluation agents, see Appendix~\ref{sec:Appendix:Tournament} for details.
It is evident that the number of passes with high OBSO consistently increases with training time, along with the average OBSO across all passes.
This result is notable, as it indicates that the trained agents learn to progressively position themselves on the pitch in ways that increases their scoring opportunity, serving as quantifiable evidence of their understanding of the dynamics of the football environment and interactions with the ball and opposing agents.

\subsection{Knowledge Representation}
\label{sec:knowledge_appendix}
We repeat and extend the analysis of agent's knowledge proposed in \cite{JaderbergCTF} to help us understand how the agent
represents its environment, what aspects of game state are emphasized or de-emphasized, and how efficiently it uses its memory module to
represent game features such that the policy head has easy access to corresponding game feature when making decisions.

We say that the agent has game-feature related \emph{knowledge} of a given piece of information if that information can be decoded
with sufficient accuracy from the agent's recurrent hidden state using a linear probe classifier.
In particular, we compare three classification methods: 1) logistic regression on the agent's raw observations of game state, 2) logistic regression on the agent's internal state, and 3) a Multi-layer perceptron (MLP) classifier on the agent's internal state.
1 vs 2 indicates which game features are emphasized or de-emphasized in the agent's internal representation, and 2 vs 3 quantifies how efficient the internal representation of a given game feature is.

We define a set of 44 binary features about present game state (listed in Figure~\ref{acc:knowledge_fea}).
Given a game feature, evaluating the three probe classifiers unfolds into the following steps:
1) For each game feature, we collect a dataset consisting of 500k timesteps sampled from 512 game plays, and for each timestep, we label it as positive if it has the feature, negative otherwise.
2) Given the dataset, we split it into training and test sets by uniform sampling with a randomization seed.
3) Fixing the dataset, we report the performance of each classifier on the test set in terms of balanced accuracy evaluated with 3-fold cross-validation.
4) Finally, we repeat Step 2) to 3) 5 times by splitting the dataset with randomization seeds, and report the test accuracy as the final performance of the probe classifiers.
The full list of results are shown in in Table~\ref{acc:knowledge_fea}.

Further insights about the geometry of the raw observation space and the representation space are gleaned by performing
a t-SNE dimensionality reduction on the raw observations fed to the agent (Figure~\ref{fig:knowledge}, B1) and
the recurrent hidden state (Figure~\ref{fig:knowledge}, Panel B2) respectively. We find strong evidence of cluster structure in the agent's representation reflecting
conjunction of game features: whether the agent has fallen, which player is closest to the ball, and whether the ball is close to home or away goal.

\begin{figure}[t]
    \centering
    \includegraphics[width=\textwidth]{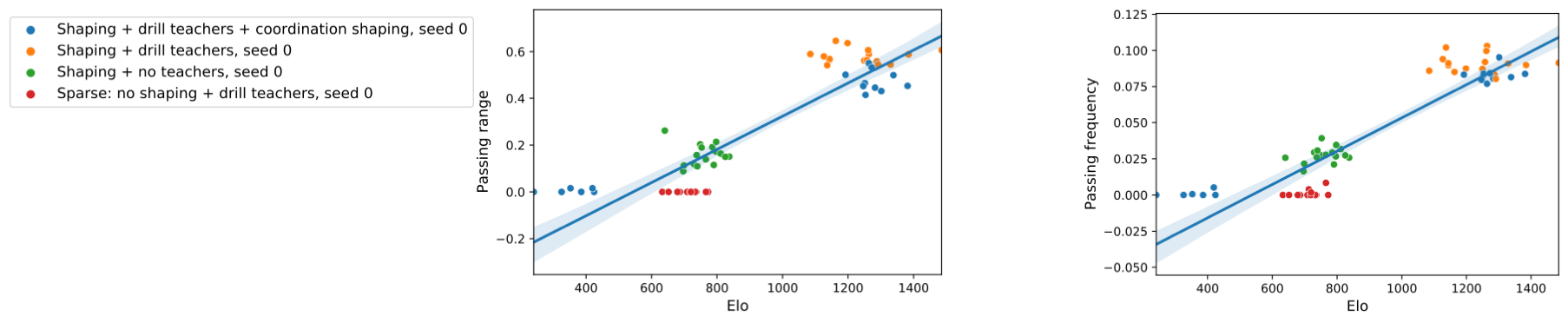}
    \caption{Correlation between agent's performance and behavioural statistics of coordination. We investigated four populations with different training regimes as reported in Section~\ref{sec:Ablation}.
    For each population, we fixed the environment steps to 40e9 and collected the coordination statistics of \emph{passing range} (left) and \emph{passing frequency} (right) for each agent.
    Each data point in the plots represent the mean value of corresponding behavioural metric collected from 64 episodes for one agent in a population.}
    \label{fig:corr_elo_coord}
\end{figure}

Finally, we investigate the correlation between agent's performance and behavioural statistics of coordination, i.e., passing range and frequency (Figure~\ref{fig:corr_elo_coord}).
To do this, we investigate the agents from four different training regimes which are used in the ablation study (Section~\ref{sec:Ablation}). For each regime, we take the snapshot of
corresponding population at 40e9 environment steps. For each agent in the snapshot, which is associated with an Elo score, we collect the behavioural statistics from 64 episodes against the set of evaluation agents
described in \ref{app:Evaluators}. The length of each episode is 3000 environment steps, and the behavioural statistics are averaged over environment steps episode-wise. Such an (Elo, average behavioural statistic) pair
correspond one data point in Figure~\ref{fig:corr_elo_coord}. We find that the coordination statistics are positively correlated with the agent's performance.

\begin{table}[t]
    \centering
    \begin{adjustbox}{width={\textwidth},totalheight={\textheight},keepaspectratio}
    \begin{tabular}{lccc}
        \toprule
        \textbf{Game feature} & \textbf{LR on raw obs.} & \textbf{LR on LSTM state} & \textbf{MLP on LSTM state} \\
        \toprule 
        Ball close to away goal ($\leq$ 2m)$^\textbf{**}$	 	 & 	0.77$\pm$0.01	 	 & 	0.96$\pm$0.00	 	 & 	0.98$\pm$0.00 \\
        Ball close to home goal ($\leq$ 2m)$^\textbf{**}$	 	 & 	0.72$\pm$0.01	 	 & 	0.93$\pm$0.01	 	 & 	0.97$\pm$0.00 \\
        Ball in away half$^\textbf{***}$ 	 & 	0.63$\pm$0.01	 	 & 	0.80$\pm$0.01	 	 & 	0.83$\pm$0.01 \\
        Ball in home half$^\textbf{***}$	 	 & 	0.62$\pm$0.01	 	 & 	0.79$\pm$0.00	 	 & 	0.84$\pm$0.00 \\
        Agent on the ground$^\textbf{}$	 	 & 	0.98$\pm$0.00	 	 & 	0.99$\pm$0.00	 	 & 	0.99$\pm$0.00 \\
        Agent in away half$^\textbf{**}$ 	 & 	0.56$\pm$0.00	 	 & 	0.75$\pm$0.01	 	 & 	0.80$\pm$0.01 \\
        Agent in home half$^\textbf{*}$	 	 & 	0.55$\pm$0.02	 	 & 	0.74$\pm$0.01	 	 & 	0.79$\pm$0.00 \\
        Dribbling$^\textbf{***}$ 	 & 	0.63$\pm$0.01	 	 & 	0.65$\pm$0.01	 	 & 	0.74$\pm$0.00 \\
        Agent intercepting the ball$^\textbf{**}$	 	 & 	0.65$\pm$0.01	 	 & 	0.70$\pm$0.01	 	 & 	0.77$\pm$0.01 \\
        Teammate intercepting the ball$^\textbf{}$	 	 & 	0.75$\pm$0.12	 	 & 	0.69$\pm$0.10	 	 & 	0.73$\pm$0.13 \\
        Opponent intercepting the ball$^\textbf{}$	 	 & 	0.71$\pm$0.03	 	 & 	0.79$\pm$0.01	 	 & 	0.81$\pm$0.03 \\
        Passing ($\geq$ 10m)$^\textbf{}$	 	 & 	0.78$\pm$0.02	 	 & 	0.84$\pm$0.01	 	 & 	0.86$\pm$0.04 \\
        Passing ($\geq$ 15m)$^\textbf{}$	 	 & 	0.82$\pm$0.01	 	 & 	0.87$\pm$0.03	 	 & 	0.90$\pm$0.02 \\
        Passing ($\geq$ 5m)$^\textbf{.}$	 	 & 	0.78$\pm$0.01	 	 & 	0.84$\pm$0.01	 	 & 	0.87$\pm$0.02 \\
        Shooting$^\textbf{}$	 	 & 	0.81$\pm$0.01	 	 & 	0.93$\pm$0.01	 	 & 	0.94$\pm$0.01 \\
        Agent close to away goal ($\leq$ 2m)$^\textbf{}$	 	 & 	0.78$\pm$0.04	 	 & 	0.91$\pm$0.03	 	 & 	0.94$\pm$0.03 \\
        Agent close to home goal ($\leq$ 2m)$^\textbf{}$	 	 & 	0.81$\pm$0.01	 	 & 	0.93$\pm$0.02	 	 & 	0.96$\pm$0.01 \\
        Agent close to ball ($\leq$ 2m)$^\textbf{}$	 	 & 	0.75$\pm$0.01	 	 & 	0.92$\pm$0.01	 	 & 	0.93$\pm$0.01 \\
        Agent closest to away goal$^\textbf{***}$	 	 & 	0.55$\pm$0.01	 	 & 	0.69$\pm$0.01	 	 & 	0.76$\pm$0.01 \\
        Agent closest to home goal$^\textbf{*}$	 	 & 	0.54$\pm$0.00	 	 & 	0.67$\pm$0.01	 	 & 	0.72$\pm$0.01 \\
        Agent closest to ball$^\textbf{}$	 	 & 	0.65$\pm$0.01	 	 & 	0.88$\pm$0.00	 	 & 	0.88$\pm$0.00 \\
        Teammate close to away goal ($\leq$ 2m)$^\textbf{}$	 	 & 	0.85$\pm$0.03	 	 & 	0.94$\pm$0.01	 	 & 	0.95$\pm$0.00 \\
        Teammate close to home goal ($\leq$ 2m)$^\textbf{*}$	 	 & 	0.80$\pm$0.02	 	 & 	0.93$\pm$0.01	 	 & 	0.96$\pm$0.01 \\
        Teammate close to ball ($\leq$ 2m)$^\textbf{}$	 	 & 	0.69$\pm$0.00	 	 & 	0.87$\pm$0.01	 	 & 	0.88$\pm$0.01 \\
        Teammate closest to away goal$^\textbf{*}$	 	 & 	0.54$\pm$0.01	 	 & 	0.65$\pm$0.01	 	 & 	0.70$\pm$0.00 \\
        Teammate closest to home goal$^\textbf{**}$	 	 & 	0.53$\pm$0.01	 	 & 	0.64$\pm$0.01	 	 & 	0.67$\pm$0.01 \\
        Teammate closest to ball$^\textbf{}$	 	 & 	0.62$\pm$0.01	 	 & 	0.84$\pm$0.01	 	 & 	0.85$\pm$0.01 \\
        Opponent close to away goal ($\leq$ 2m)$^\textbf{*}$	 	 & 	0.72$\pm$0.00	 	 & 	0.89$\pm$0.02	 	 & 	0.95$\pm$0.00 \\
        Opponent close to home goal ($\leq$ 2m)$^\textbf{**}$	 	 & 	0.77$\pm$0.01	 	 & 	0.89$\pm$0.01	 	 & 	0.95$\pm$0.01 \\
        Opponent close to ball ($\leq$ 2m)$^\textbf{}$	 	 & 	0.67$\pm$0.01	 	 & 	0.82$\pm$0.00	 	 & 	0.82$\pm$0.02 \\
        Opponent closest to away goal$^\textbf{}$	 	 & 	0.54$\pm$0.00	 	 & 	0.63$\pm$0.02	 	 & 	0.64$\pm$0.02 \\
        Opponent closest to home goal$^\textbf{}$	 	 & 	0.56$\pm$0.01	 	 & 	0.64$\pm$0.01	 	 & 	0.64$\pm$0.00 \\
        Opponent closest to ball$^\textbf{}$	 	 & 	0.62$\pm$0.01	 	 & 	0.80$\pm$0.01	 	 & 	0.79$\pm$0.01 \\
        Teammate on the ground$^\textbf{}$	 	 & 	0.87$\pm$0.01	 	 & 	0.79$\pm$0.01	 	 & 	0.81$\pm$0.01 \\
        Teammate in home half$^\textbf{}$	 	 & 	0.57$\pm$0.00	 	 & 	0.73$\pm$0.00	 	 & 	0.75$\pm$0.01 \\
        Teammate in away half$^\textbf{}$	 	 & 	0.56$\pm$0.00	 	 & 	0.73$\pm$0.01	 	 & 	0.76$\pm$0.02 \\
        Opponent on the ground$^\textbf{.}$	 	 & 	0.85$\pm$0.00	 	 & 	0.68$\pm$0.00	 	 & 	0.69$\pm$0.01 \\
        Opponent in home half$^\textbf{}$	 	 & 	0.57$\pm$0.01	 	 & 	0.74$\pm$0.02	 	 & 	0.77$\pm$0.01 \\
        Opponent in away half$^\textbf{*}$	 	 & 	0.59$\pm$0.02	 	 & 	0.76$\pm$0.01	 	 & 	0.79$\pm$0.01 \\
        \bottomrule
    \end{tabular}
    \end{adjustbox}
    \caption{Representational efficiency of the agent's LSTM state on a list of high-level game features. Each cell represents the mean classification accuracy and standard deviation of corresponding classification method on a given game feature. The asterisk on the name of a game feature indicates the significance level of a two-sided t-test comparing the performance of corresponding linear classifier and  MLP classifier on the agent's LSTM state. No asterisk means that the linear classifier perform as as well as the MLP classifier and suggest that the agent policy head has easy access to the knowledge encoded in its LSTM state when generating actions.}
    \label{acc:knowledge_fea}
\end{table}

\end{document}